\documentclass{article} % For LaTeX2e
\usepackage[final]{colm2026_conference}
\usepackage{microtype}
\usepackage{hyperref}
\usepackage{url}
\usepackage{booktabs}
\usepackage{tabularx}
\usepackage{amsmath}
\usepackage[T1]{fontenc}
\usepackage{listings}
\usepackage{multirow}
\usepackage{xcolor}
\usepackage{subcaption}
\usepackage{tcolorbox}
\usepackage{nicefrac}
\usepackage{bm}
\usepackage{placeins}
\tcbuselibrary{listings,skins}

\definecolor{codebg}{gray}{0.95}

\lstdefinestyle{python}{
    language=Python,
    basicstyle=\small\ttfamily,
    keywordstyle=\color{blue!70!black},
    stringstyle=\color{red!60!black},
    commentstyle=\color{green!50!black},
    showstringspaces=false,
    breaklines=true,
    numbers=none,
    mathescape=false,
}

\newcounter{listingnum}

\newtcblisting{code}[1]{
    listing only,
    listing options={style=python},
    colback=codebg,
    colframe=codebg,
    arc=5pt,
    boxrule=0pt,
    left=8pt,
    right=8pt,
    top=6pt,
    bottom=6pt,
    before skip=8pt,
    after skip=0pt,
    after={\refstepcounter{listingnum}\noindent\hfil{\small\textit{Listing \thelistingnum: #1}}\hfil\vspace{8pt}},
}

\newtcblisting{codeoutput}[1]{
    listing only,
    listing options={basicstyle=\footnotesize\ttfamily, breaklines=true, numbers=none},
    colback=codebg,
    colframe=codebg,
    arc=5pt,
    boxrule=0pt,
    left=8pt,
    right=8pt,
    top=6pt,
    bottom=6pt,
    before skip=8pt,
    after skip=0pt,
    after={\refstepcounter{listingnum}\noindent\hfil{\small\textit{Listing \thelistingnum: #1}}\hfil\vspace{8pt}},
}

\usepackage{lineno}

\definecolor{darkblue}{rgb}{0, 0, 0.5}
\hypersetup{
    colorlinks=true,
    citecolor=darkblue,
    linkcolor=darkblue,
    urlcolor=darkblue,
    pdftitle={Autorubric: A Unifying Framework for Rubric-Based LLM Evaluation on Non-Verifiable Tasks},
    pdfauthor={Delip Rao and Chris Callison-Burch}
}

\title{\texttt{Autorubric}: A Unifying Framework for Rubric-Based LLM Evaluation on Non-Verifiable Tasks}

\author{Delip Rao\thanks{Corresponding author} \\
University of Pennsylvania\\
\texttt{delip@seas.upenn.edu} \\
\And Chris Callison-Burch\\
University of Pennsylvania\\
\texttt{ccb@seas.upenn.edu}
}

\begin{document}

\ifcolmsubmission
\linenumbers
\fi

\maketitle
\begin{abstract}
Rubric-based LLM judges have become indispensable for evaluating and optimizing systems on non-verifiable tasks, where success cannot be reduced to exact programmatic checks. Yet the underlying judges remain vulnerable to position bias, stochastic inconsistency, criterion conflation, forced judgments under uncertainty, and model-dependent calibration. Rubrics structure evaluation but do not eliminate these failures; they add consequential choices about criterion design, scale types, weighting, aggregation, abstention, calibration, and reliability measurement. Despite extensive research across LLM evaluation, educational measurement, and psychometrics, the relevant methods remain scattered across papers and partial implementations. Researchers therefore pay a \emph{reinvention tax}, repeatedly rebuilding evaluation machinery instead of accumulating knowledge on a common substrate. We introduce \texttt{Autorubric}, an open-source framework that makes rubric and judge choices explicit, reusable, and auditable. Through a unified API, it supports atomic evaluation of mixed criterion types alongside configurable bias mitigations, calibration, ensembling, abstention, and psychometric diagnostics. Evaluations spanning college chemistry grading, deep-research systems, and CHARM-100---a new mixed-criterion chatbot benchmark---reveal criterion-specific failures, systematic judge-family differences, and configuration effects that do not support a universal mitigation stack. We further demonstrate how per-criterion scores and explanations can support agent skill revision and reward modeling for reinforcement learning. By providing shared infrastructure from measurement through optimization, \texttt{Autorubric} gives the community a common basis for comparing methods, reproducing evaluation choices, and accumulating evidence across studies.
\end{abstract}

\section{Introduction}
\label{sec:introduction}

The LLM-as-a-Judge (LaaJ) paradigm is the default approach for evaluating text generation at scale~\citep{zheng2023judging, liu2023geval, gu2024survey}, yet the techniques that make LaaJ evaluation reliable---ensemble judging~\citep{verga2024replacing}, position bias mitigation~\citep{wang2023fairevaluators}, explicit abstention~\citep{xin2021abstention}---are scattered across papers with inconsistent terminology and not consistently applied in LLM-based rubric evaluations. Practitioners must compose these techniques ad hoc, resulting in repeated reimplementation. A separate gap compounds the problem: educational measurement and psychometrics offer decades of methodology on designing and scoring rubrics and inter-rater reliability~\citep{mckeown2018heqco, brookhart2018criteria}, yet this body of work has yet to be systematically applied to LaaJ evaluations.

Part of this problem is the perception that a rubric is \textit{just} a prompt. However, when examined deeply, rubrics offer a plethora of design and operationalization choices, each consequential to the evaluation problem and judge model at hand. For starters, rubrics can be `holistic' (single score over multiple criteria) or `analytical'. Analytic rubrics decompose evaluation into \textbf{independent} criteria scored separately, conferring three advantages: (1) per-criterion evaluation reduces the risk of criterion conflation and halo effects~\citep{lee2025checkeval, wei2025rocketeval}; (2) independent criterion scores enable reliability measurement via Cohen's $\kappa$ and weighted $\kappa$, so practitioners can identify which criteria are unreliable; and (3) structured per-criterion verdicts and explanations serve as optimization signals---a system that knows which criteria it fails can target those dimensions, whereas a holistic score provides no such decomposition. We review many more such design choices in Section~\ref{sec:background}.

\begin{figure*}[t]
    \centering
    \includegraphics[width=\linewidth]{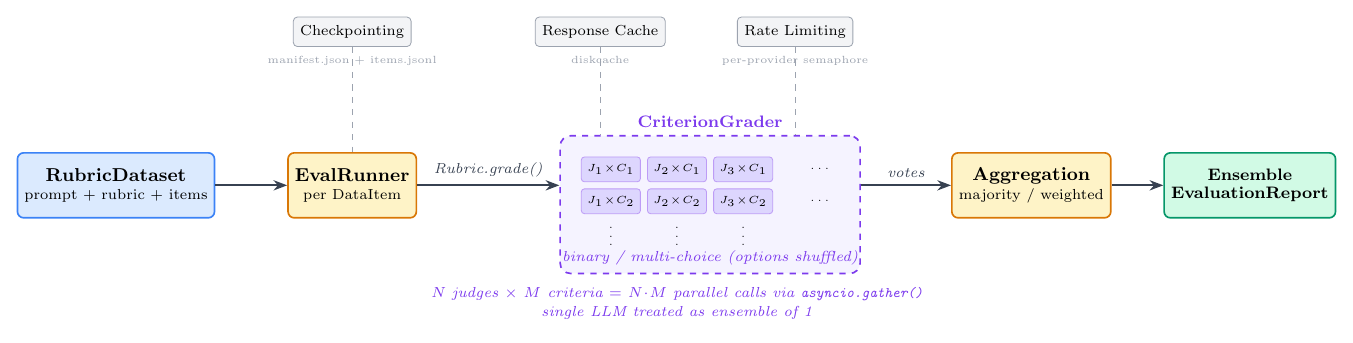}
    \caption{\texttt{Autorubric} evaluation pipeline. A \texttt{RubricDataset} packages a task prompt, rubric, and submissions. The \texttt{CriterionGrader} issues $N \times M$ independent LLM calls ($N$ judges, $M$ criteria) in parallel, each evaluating one criterion under one judge with optional option shuffling. Per-criterion votes are combined by a configurable aggregation strategy to produce an \texttt{EnsembleEvaluationReport}. Response caching, per-provider rate limiting, checkpoint-based resumable evaluation, concurrent execution, and cost tracking per LLM call with per-item and run-level totals operate under an \texttt{EvalRunner}. Details are in Appendix~\ref{appendix:framework-details}.}
    \label{fig:architecture}
\end{figure*}

This paper presents \texttt{Autorubric}, an open-source framework\footnote{Project website and documentation: \url{https://autorubric.org/}. Framework and API descriptions in this paper target release v1.5.3; readers should consult the public site for the latest API documentation. Reported experiments retain their contemporaneous configurations and were not rerun under v1.5.3.} that unifies rubric-based LLM evaluation design and operationalization choices with opinionated defaults drawn from LLM-as-a-judge, education testing, and psychometrics literature. Its design centers on per-criterion atomic evaluation, bias mitigations for position and criterion conflation, ensemble judging with configurable aggregation, few-shot calibration with verdict-balanced sampling, and psychometric reliability metrics. The framework supports heterogeneous mixtures of binary, ordinal, and nominal criteria within a single rubric, enabling richer evaluations.  Figure~\ref{fig:architecture} shows the evaluation pipeline.

Our contributions are:
\begin{enumerate}
    \item \texttt{Autorubric}, an open-source framework unifying rubric-based LLM evaluation---analytic rubrics with mixed criterion types, ensemble judging, few-shot calibration, bias mitigations, and psychometric reliability metrics---with opinionated defaults (Section~\ref{sec:background}).
    \item CHARM-100, a synthetically authored chatbot evaluation dataset combining binary, ordinal, and nominal criteria with per-sample reference labels and a recorded 50-item second-annotation agreement audit (Section~\ref{sec:charm100}).
    \item Validation on three benchmarks with an exploratory configuration-sensitivity analysis across three judge-model families (Section~\ref{sec:evaluation}).
    \item Two downstream applications of \texttt{Autorubric} demonstrating that per-criterion rubric scores serve as optimization signals for agent skill improvement (Section~\ref{sec:skill-improvement}) and reinforcement learning with rubric-based rewards (Section~\ref{sec:rlrr}).
\end{enumerate}

\section{Background and framework design}
\label{sec:background}

A rubric is a scoring instrument comprising \emph{criteria} that specify what is being evaluated and performance-level descriptions that characterize quality~\citep{brookhart2018criteria}. Rubric criteria can have associated weights~\citep{kim2024prometheus, gunjal2025rubrics, hong2026rulers}. While holistic rubrics assign a single overall score, analytic rubrics decompose evaluation into separate criteria scored independently~\citep{mrangu2022rubric}. \texttt{Autorubric} adopts analytic rubrics as the default, as motivated in Section~\ref{sec:introduction}. Figure~\ref{fig:design-space} illustrates the space of rubric design and implementation choices that we explore in our framework.

\textbf{Criterion types.} Individual criteria take three forms. \emph{Binary} criteria (\texttt{MET}/\texttt{UNMET}) express thresholded decisions using the smallest label space. \emph{Ordinal} criteria use ordered levels (Likert scales) to capture gradations; we encourage bounded scales with clear behavioral anchors. Direct LLM scoring can concentrate on a single label (e.g., 3 on a 1--5 scale) and produce tied integer ratings even when decimals are requested~\citep{liu2023geval}. \emph{Nominal} criteria offer unordered categories for classification-style evaluation. \texttt{Autorubric} supports all three types, with explicit numeric values (0--1) for multi-choice options to decouple score from presentation order. Reliability depends on the construct, label design, and judge configuration, so agreement should be assessed with type-appropriate metrics rather than assumed from criterion type alone. As a design choice, continuous-valued criteria are intentionally excluded in favor of bounded categorical options with explicit numeric mappings.

\textbf{Weighting and aggregation.} Criteria carry configurable positive or negative weights. Negative criteria apply explicit penalties when specified anti-patterns are detected. After applying the configured abstention strategy, let $\mathcal{I}$ be the criteria retained for scoring and $\tilde{v}_i$ their effective values. Define $S = \sum_{i\in\mathcal{I}} \tilde{v}_i w_i$, $W_+ = \sum_{i\in\mathcal{I}:w_i>0} w_i$, and $W_- = \sum_{i\in\mathcal{I}:w_i<0} |w_i|$. The normalized score is
\begin{equation}
\label{eq:score-aggregation}
\text{score} =
\begin{cases}
\operatorname{clip}_{[0,1]}(S/W_+), & W_+ > 0, \\
\operatorname{clip}_{[0,1]}(1 + S/W_-), & W_+ = 0,\ W_- > 0, \\
0, & W_+ = W_- = 0,
\end{cases}
\end{equation}
where $w_i$ denotes weight and, for ordinary verdicts, $\tilde{v}_i$ is 1 for \texttt{MET}, 0 for \texttt{UNMET}, or the option's explicit value for multi-choice criteria; retained abstentions use the strategy-defined effective value. Also, $\operatorname{clip}_{[0,1]}(x)=\max(0,\min(1,x))$. When $W_+>0$, negative weights subtract penalties while $W_+$ defines the maximum attainable positive score. When $W_+=0$ and $W_->0$, the effective rubric is penalty-only: its score starts at 1 and falls in proportion to the triggered penalty weight. If no nonzero-weight criteria remain after abstention handling, the score is 0. Listings 1--3 in Appendix~\ref{appendix:framework-details} illustrate rubric definitions with all three criterion types.

\textbf{Evaluation modes and judging strategies.} \texttt{Autorubric} implements pointwise, reference-free evaluation by default; optional reference submissions are supported. Each criterion is evaluated in a separate LLM call to reduce criterion conflation and halo effects~\citep{lee2025checkeval, wei2025rocketeval}. Ensemble grading via diverse-model panels is supported because such panels can outperform individual judges in some settings~\citep{verga2024replacing}, with majority vote, weighted vote, unanimous, and any-vote aggregation strategies (Listing 4). The grader makes $N \times M$ concurrent calls ($N$ judges, $M$ criteria) with mean inter-judge agreement tracked as a reliability indicator.

\textbf{Calibration and reasoning.} Few-shot calibration includes example submissions with correct verdicts drawn from a training split, with verdict balancing to prevent the judge from inferring a base-rate prior~\citep{hong2026rulers}. Reasoning-enhanced judging is configurable via thinking levels or explicit token budgets per judge. Evidence for reasoning benefits is mixed~\citep{gunjal2025rubrics, haldar2025ratingroulette}, so thinking is best treated as a tunable option. Code examples are in Appendix~\ref{appendix:framework-details}.

\section{Failure modes in LLM-based evaluation and \texttt{Autorubric} mitigations}
\label{sec:failure-modes}

\textbf{Position bias.} Many LLM judges are sensitive to option position~\citep{wang2023fairevaluators, zheng2023judging}. \texttt{Autorubric} randomizes option order per evaluation by default, using explicit numeric values to decouple score from position. Shuffling is deterministic via a per-item seed derived from the master seed, so reruns are reproducible.

\begin{table}[h]
\centering
\small
\begin{tabularx}{\linewidth}{@{}llX@{}}
\toprule
Failure Mode & Mitigation & \texttt{Autorubric} Mechanism \\
\midrule
Position bias & Option shuffling & \texttt{shuffle\_options} + value-based scoring \\
Low reliability & Ensemble judging & Multi-judge \texttt{CriterionGrader} \\
Criterion conflation & Atomic decomposition & Independent per-criterion LLM calls \\
Uncertainty & Explicit abstention & \texttt{CANNOT\_ASSESS} + strategies \\
Opacity & Per-criterion explanations & Mandatory \texttt{reason} field per verdict \\
\bottomrule
\end{tabularx}
\caption{Failure modes in LLM-based evaluation and corresponding \texttt{Autorubric} mitigations.}
\label{tab:failure-modes}
\end{table}

\textbf{Low reliability.} Individual LLM judgments exhibit high unexplained variance~\citep{feuer2025judgment} and stochastic self-inconsistency~\citep{haldar2025ratingroulette}. Multi-judge ensembles with per-criterion voting reduce variance and mitigate model-specific biases. \texttt{Autorubric} controls non-LLM randomness via a single master seed persisted in the experiment manifest.

\textbf{Criterion conflation.} When multiple quality dimensions are evaluated together, judgments conflate distinct constructs~\citep{lee2025checkeval, wei2025rocketeval}. \texttt{Autorubric} evaluates each criterion in a separate LLM call, with concurrent execution and prompt caching offsetting evaluation throughput and cost respectively.

\textbf{Uncertainty.} Selective prediction permits abstention rather than a forced label on uncertain examples~\citep{xin2021abstention}. Some submissions similarly do not provide enough information to assess a particular rubric criterion. \texttt{Autorubric} provides a native \texttt{CANNOT\_ASSESS} verdict with configurable strategies: \texttt{SKIP}, \texttt{ZERO}, \texttt{PARTIAL}, or \texttt{FAIL}.

\textbf{Interpretability and auditability.} Every verdict includes a mandatory explanation field for enabling evaluation audits and downstream optimization (See Sec.~\ref{sec:skill-improvement} for an example application). In ensemble mode, per-judge reasoning is preserved so reviewers can inspect disagreement rationales, though LLM explanations are post-hoc rationalizations~\citep{turpin2023language} and not a replacement for human review.

\section{Evaluation}
\label{sec:evaluation}

In addition to an extensive suite of unit tests for each \texttt{Autorubric} feature, we empirically sanity check the framework with benchmarks.

\subsection{College-level chemistry grading (RiceChem)}
\label{sec:ricechem}

The RiceChem dataset~\citep{sonkar2024automatedlonganswergrading} comprises 1,240 human-graded long-form student responses to four college-level chemistry exam questions, each graded against an independent binary rubric for a total of 27 criteria. The four questions cover ionization energies (Q1, 8 criteria, 327 students), quantized absorption versus photoejection (Q2, 6 criteria, 317), hybrid orbital analysis (Q3, 7 criteria, 298), and the Law of Multiple Proportions (Q4, 6 criteria, 298). Responses average approximately 120 words. \citet{sonkar2024automatedlonganswergrading} frame grading as \emph{rubric entailment}: for each (response, criterion) pair, determine whether the response satisfies the criterion. Criterion weights, not provided in the raw data, were inferred via least-squares regression from TA-assigned scores. The inferred weights cluster near integer values with $R^2 \geq 0.985$ for three of four questions (Appendix~\ref{appendix:ricechem}).

This benchmark exercises two capabilities: per-criterion binary evaluation with criterion-level agreement metrics, and few-shot calibration via \texttt{FewShotConfig} with verdict-balanced sampling.\footnote{Unless otherwise noted, benchmark runs use one judge, option shuffling, majority aggregation for binary criteria, and \texttt{SKIP} for unassessable criteria. Few-shot use is experiment-specific: the primary RiceChem comparison is 0-shot versus 5-shot, the primary ResearcherBench and CHARM-100 evaluations are zero-shot, and the CHARM configuration-sensitivity study uses 3-shot calibration except in the explicitly ablated conditions. Each criterion is evaluated atomically via a structured prompt that elicits a JSON verdict (\texttt{MET}/\texttt{UNMET}/\texttt{CANNOT\_ASSESS}) with a 1--2 sentence evidence-citing explanation. Appendix~\ref{appendix:framework-details} distinguishes historical experiment provenance from the v1.5.3 API. All models were accessed in Feb.--March 2026.} Published baselines---zero-shot GPT-4 (70.9\% accuracy) and fine-tuned RoBERTa+MNLI (86.8\%)---serve as reference points.

\begin{table}[h]
\centering
\small
\begin{tabular}{lc}
\toprule
\textbf{Method} & \textbf{Accuracy} \\
\midrule
\multicolumn{2}{l}{\citep{sonkar2024automatedlonganswergrading}:} \\
\quad $\hookrightarrow$ GPT-4 zero-shot & 70.9\% \\
\quad $\hookrightarrow$ RoBERTa+MNLI fine-tuned & 86.8\% \\
\midrule
\multicolumn{2}{l}{\texttt{Autorubric} + Gemini-3-Flash:} \\
\quad $\hookrightarrow$ 0-shot & 78.0\% \\
\quad $\hookrightarrow$ 5-shot & 80.7\% \\
\bottomrule
\end{tabular}
\caption{Comparison on the RiceChem rubric-entailment task (1,240 student responses, 27 binary criteria). \texttt{Autorubric} values are from the paired rerun over all 819 held-out response--criterion decisions; baselines are from \citet{sonkar2024automatedlonganswergrading}.}
\label{tab:ricechem-results}
\end{table}

Following \citet{sonkar2024automatedlonganswergrading}, we use the same 80-10-10 split and report micro-averaged accuracy on the held-out test set. In a paired rerun over all 819 response--criterion decisions, 5-shot calibration improves accuracy from 78.0\% (639/819) to 80.7\% (661/819), a gain of 2.7 percentage points; 54 decisions improve and 32 degrade. Because multiple criterion decisions are nested within each student response, we treat this gain as descriptive and do not use a decision-level McNemar test or flat bootstrap interval as cluster-adjusted inference. The gap to fine-tuned RoBERTa+MNLI (86.8\%) is expected for a zero/few-shot approach. Appendix~\ref{appendix:ricechem} reports the paired per-question results and clearly labels two independent diagnostic runs.

\subsection{Deep research system evaluation (ResearcherBench)}
\label{sec:researcherbench}

ResearcherBench~\citep{xu2025researcherbenchevaluatingdeepai} evaluates Deep Research\footnote{\url{https://openai.com/index/introducing-deep-research/}} like systems---agents that perform multi-step literature search, synthesis, and reasoning---on 65 expert-curated questions spanning 34 AI research subjects. Each question carries a per-item rubric of weighted binary criteria designed by experienced AI researchers. Rubrics contain 931 total criteria (mean 14.3 per question), with weights from 1 (nice-to-have) to 3 (core finding). The benchmark defines a coverage score\footnote{The coverage score, introduced by \citet{xu2025researcherbenchevaluatingdeepai}, is the weighted proportion of binary criteria satisfied: $\text{Coverage} = \sum_i w_i c_i / \sum_i w_i$, where $c_i \in \{0,1\}$ indicates whether criterion $i$ is met and $w_i$ is its importance weight.} identical to \texttt{Autorubric}'s normalized weighted score for binary criteria (Equation~\ref{eq:score-aggregation}).

We evaluate three systems (OpenAI DeepResearch, Gemini DeepResearch, Grok3 DeepSearch) using two judges: Claude Sonnet-4.5 and Gemini-3-Flash, for a total of 5,586 criterion-level judgments.

\begin{table}[h]
\centering
\small
\begin{tabular}{lccc}
\toprule
& \multicolumn{2}{c}{\textbf{Autorubric}} & \textbf{\citet{xu2025researcherbenchevaluatingdeepai}} \\
\cmidrule(lr){2-3} \cmidrule(lr){4-4}
& \textbf{Sonnet-4.5} & \textbf{Gemini-3-Flash} & \textbf{Sonnet-3.5} \\
\midrule
OpenAI DeepResearch & \shortstack{0.620\\{[.569, .670]}} & \shortstack{0.771\\{[.723, .818]}} & 0.703 \\
Gemini DeepResearch & \shortstack{0.692\\{[.640, .741]}} & \shortstack{0.810\\{[.751, .864]}} & 0.693 \\
Grok3 DeepSearch    & \shortstack{0.579\\{[.520, .637]}} & \shortstack{0.618\\{[.569, .667]}} & 0.441 \\
\bottomrule
\end{tabular}
\caption{ResearcherBench mean coverage scores with 95\% bootstrap CIs (65 questions, 931 criteria). The top-two systems' CIs overlap for Sonnet-4.5; a paired permutation test (Table~\ref{tab:rb-paired-permutation}) shows the gap is significant under Sonnet-4.5 but not Gemini-3-Flash. The \citet{xu2025researcherbenchevaluatingdeepai} column uses Sonnet-3.5 (CIs not available). Cost analysis is in Appendix~\ref{appendix:researcherbench}.}
\label{tab:researcherbench-results}
\end{table}

Table~\ref{tab:researcherbench-results} reports mean coverage scores with 95\% bootstrap confidence intervals. Both \texttt{Autorubric} judges produce the same aggregate ranking (Gemini $>$ OpenAI $>$ Grok3), though CIs overlap for the top two. A paired permutation test (Table~\ref{tab:rb-paired-permutation}) confirms the Gemini--OpenAI gap is significant under Sonnet-4.5 ($p = 0.003$, Cohen's $d = 0.39$) but not under Gemini-3-Flash ($p = 0.219$), making the top-two ranking judge-dependent. Cross-judge Spearman correlations on per-question scores are moderate to strong ($\rho = 0.54$--$0.82$, $p < 0.001$), indicating judges agree on \emph{which questions are hard} more consistently than on \emph{which system is best}. At the criterion level, inter-judge agreement is moderate (pooled $\kappa = 0.53$, 78.5\% raw agreement), with the non-exclusive keyword taxonomy yielding a higher disagreement rate for critical-analysis criteria (27.9\%) than for enumeration and depth criteria (14--15\%; Table~\ref{tab:rb-disagreement-taxonomy}). \texttt{Autorubric}'s cost and token usage tracking, for example, reveals Gemini-3-Flash is 5--6$\times$ cheaper while preserving the aggregate ranking (Appendix~\ref{appendix:researcherbench}), giving practitioners evidence for cost--quality tradeoffs in large-scale deployments.

\subsection{Chatbot assessment with heterogeneous criteria (CHARM-100)}
\label{sec:charm100}

To validate \texttt{Autorubric}'s support for multiple criterion types and heterogeneous criteria, we introduce CHARM-100 (\textbf{Ch}atbot \textbf{A}ssessment with Mixed \textbf{R}ubric \textbf{M}etrics), 100 annotated single-turn chatbot conversations evaluated against six criteria spanning three types: one binary (factual accuracy), four ordinal (satisfaction, helpfulness, naturalness, specificity), and one nominal (response length)\footnote{Existing evaluation benchmarks adopt a single scale type: Likert scales~\citep{zheng2023judging}, pairwise judgments~\citep{li2025arenahard}, binary checklists~\citep{lin2024wildbench}, or multi-dimensional ratings~\citep{wang2024helpsteer2, hashemi2024llmrubric}. None combine ordinal, nominal, and binary criteria in a single rubric with per-sample reference labels.}. The conversations and reference labels were generated synthetically to control quality-label distributions. Cross-criteria conflicts (e.g., factually wrong but naturally written) are designed to expose reliance on a single overall-quality shortcut. Reference labels are broadly distributed (mean normalized entropy 0.92), and a recorded second annotation pass covers a stratified 50-item subset. The full annotation schema is in Table~\ref{tab:charm100-full-criteria}; dataset details are in Appendix~\ref{appendix:charm100-dataset}.

\begin{figure}[t]
    \centering
    \includegraphics[width=0.7\linewidth]{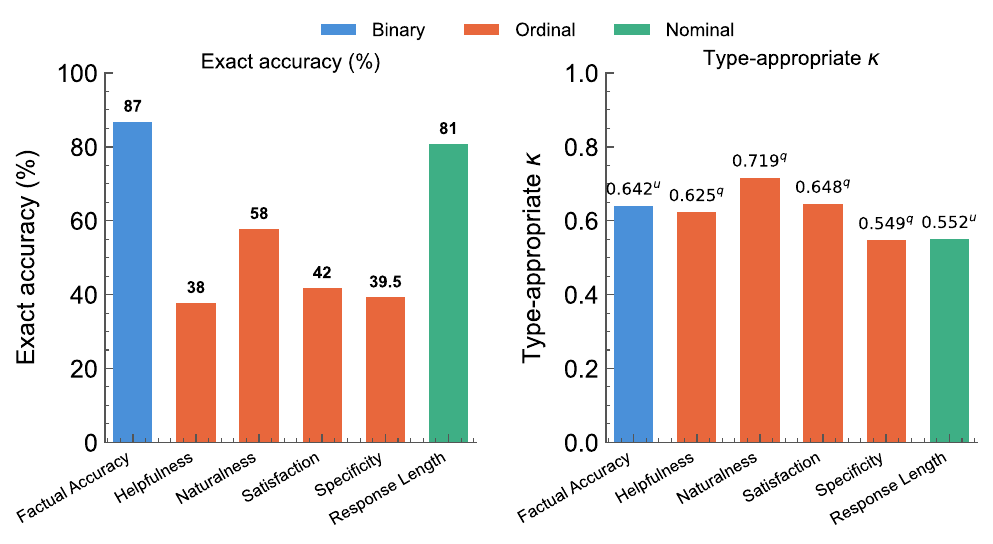}
    \caption{CHARM-100 per-criterion exact accuracy (left) and type-appropriate $\kappa$ (right) under Gemini-3-Flash, grouped by criterion type. The right panel uses unweighted $\kappa$ for binary and nominal criteria and quadratic-weighted $\kappa$ for ordinal criteria, so it is not a common-metric ranking of types. Factual accuracy has the highest exact accuracy; naturalness has the highest displayed $\kappa$.}
    \label{fig:charm100-kappa}
\end{figure}

Table~\ref{tab:charm100-per-criterion} (Appendix~\ref{appendix:charm100-results}) reports per-criterion results using Gemini-3-Flash as the judge\footnote{For the CHARM-100 evaluation results, $\kappa$ compares judge predictions with reference labels. Ordinal criteria use quadratic-weighted $\kappa$; binary and nominal criteria use unweighted $\kappa$. A separate second-annotation audit reports mean agreement with the reference labels of $\kappa = 0.69$ on 50 items (Appendix~\ref{appendix:charm100-iaa}).}. Under this configuration, factual accuracy, the sole binary criterion, has the highest exact accuracy (87.0\%; unweighted $\kappa = 0.642$). The four ordinal criteria show lower exact agreement (38--58\%) despite high adjacent accuracy (85--93\%): the judge is usually within one step of ground truth but clusters toward scale extremes. Naturalness has the highest reported type-appropriate $\kappa$ (0.719, quadratic-weighted), and the ordinal criteria show strong rank correlations (Spearman 0.698--0.786). The sole nominal criterion, response length, achieves 81.0\% exact accuracy but detects brevity (0.70 recall) far better than verbosity (0.14 recall). Because CHARM-100 contains only one binary and one nominal criterion and uses type-specific $\kappa$ weighting, these results describe criterion-specific error profiles rather than a general reliability hierarchy. Figure~\ref{fig:charm100-kappa} visualizes these patterns.

In aggregate, the model exhibits a positive bias (+0.170), consistent with documented LLM judge leniency~\citep{zheng2023judging, ye2024flask}, though score-level ranking remains strong (Spearman = 0.810). Full confusion matrices are in Appendix~\ref{appendix:charm100-results}.

\subsection{Configuration sensitivity}
\label{sec:ablation}

We conduct an exploratory configuration-sensitivity analysis on CHARM-100 across three model families: Gemini-3-Flash, GPT-5.4-nano, and LLaMA-3.1-8B. The Default configuration uses option shuffling, 3-shot verdict-balanced calibration, and the SKIP abstention strategy. The $-$Shuffle and $-$Few-shot rows remove one component; the ensemble rows add repeated judges, Bare removes multiple components, and Cross-family combines the three Default runs. All configurations use the same 80-item test set (20 items reserved for few-shot training, stratified split, seed 42). These runs are independent stochastic evaluations from the primary CHARM result above. Because each configuration was run once, differences are descriptive and include model-output variability.

\begin{table*}[t]
\centering
\small
\setlength{\tabcolsep}{1pt}
\begin{tabular}{@{}lcccccccccccc@{}}
\toprule
& \multicolumn{4}{c}{\textbf{Gemini-3-Flash}} & \multicolumn{4}{c}{\textbf{GPT-5.4-nano}} & \multicolumn{4}{c}{\textbf{LLaMA-3.1-8B}} \\
\textbf{Configuration} & Exact & $\kappa$ & $\rho$ & RMSE & Exact & $\kappa$ & $\rho$ & RMSE & Exact & $\kappa$ & $\rho$ & RMSE \\
\midrule
Default & 60.4\% & 0.679 & 0.800 & 0.213 & 51.0\% & 0.462 & 0.562 & 0.242 & 19.4\% & -0.001 & 0.022 & 0.605 \\
$-$Shuffle & 60.0\% & 0.660 & 0.828 & 0.217 & 46.7\% & 0.425 & 0.539 & 0.262 & 24.0\% & 0.033 & 0.200 & 0.512 \\
$-$Few-shot & 59.8\% & 0.657 & 0.825 & 0.231 & 48.3\% & 0.434 & 0.614 & 0.248 & 21.5\% & 0.087 & 0.071 & 0.560 \\
\midrule
+Ens($k{=}3$, maj.) & 61.3\% & 0.673 & 0.813 & 0.217 & 46.9\% & 0.437 & 0.493 & 0.272 & 21.5\% & 0.018 & 0.125 & 0.575 \\
+Ens($k{=}5$, maj.) & 61.5\% & 0.678 & 0.812 & 0.216 & 49.0\% & 0.456 & 0.533 & 0.263 & 24.0\% & 0.044 & 0.310 & 0.521 \\
+Ens($k{=}3$, mean) & 61.3\% & 0.670 & 0.828 & 0.215 & 47.9\% & 0.451 & 0.504 & 0.264 & 24.8\% & 0.043 & 0.135 & 0.509 \\
\midrule
Bare (no mitigations) & 57.3\% & 0.629 & 0.767 & 0.225 & 48.3\% & 0.428 & 0.644 & 0.261 & 19.2\% & -0.038 & -0.053 & 0.594 \\
Cross-family ens. & 57.1\% & 0.626 & 0.769 & 0.200 & \multicolumn{4}{c}{---} & \multicolumn{4}{c}{---} \\
\bottomrule
\end{tabular}
\caption{Configuration sensitivity on CHARM-100 (80 test items, 6 criteria). Exact is pooled exact-label accuracy over all 480 item--criterion decisions, including N/A as a label. $\kappa$ is the unweighted mean of six per-criterion Cohen's $\kappa$ values (quadratic-weighted for ordinal criteria; N/A pairs excluded); $\rho$ and RMSE are item-score metrics. The cross-family ensemble majority-votes across the three Default runs.}
\label{tab:ablation}
\end{table*}

Judge family is the largest source of variation: under Default, pooled exact accuracy/$\kappa$ is 60.4\%/0.679 for Gemini, 51.0\%/0.462 for GPT, and 19.4\%/$-0.001$ for LLaMA. Configuration effects are not universal. Removing few-shot examples changes pooled exact accuracy by $-0.6$, $-2.7$, and $+2.1$pp for Gemini, GPT, and LLaMA, respectively; removing shuffling changes it by $-0.4$, $-4.3$, and $+4.6$pp. Same-model ensembles raise Gemini exact accuracy by at most 1.1pp, lower GPT accuracy by 2.0--4.1pp, and raise LLaMA accuracy by 2.1--5.4pp while leaving its $\kappa$ near zero. Relative to Bare, Default improves pooled exact accuracy by 3.1pp (Gemini), 2.7pp (GPT), and 0.2pp (LLaMA). The cross-family panel has lower exact accuracy and $\kappa$ than Gemini alone (57.1\%/0.626 vs.\ 60.4\%/0.679), although its score RMSE is slightly lower (0.200 vs.\ 0.213), and it costs more. These single-run comparisons support model- and criterion-dependent configuration choices, not a universally beneficial mitigation stack; Appendix~\ref{appendix:ablation} provides the per-type breakdown.

\section{Application: agent skill improvement}
\label{sec:skill-improvement}

Per-criterion measurements can serve as an optimization signal to improve an LLM agent's \emph{skill}---a structured collection of instructions that enable an agent to execute a specific task reliably\footnote{See \url{https://agentskills.io/what-are-skills}}.

\begin{figure*}[h]
    \centering
    \begin{subfigure}[b]{0.60\linewidth}
        \centering
        \includegraphics[width=\linewidth]{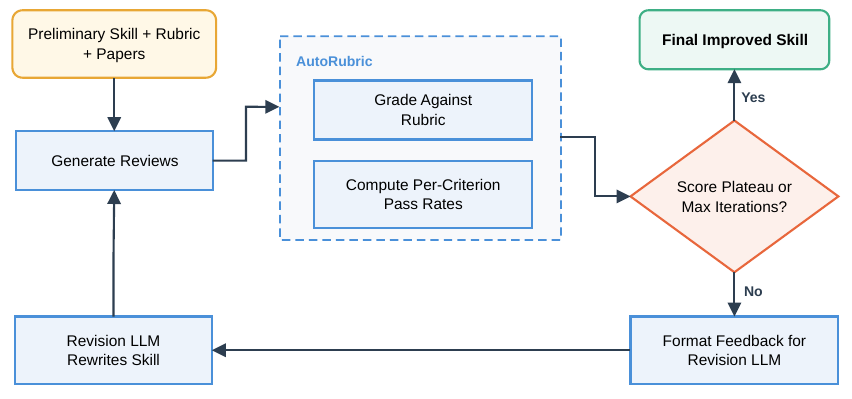}
        \caption{Rubric-guided skill improvement loop}
    \end{subfigure}
    \hfill
    \begin{subfigure}[b]{0.35\linewidth}
        \centering
        \includegraphics[width=\linewidth]{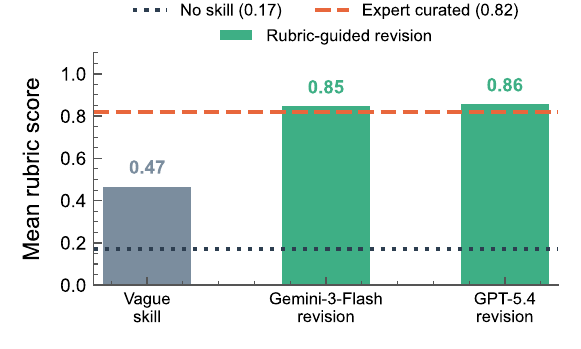}
        \caption{Improvement results}
    \end{subfigure}
    \caption{Agent skill improvement via rubric-guided feedback. (a) The iterative loop uses \texttt{Autorubric} grading as the optimization signal. (b) Starting from a vague skill (0.47), a single rubric-guided revision produces point estimates of 0.85 (Gemini-3-Flash revision) and 0.86 (GPT-5.4 revision), compared with 0.82 for the expert-curated baseline (dashed); all three 95\% bootstrap CIs overlap. Gemini-3-Flash served as the rubric grading model in both revision conditions, but the conditions independently sampled the feedback used for revision and the reviews used for evaluation. Their similar point estimates therefore do not isolate revision-model self-preference or establish equivalence or superiority. Confidence intervals are in Table~\ref{tab:skill-cross-judge} (Appendix~\ref{appendix:skill-improvement}).}
    \label{fig:skill-improvement}
\end{figure*}

As a demonstration, we consider the task of peer reviewing 10 scientific papers using Llama 3.1 8B, graded against a 10-criterion binary rubric covering outcome quality, style, efficiency, and a factual misrepresentation penalty. We establish performance boundaries: no skill (score: 0.17), vague one-line skill (0.47), and expert-curated skill (0.82). In a setup similar to prompt induction~\citep{agrawal2026gepa, yuksekgonul2024textgrad}, an improvement loop grades reviews against the rubric, formats failing criteria into feedback, and sends the feedback to a revision LLM. Starting from the vague skill, a single revision raises the score to 0.85. This slightly exceeds the expert-curated point estimate of 0.82, but their bootstrap CIs overlap and no revised-versus-curated superiority test was conducted (Figure~\ref{fig:skill-improvement}). As a descriptive robustness check, we repeat the full pipeline with GPT-5.4 as the revision model (score: 0.86, overlapping CIs). The two revision arms independently sample the initial reviews and Gemini grading feedback supplied to the reviser, as well as the post-revision reviews and grades, so their similar point estimates do not isolate revision-model self-preference. With GPT-5.4-mini as a second grader on freshly generated reviews, the vague-to-revised increase persists ($0.58 \to 0.74$), while the revised point estimate remains below the expert-curated point estimate (0.78). Across these end-to-end robustness runs, improvement over the vague skill appears under both graders, but the design does not establish superiority over expert curation or isolate revision-model or grading-model effects (Table~\ref{tab:skill-cross-judge} in Appendix~\ref{appendix:skill-improvement}).

\section{Application: reinforcement learning with rubric-based rewards}
\label{sec:rlrr}

In contrast to per-criterion explanations, the per-criterion scores can serve as a continuous reward signal for reinforcement learning.\footnote{Unlike RLVR where rewards derive from programmatic checks, rubric-based rewards are interpretable but not deterministically verifiable.} We demonstrate this on the \texttt{complex\_if\_single\_turn\_v5} subset of AdvancedIF~\citep{he2025advancedif}: 402 single-turn prompts with 7.44 expert-curated binary criteria on average, split 80/20 into training and validation\footnote{This subset is the most uniform of the three AdvancedIF categories (rubric counts 4--9, mean 7.4), reducing confounds from rubric-count variance.}. The policy model is \texttt{Qwen/Qwen3-4B-Instruct}, trained with LoRA via the Tinker RL framework\footnote{\url{https://thinkingmachines.ai/tinker/}} (rank 32, LR $5 \times 10^{-4}$, group size 8, batch size 32). The reward is the \texttt{Autorubric} normalized score from \texttt{gemini-3-flash-preview}. Total judge cost: approximately \$12 for 50 training steps (5 epochs).

Mean training rubric score rises from 0.774 (epoch 1) to 0.825 (epoch 3), then declines in epochs 4--5. We evaluate the 81-prompt validation split every five steps and report step~25, the highest-scoring of the ten monitored checkpoints (Figure~\ref{fig:rlrr-eval-progression}). On this same validation split used for checkpoint selection, mean rubric score increases from 0.756 to 0.795 (+0.039; nominal paired two-sided Wilcoxon signed-rank $p = 0.032$; Cohen's $d = 0.26$, 95\% bootstrap CI $[0.04, 0.51]$). This $p$-value is unadjusted for checkpoint selection and is descriptive rather than an independent test. Responses with rubric score 1.0 increase from 21/81 to 30/81 (+43\%), and mean validation response length falls 33\% (1{,}097 to 733 tokens). Across all 50 training batches, response length is non-monotonic and has no detectable linear trend (+1.9 tokens/step, $r = 0.08$, $p = 0.568$); KL divergence remains below 0.003 and format compliance remains stable (Appendix~\ref{appendix:rlrr}).

To probe judge dependence, we independently regenerate responses from the base and step-25 models and grade them with GPT-5.4-mini. This check yields a smaller positive difference (+0.023; nominal paired two-sided Wilcoxon $p = 0.243$; Cohen's $d = 0.13$, 95\% bootstrap CI $[-0.09, 0.35]$). Because it changes both sampled responses and judge, it is directionally consistent but inconclusive and does not isolate judge coupling (Appendix~\ref{appendix:rlrr}).

\begin{figure*}[h]
    \centering
    \includegraphics[width=\linewidth]{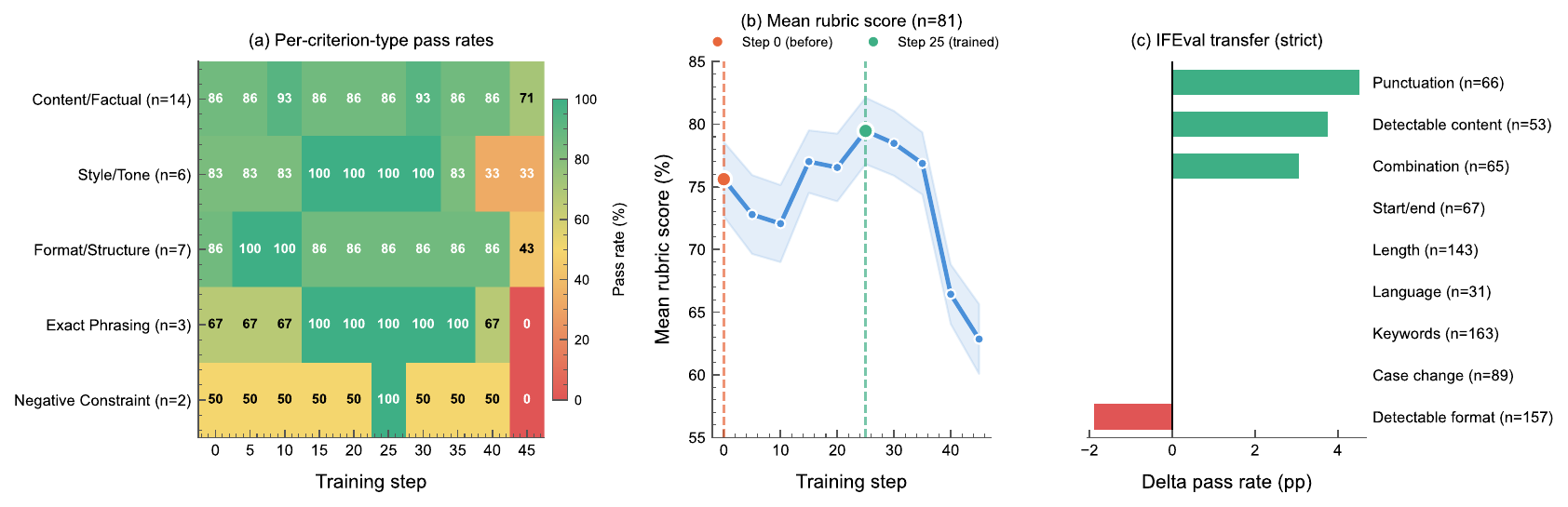}
    \caption{Reinforcement learning with \texttt{Autorubric} rewards. (a) Exploratory keyword-derived category pass rates from four rollout groups per checkpoint. (b) Mean score on 81 validation prompts; step~25 is the best monitored checkpoint (0.795 vs.\ 0.756). (c) IFEval~\citep{zhou2023ifeval} strict pass-rate changes; eight of nine instruction types show non-negative transfer. Qualifications and diagnostics are in Appendix~\ref{appendix:rlrr}.}
    \label{fig:rlrr}
\end{figure*}

Figure~\ref{fig:rlrr}(a--b) reveals which criterion groups benefit in this run. Style, phrasing, and format criteria show the largest gains through the selected checkpoint, while some content and factual criteria remain unresolved. On IFEval~\citep{zhou2023ifeval} (Figure~\ref{fig:rlrr}(c)), eight of nine instruction types show non-negative out-of-domain transfer, with the largest gains in punctuation (+4.5pp) and detectable content (+3.8pp). The transfer gains are directionally positive but not statistically significant (McNemar $p = 0.28$).

\section{Additional Related Works}
\label{sec:foundations}

Rubric-based evaluation draws on educational measurement and psychometrics. Three rubric-authoring principles motivate \texttt{Autorubric}'s criterion-level structure: \emph{unidimensionality}, \emph{behavioral anchors}, and \emph{construct alignment}~\citep{mckeown2018heqco}. These remain author responsibilities; the framework does not enforce or certify them. From psychometrics~\citep{comer2009rubrics}, it provides Cohen's $\kappa$, quadratic weighted $\kappa$, rank correlations, and Earth Mover's Distance~\citep{he2025llmjudge}. LLM-specific concerns---position bias, prompt sensitivity, self-preference bias~\citep{llm-judge-self-bias-panickssery, llm-judge-self-bias-watoka}---have no educational precedent and require the mitigations in Section~\ref{sec:failure-modes}.

Several frameworks address bias and calibration in LLM evaluation. RULERS~\citep{hong2026rulers} learns a distributional mapping from 200+ labeled examples per dataset, operating in the supervised regime with fixed rubrics. \texttt{Autorubric} targets a different setting: user-defined analytic rubrics without task-specific calibration data, relying on generation-time mitigations. In the pairwise paradigm, \citet{zhou2024zepo} show that optimizing toward balanced decision rates improves alignment; the analogous concern in pointwise evaluation is option position bias, addressed by \texttt{Autorubric}'s automatic shuffling. At the classification level, Batch Calibration~\citep{zhou2023batch} removes contextual bias via batch statistics over logit vectors; the principle that prompt-induced priors distort judgments motivates \texttt{Autorubric}'s verdict-balanced few-shot design. \citet{xia2025calibn} produce calibrated confidence from inter-model agreement for factoid QA; \texttt{Autorubric}'s ensemble mode surfaces a related quantity as a reliability diagnostic rather than a confidence score.

\section{Discussion and conclusion}
\label{sec:discussion}

\textbf{Criterion format and construct shape error profiles.} In the CHARM-100 Gemini-3-Flash evaluation, factual accuracy has the highest exact accuracy (87\%), whereas the ordinal criteria show lower exact accuracy (38--58\%) but high adjacent accuracy (85--93\%); naturalness has the highest reported type-appropriate $\kappa$ (0.719). With only one binary and one nominal criterion, criterion identity is confounded with type, and the type-specific $\kappa$ variants are not directly comparable as a reliability hierarchy. The practical implication is to match criterion format to the construct---binary for defensible thresholds and ordinal for meaningful gradations---and report type-appropriate metrics rather than treating one format as universally preferable.

\textbf{Per-criterion decomposition as measurement and optimization.} The central finding across Sections~\ref{sec:charm100}--\ref{sec:rlrr} is that per-criterion analytic rubrics serve two purposes with a single representation. For measurement, per-criterion scores diagnose where judges agree and disagree---e.g., CHARM-100 reveals 0.70 recall for brevity but 0.14 for verbosity, a distinction invisible in holistic scores. The same scores provide actionable optimization feedback: the skill improvement loop uses per-criterion pass rates to raise a peer review agent's score from 0.47 to 0.85, and per-criterion RL rewards yield a positive validation-selected difference on AdvancedIF with directionally positive out-of-domain transfer to IFEval. Holistic scores can support scalar black-box optimization, but they do not directly localize which requirements failed.

\texttt{Autorubric} provides a common implementation, but consistency cannot establish construct validity.

This paper represents a snapshot of \texttt{Autorubric} as it existed at the time of the COLM peer-review submission in March 2026. By the camera-ready submission in August 2026, \texttt{Autorubric} had already gained substantial new capabilities beyond that snapshot, and it continues to evolve. For example, users can now integrate \texttt{Autorubric} directly into agent harnesses through a skill without needing to write any code, assess rubric quality, induce rubrics, and more. We strongly encourage readers to consult the extensive documentation, cookbook recipes, and latest MIT-licensed source code at \href{https://autorubric.org/}{https://autorubric.org/}.

\section*{Limitations}
\label{sec:limitations}
We group limitations into two categories. \emph{Evaluation limitations}: (1) LLM judges exhibit ordinal scale-extreme clustering, producing misleadingly low exact accuracy on graded criteria; coarser scales or reporting adjacent accuracy as the primary metric partially mitigate this; batch calibration or ordinal regression methods may further reduce scale clustering, but we leave this to future work. (2) Some constructs resist binary categorization; multi-choice criteria address this at the cost of more careful rubric design. (3) Gemini-3-Flash at 5--6$\times$ lower cost produced the same aggregate ranking as Sonnet-4.5 on ResearcherBench (Section~\ref{sec:researcherbench}), although bootstrap CIs overlap for the top two systems; inexpensive judges may not suffice for criterion-level analysis where absolute calibration matters. (4) Ensemble judging with $k$ judges requires $k\times$ LLM calls; the configuration-sensitivity analysis shows small and model-dependent changes for same-model ensembles, limiting the evidence for systematic error correction. Adaptive ensembling---using multiple judges only for low-confidence items---could reduce cost while targeting the cases most likely to benefit. (5) Verbosity bias~\citep{dubois2024lengthcontrolled} is not explicitly mitigated in the current framework; length-controlled evaluation or post-hoc calibration methods are potential directions. (6) Each CHARM configuration was run once, so comparisons include stochastic model-output variation and do not identify causal mitigation effects. Judge-family differences are larger than the within-model ranges for Gemini and GPT, but the framework's mitigations cannot compensate for the weakest judge tested. (7) All experiments and default prompts are English-only. The framework is language-agnostic in design, but criterion evaluation prompts and few-shot exemplars may require adaptation for other languages; we do not test this. \emph{Optimization limitations}: (8) The optimized skill is tightly coupled to both rubric and model; a different rubric may reward different tradeoffs, and factual-grounding failures were not eliminated in this setup. (9) The AdvancedIF validation split is used both to select step~25 and estimate its gain; the nominal $p$-value is unadjusted for selection, and no independent test split was evaluated. The independent-judge check regenerates responses, so it cannot isolate judge effects from generation variability. (10) The framework assumes rubrics are well-designed. Rubric quality assessment---validating that criteria measure what they claim to measure---remains an open problem.

\section*{Acknowledgments}
This research was developed with funding from the Defense Advanced Research Projects Agency's (DARPA) SciFy program (Agreement No. HR00112520300) and is based upon work supported in part by the Office of the Director of National Intelligence (ODNI), Intelligence Advanced Research Projects Activity (IARPA), via 56000026C0019 (the BENGAL program). The views and conclusions contained herein are those of the authors and should not be interpreted as necessarily representing the official policies, either expressed or implied, of DARPA, ODNI, IARPA, the Department of Defense, or the U.S. Government. The U.S. Government is authorized to reproduce and distribute reprints for governmental purposes notwithstanding any copyright annotation therein.

\section*{Generative AI Use Disclosure}
\label{sec:generative-ai-use-disclosure}

The authors used LLMs both as experimental components and in preparing this manuscript. As described in the corresponding methods, LLMs served as evaluation judges, revision models, and a reinforcement-learning reward model; a language model also generated the CHARM-100 conversations and reference labels, followed by a recorded second annotation pass on 50 items. For manuscript preparation, the authors used Gemini 3 Pro Preview and Claude Sonnet 4.5 to organize references, proofread drafts, make plots, and provide light rewrites. The authors reviewed the resulting prose, figures, data artifacts, and citations and take full responsibility for the paper's contents.

\section*{Ethics Statement}
\label{sec:ethics}
A consolidated framework like \texttt{Autorubric} creates ethical tradeoffs. By lowering barriers to deployment, we enable evaluation at scales previously infeasible---with benefits (practitioners gain access to bias mitigations) but also risks (reduced friction may prompt deployment without careful consideration of appropriateness). Clean APIs can produce overconfidence: a score between 0 and 1 obscures uncertainty and systematic error. Those deploying automated evaluation benefit from cost savings while those being evaluated bear the risk of biased judgments. \texttt{Autorubric} surfaces uncertainty indicators (\texttt{mean\_agreement}, per-criterion vote distributions) alongside scores, and we recommend treating low-agreement cases as requiring human review. The analytic rubric design produces per-criterion explanations suitable for disclosure to evaluated parties. We caution against interpreting this framework as endorsement of automated evaluation in high-stakes human assessment; organizations should validate against human judgments for their specific populations and maintain oversight proportional to decision stakes.

\bibliography{paper}

\begin{thebibliography}{36}
\providecommand{\natexlab}[1]{#1}
\providecommand{\url}[1]{\texttt{#1}}
\expandafter\ifx\csname urlstyle\endcsname\relax
  \providecommand{\doi}[1]{doi: #1}\else
  \providecommand{\doi}{doi: \begingroup \urlstyle{rm}\Url}\fi

\bibitem[Agrawal et~al.(2026)Agrawal, Tan, Soylu, Ziems, Khare, Opsahl-Ong,
  Singhvi, Shandilya, Ryan, Jiang, Potts, Sen, Dimakis, Stoica, Klein, Zaharia,
  and Khattab]{agrawal2026gepa}
Lakshya~A Agrawal, Shangyin Tan, Dilara Soylu, Noah Ziems, Rishi Khare, Krista
  Opsahl-Ong, Arnav Singhvi, Herumb Shandilya, Michael~J Ryan, Meng Jiang,
  Christopher Potts, Koushik Sen, Alexandros~G. Dimakis, Ion Stoica, Dan Klein,
  Matei Zaharia, and Omar Khattab.
\newblock {GEPA}: Reflective prompt evolution can outperform reinforcement
  learning, 2026.
\newblock URL \url{https://arxiv.org/abs/2507.19457}.

\bibitem[Brookhart(2018)]{brookhart2018criteria}
Susan~M. Brookhart.
\newblock Appropriate criteria: Key to effective rubrics.
\newblock \emph{Frontiers in Education}, 3:\penalty0 22, 2018.
\newblock \doi{10.3389/feduc.2018.00022}.

\bibitem[Comer(2009)]{comer2009rubrics}
Keith Comer.
\newblock Developing valid and reliable rubrics for writing assessment:
  Research and practice.
\newblock Technical report, Ako Aotearoa, Wellington, New Zealand, 2009.
\newblock URL \url{https://mro.massey.ac.nz/handle/10179/10631}.

\bibitem[Dubois et~al.(2024)Dubois, Galambosi, Liang, and
  Hashimoto]{dubois2024lengthcontrolled}
Yann Dubois, Bal{\'a}zs Galambosi, Percy Liang, and Tatsunori~B. Hashimoto.
\newblock Length-controlled {AlpacaEval}: A simple way to debias automatic
  evaluators.
\newblock In \emph{First Conference on Language Modeling}, 2024.
\newblock URL \url{https://openreview.net/forum?id=CybBmzWBX0}.

\bibitem[Feuer et~al.(2025)Feuer, Tseng, Lathe, Elachqar, and
  Dickerson]{feuer2025judgment}
Benjamin Feuer, Chiung-Yi Tseng, Astitwa~Sarthak Lathe, Oussama Elachqar, and
  John~P. Dickerson.
\newblock When judgment becomes noise: How design failures in {LLM} judge
  benchmarks silently undermine validity.
\newblock \emph{arXiv preprint}, 2025.
\newblock \doi{10.48550/arXiv.2509.20293}.

\bibitem[Gu et~al.(2026)Gu, Jiang, Shi, Tan, Zhai, Xu, Li, Shen, Ma, Liu, Wang,
  Zhang, Lin, Zhang, Ni, Gao, Wang, and Guo]{gu2024survey}
Jiawei Gu, Xuhui Jiang, Zhichao Shi, Hexiang Tan, Xuehao Zhai, Chengjin Xu, Wei
  Li, Yinghan Shen, Shengjie Ma, Honghao Liu, Saizhuo Wang, Kun Zhang, Zhouchi
  Lin, Bowen Zhang, Lionel Ni, Wen Gao, Yuanzhuo Wang, and Jian Guo.
\newblock A survey on {LLM}-as-a-judge.
\newblock \emph{The Innovation}, 7\penalty0 (6):\penalty0 101253, 2026.
\newblock \doi{10.1016/j.xinn.2025.101253}.

\bibitem[Gunjal et~al.(2025)Gunjal, Wang, Lau, Nath, He, Liu, and
  Hendryx]{gunjal2025rubrics}
Anisha Gunjal, Anthony Wang, Elaine Lau, Vaskar Nath, Yunzhong He, Bing Liu,
  and Sean Hendryx.
\newblock Rubrics as rewards: Reinforcement learning beyond verifiable domains.
\newblock \emph{arXiv preprint}, 2025.
\newblock \doi{10.48550/arXiv.2507.17746}.

\bibitem[Haldar \& Hockenmaier(2025)Haldar and
  Hockenmaier]{haldar2025ratingroulette}
Rajarshi Haldar and Julia Hockenmaier.
\newblock Rating roulette: Self-inconsistency in {LLM}-as-a-judge frameworks.
\newblock In \emph{Findings of the Association for Computational Linguistics:
  EMNLP 2025}, pp.\  24986--25004. Association for Computational Linguistics,
  2025.
\newblock \doi{10.18653/v1/2025.findings-emnlp.1361}.

\bibitem[Hashemi et~al.(2024)Hashemi, Eisner, Rosset, Van~Durme, and
  Kedzie]{hashemi2024llmrubric}
Helia Hashemi, Jason Eisner, Corby Rosset, Benjamin Van~Durme, and Chris
  Kedzie.
\newblock {LLM}-rubric: A multidimensional, calibrated approach to automated
  evaluation of natural language texts.
\newblock In \emph{Proceedings of the 62nd Annual Meeting of the Association
  for Computational Linguistics (Volume 1: Long Papers)}, pp.\  13806--13834.
  Association for Computational Linguistics, 2024.
\newblock \doi{10.18653/v1/2024.acl-long.745}.

\bibitem[He et~al.(2025{\natexlab{a}})He, Shi, Zhuo, Treude, Sun, Xing, Du, and
  Lo]{he2025llmjudge}
Junda He, Jieke Shi, Terry~Yue Zhuo, Christoph Treude, Jiamou Sun, Zhenchang
  Xing, Xiaoning Du, and David Lo.
\newblock {LLM}-as-a-judge for software engineering: Literature review, vision,
  and the road ahead.
\newblock \emph{arXiv preprint}, 2025{\natexlab{a}}.
\newblock \doi{10.48550/arXiv.2510.24367}.

\bibitem[He et~al.(2025{\natexlab{b}})He, Li, Zhang, Li, Mandyam, Khosla,
  Xiong, Wang, Peng, Li, Bi, Patil, Qi, Feng, Katz-Samuels, Pang, Gonugondla,
  Lang, Yu, Qian, Fazel-Zarandi, Yu, Benhalloum, Awadalla, and
  Faruqui]{he2025advancedif}
Yun He, Wenzhe Li, Hejia Zhang, Songlin Li, Karishma Mandyam, Sopan Khosla,
  Yuanhao Xiong, Nanshu Wang, Xiaoliang Peng, Beibin Li, Shengjie Bi,
  Shishir~G. Patil, Qi~Qi, Shengyu Feng, Julian Katz-Samuels, Richard~Yuanzhe
  Pang, Sujan Gonugondla, Hunter Lang, Yue Yu, Yundi Qian, Maryam
  Fazel-Zarandi, Licheng Yu, Amine Benhalloum, Hany Awadalla, and Manaal
  Faruqui.
\newblock {AdvancedIF}: Rubric-based benchmarking and reinforcement learning
  for advancing {LLM} instruction following, 2025{\natexlab{b}}.
\newblock URL \url{https://arxiv.org/abs/2511.10507}.

\bibitem[Hong et~al.(2026)Hong, Yao, Shen, Xu, Wei, and Dong]{hong2026rulers}
Yihan Hong, Huaiyuan Yao, Bolin Shen, Wanpeng Xu, Hua Wei, and Yushun Dong.
\newblock From rubrics to reliable scores: Evidence-grounded text evaluation
  with {LLM} judges.
\newblock \emph{arXiv preprint}, 2026.
\newblock \doi{10.48550/arXiv.2601.08654}.

\bibitem[Kim et~al.(2024)Kim, Suk, Longpre, Lin, Shin, Welleck, Neubig, Lee,
  Lee, and Seo]{kim2024prometheus}
Seungone Kim, Juyoung Suk, Shayne Longpre, Bill~Yuchen Lin, Jamin Shin, Sean
  Welleck, Graham Neubig, Moontae Lee, Kyungjae Lee, and Minjoon Seo.
\newblock Prometheus 2: An open source language model specialized in evaluating
  other language models.
\newblock In \emph{Proceedings of the 2024 Conference on Empirical Methods in
  Natural Language Processing}, pp.\  4334--4353, 2024.
\newblock \doi{10.18653/v1/2024.emnlp-main.248}.

\bibitem[Lee et~al.(2025)Lee, Kim, Kim, Cho, Kang, Kang, and
  Kim]{lee2025checkeval}
Yukyung Lee, JoongHoon Kim, Jaehee Kim, Hyowon Cho, Jaewook Kang, Pilsung Kang,
  and Najoung Kim.
\newblock {CheckEval}: A reliable {LLM}-as-a-judge framework for evaluating
  text generation using checklists.
\newblock In \emph{Proceedings of the 2025 Conference on Empirical Methods in
  Natural Language Processing}, pp.\  15782--15809, 2025.
\newblock \doi{10.18653/v1/2025.emnlp-main.796}.

\bibitem[Li et~al.(2025)Li, Chiang, Frick, Dunlap, Wu, Zhu, Gonzalez, and
  Stoica]{li2025arenahard}
Tianle Li, Wei-Lin Chiang, Evan Frick, Lisa Dunlap, Tianhao Wu, Banghua Zhu,
  Joseph~E. Gonzalez, and Ion Stoica.
\newblock From crowdsourced data to high-quality benchmarks: {Arena-Hard} and
  {BenchBuilder} pipeline.
\newblock In \emph{Proceedings of the 42nd International Conference on Machine
  Learning}, volume 267 of \emph{Proceedings of Machine Learning Research},
  pp.\  34209--34231. PMLR, 2025.
\newblock URL \url{https://proceedings.mlr.press/v267/li25h.html}.

\bibitem[Lin et~al.(2025)Lin, Deng, Chandu, Brahman, Ravichander, Pyatkin,
  Dziri, Bras, and Choi]{lin2024wildbench}
Bill~Yuchen Lin, Yuntian Deng, Khyathi Chandu, Faeze Brahman, Abhilasha
  Ravichander, Valentina Pyatkin, Nouha Dziri, Ronan~Le Bras, and Yejin Choi.
\newblock {WildBench}: Benchmarking {LLMs} with challenging tasks from real
  users in the wild.
\newblock In \emph{International Conference on Learning Representations
  (ICLR)}, pp.\  47852--47870, 2025.
\newblock \doi{10.48550/arXiv.2406.04770}.

\bibitem[Liu et~al.(2023)Liu, Iter, Xu, Wang, Xu, and Zhu]{liu2023geval}
Yang Liu, Dan Iter, Yichong Xu, Shuohang Wang, Ruochen Xu, and Chenguang Zhu.
\newblock {G-Eval}: {NLG} evaluation using {GPT-4} with better human alignment.
\newblock In \emph{Proceedings of the 2023 Conference on Empirical Methods in
  Natural Language Processing}, pp.\  2511--2522. Association for Computational
  Linguistics, 2023.
\newblock \doi{10.18653/v1/2023.emnlp-main.153}.

\bibitem[McKeown \& Lenarcic~Biss(2018)McKeown and
  Lenarcic~Biss]{mckeown2018heqco}
Jessica McKeown and Diane Lenarcic~Biss.
\newblock {HEQCO}'s guide to developing valid and reliable rubrics.
\newblock Technical report, Higher Education Quality Council of Ontario,
  Toronto, ON, 2018.
\newblock URL
  \url{https://heqco.ca/wp-content/uploads/2020/06/Formatted_Rubric-Guide_FINAL.pdf}.

\bibitem[Mrangu(2022)]{mrangu2022rubric}
Leah Mrangu.
\newblock Rubric as assessment tool for lecturers and students in higher
  education institution.
\newblock \emph{Acta Pedagogia Asiana}, 1\penalty0 (1):\penalty0 26--33, 2022.
\newblock \doi{10.53623/apga.v1i1.98}.

\bibitem[Panickssery et~al.(2024)Panickssery, Bowman, and
  Feng]{llm-judge-self-bias-panickssery}
Arjun Panickssery, Samuel~R. Bowman, and Shi Feng.
\newblock {LLM} evaluators recognize and favor their own generations.
\newblock In \emph{Advances in Neural Information Processing Systems
  (NeurIPS)}, pp.\  68772--68802, 2024.
\newblock \doi{10.52202/079017-2197}.

\bibitem[Sonkar et~al.(2024)Sonkar, Ni, Lu, Kincaid, Hutchinson, and
  Baraniuk]{sonkar2024automatedlonganswergrading}
Shashank Sonkar, Kangqi Ni, Lesa~Tran Lu, Kristi Kincaid, John~S. Hutchinson,
  and Richard~G. Baraniuk.
\newblock Automated long answer grading with {RiceChem} dataset.
\newblock In \emph{Proceedings of the 25th International Conference on
  Artificial Intelligence in Education (AIED 2024)}, volume 14829 of
  \emph{Lecture Notes in Computer Science}, pp.\  163--176. Springer, 2024.
\newblock \doi{10.1007/978-3-031-64302-6_12}.
\newblock URL \url{https://arxiv.org/abs/2404.14316}.

\bibitem[Turpin et~al.(2023)Turpin, Michael, Perez, and
  Bowman]{turpin2023language}
Miles Turpin, Julian Michael, Ethan Perez, and Samuel~R. Bowman.
\newblock Language models don't always say what they think: Unfaithful
  explanations in chain-of-thought prompting.
\newblock In \emph{Advances in Neural Information Processing Systems
  (NeurIPS)}, 2023.
\newblock URL
  \url{https://proceedings.neurips.cc/paper_files/paper/2023/hash/ed3fea9033a80fea1376299fa7863f4a-Abstract-Conference.html}.

\bibitem[Verga et~al.(2024)Verga, Hofstatter, Althammer, Su, Piktus,
  Arkhangorodsky, Xu, White, and Lewis]{verga2024replacing}
Pat Verga, Sebastian Hofstatter, Sophia Althammer, Yixuan Su, Aleksandra
  Piktus, Arkady Arkhangorodsky, Minjie Xu, Naomi White, and Patrick Lewis.
\newblock Replacing judges with juries: Evaluating {LLM} generations with a
  panel of diverse models.
\newblock \emph{arXiv preprint}, 2024.
\newblock \doi{10.48550/arXiv.2404.18796}.

\bibitem[Wang et~al.(2024{\natexlab{a}})Wang, Li, Chen, Cai, Zhu, Lin, Cao,
  Kong, Liu, Liu, and Sui]{wang2023fairevaluators}
Peiyi Wang, Lei Li, Liang Chen, Zefan Cai, Dawei Zhu, Binghuai Lin, Yunbo Cao,
  Lingpeng Kong, Qi~Liu, Tianyu Liu, and Zhifang Sui.
\newblock Large language models are not fair evaluators.
\newblock In \emph{Proceedings of the 62nd Annual Meeting of the Association
  for Computational Linguistics (Volume 1: Long Papers)}, pp.\  9440--9450.
  Association for Computational Linguistics, 2024{\natexlab{a}}.
\newblock \doi{10.18653/v1/2024.acl-long.511}.

\bibitem[Wang et~al.(2024{\natexlab{b}})Wang, Dong, Delalleau, Zeng, Shen,
  Egert, Zhang, Sreedhar, and Kuchaiev]{wang2024helpsteer2}
Zhilin Wang, Yi~Dong, Olivier Delalleau, Jiaqi Zeng, Gerald Shen, Daniel Egert,
  Jimmy~J. Zhang, Makesh~Narsimhan Sreedhar, and Oleksii Kuchaiev.
\newblock {HelpSteer} 2: Open-source dataset for training top-performing reward
  models.
\newblock In \emph{Advances in Neural Information Processing Systems},
  volume~37, pp.\  1474--1501. Curran Associates, Inc., 2024{\natexlab{b}}.
\newblock \doi{10.52202/079017-0047}.

\bibitem[Wataoka et~al.(2024)Wataoka, Takahashi, and
  Ri]{llm-judge-self-bias-watoka}
Koki Wataoka, Tsubasa Takahashi, and Ryokan Ri.
\newblock Self-preference bias in {LLM}-as-a-judge.
\newblock In \emph{NeurIPS Safe Generative AI Workshop}, 2024.
\newblock \doi{10.48550/arXiv.2410.21819}.

\bibitem[Wei et~al.(2025)Wei, Wen, Qiao, Sun, and Ma]{wei2025rocketeval}
Tianjun Wei, Wei Wen, Ruizhi Qiao, Xing Sun, and Jianghong Ma.
\newblock {RocketEval}: Efficient automated {LLM} evaluation via grading
  checklist.
\newblock In \emph{International Conference on Learning Representations
  (ICLR)}, pp.\  58593--58619, 2025.
\newblock \doi{10.48550/arXiv.2503.05142}.

\bibitem[Xia et~al.(2025)Xia, Luz De~Araujo, Zaporojets, and
  Roth]{xia2025calibn}
Yuxi Xia, Pedro~Henrique Luz De~Araujo, Klim Zaporojets, and Benjamin Roth.
\newblock Influences on {LLM} calibration: A study of response agreement, loss
  functions, and prompt styles.
\newblock In \emph{Proceedings of the 63rd Annual Meeting of the Association
  for Computational Linguistics (Volume 1: Long Papers)}, pp.\  3740--3761.
  Association for Computational Linguistics, 2025.
\newblock \doi{10.18653/v1/2025.acl-long.188}.

\bibitem[Xin et~al.(2021)Xin, Tang, Yu, and Lin]{xin2021abstention}
Ji~Xin, Raphael Tang, Yaoliang Yu, and Jimmy Lin.
\newblock The art of abstention: Selective prediction and error regularization
  for natural language processing.
\newblock In \emph{Proceedings of the 59th Annual Meeting of the Association
  for Computational Linguistics and the 11th International Joint Conference on
  Natural Language Processing (Volume 1: Long Papers)}, pp.\  1040--1051,
  Online, 2021. Association for Computational Linguistics.
\newblock \doi{10.18653/v1/2021.acl-long.84}.
\newblock URL \url{https://aclanthology.org/2021.acl-long.84}.

\bibitem[Xu et~al.(2025)Xu, Lu, Ye, Hu, and
  Liu]{xu2025researcherbenchevaluatingdeepai}
Tianze Xu, Pengrui Lu, Lyumanshan Ye, Xiangkun Hu, and Pengfei Liu.
\newblock {ResearcherBench}: Evaluating deep {AI} research systems on the
  frontiers of scientific inquiry, 2025.
\newblock URL \url{https://arxiv.org/abs/2507.16280}.

\bibitem[Ye et~al.(2024)Ye, Kim, Kim, Hwang, Kim, Jo, Thorne, Kim, and
  Seo]{ye2024flask}
Seonghyeon Ye, Doyoung Kim, Sungdong Kim, Hyeonbin Hwang, Seungone Kim, Yongrae
  Jo, James Thorne, Juho Kim, and Minjoon Seo.
\newblock {FLASK}: Fine-grained language model evaluation based on alignment
  skill sets.
\newblock In \emph{Proceedings of the Twelfth International Conference on
  Learning Representations}, pp.\  55361--55414, 2024.
\newblock \doi{10.48550/arXiv.2307.10928}.

\bibitem[Yuksekgonul et~al.(2025)Yuksekgonul, Bianchi, Boen, Liu, Lu, Huang,
  Guestrin, and Zou]{yuksekgonul2024textgrad}
Mert Yuksekgonul, Federico Bianchi, Joseph Boen, Sheng Liu, Pan Lu, Zhi Huang,
  Carlos Guestrin, and James Zou.
\newblock Optimizing generative {AI} by backpropagating language model
  feedback.
\newblock \emph{Nature}, 639\penalty0 (8055):\penalty0 609--616, 2025.
\newblock \doi{10.1038/s41586-025-08661-4}.

\bibitem[Zheng et~al.(2023)Zheng, Chiang, Sheng, Zhuang, Wu, Zhuang, Lin, Li,
  Li, Xing, Zhang, Gonzalez, and Stoica]{zheng2023judging}
Lianmin Zheng, Wei-Lin Chiang, Ying Sheng, Siyuan Zhuang, Zhanghao Wu, Yonghao
  Zhuang, Zi~Lin, Zhuohan Li, Dacheng Li, Eric~P. Xing, Hao Zhang, Joseph~E.
  Gonzalez, and Ion Stoica.
\newblock Judging {LLM}-as-a-judge with {MT-Bench} and {Chatbot Arena}.
\newblock In \emph{Advances in Neural Information Processing Systems (Datasets
  and Benchmarks Track)}, volume~36, pp.\  46595--46623, 2023.

\bibitem[Zhou et~al.(2024{\natexlab{a}})Zhou, Wan, Liu, Collier, Vuli{\'c}, and
  Korhonen]{zhou2024zepo}
Han Zhou, Xingchen Wan, Yinhong Liu, Nigel Collier, Ivan Vuli{\'c}, and Anna
  Korhonen.
\newblock Fairer preferences elicit improved human-aligned large language model
  judgments.
\newblock In \emph{Proceedings of the 2024 Conference on Empirical Methods in
  Natural Language Processing}, pp.\  1241--1252. Association for Computational
  Linguistics, 2024{\natexlab{a}}.
\newblock \doi{10.18653/v1/2024.emnlp-main.72}.

\bibitem[Zhou et~al.(2024{\natexlab{b}})Zhou, Wan, Proleev, Mincu, Chen,
  Heller, and Roy]{zhou2023batch}
Han Zhou, Xingchen Wan, Lev Proleev, Diana Mincu, Jilin Chen, Katherine Heller,
  and Subhrajit Roy.
\newblock Batch calibration: Rethinking calibration for in-context learning and
  prompt engineering.
\newblock In \emph{Proceedings of the Twelfth International Conference on
  Learning Representations}, pp.\  49--70, 2024{\natexlab{b}}.
\newblock \doi{10.48550/arXiv.2309.17249}.

\bibitem[Zhou et~al.(2023)Zhou, Lu, Mishra, Brahma, Basu, Luan, Zhou, and
  Hou]{zhou2023ifeval}
Jeffrey Zhou, Tianjian Lu, Swaroop Mishra, Siddhartha Brahma, Sujoy Basu,
  Yi~Luan, Denny Zhou, and Le~Hou.
\newblock Instruction-following evaluation for large language models, 2023.
\newblock URL \url{https://arxiv.org/abs/2311.07911}.

\end{thebibliography}
\bibliographystyle{colm2026_conference}

\appendix
\section*{Appendix}

\section*{Outline of the Appendix}

The appendix spans ten sections containing implementation details, dataset documentation, and extended results that support but are not essential for following the main text. We summarize them here for navigability.

\begin{itemize}
\item \textbf{Appendix~\ref{appendix:design-space}} presents a visual taxonomy of rubric-based evaluation design choices across five dimensions.
\item \textbf{Appendix~\ref{appendix:framework-details}} provides code examples, educational measurement foundations, production infrastructure details, default evaluation prompts, hyperparameters, and correctness verification.
\item \textbf{Appendix~\ref{appendix:rubric-evaluation-report-example}} shows an example concrete evaluation report produced by \texttt{Autorubric}.
\item \textbf{Appendix~\ref{appendix:ricechem}} details the RiceChem dataset conversion, per-question rubrics, paired inference, an independent per-criterion diagnostic, an exploratory few-shot sweep, weight inference, score distributions, cold-start results, and prior baselines.
\item \textbf{Appendix~\ref{appendix:researcherbench}} covers additional details of the ResearcherBench dataset, rubric structure, cross-judge agreement, statistical significance tests, disagreement taxonomy, score calibration, cost analysis, and rubric statistics.
\item \textbf{Appendix~\ref{appendix:charm100-dataset}} documents the CHARM-100 dataset: motivation, annotation schema, design principles, topic coverage, label distributions, anti-pattern taxonomy, edge cases, second-annotation agreement, and limitations.
\item \textbf{Appendix~\ref{appendix:charm100-results}} reports CHARM-100 evaluation results including confusion matrices, aggregate metrics, N/A handling, and the configuration-sensitivity analysis.
\item \textbf{Appendix~\ref{appendix:skill-improvement}} describes cross-model robustness probes for skill revision and their limits for identifying judge self-preference.
\item \textbf{Appendix~\ref{appendix:cross-benchmark}} synthesizes reliability patterns across all three benchmarks.
\item \textbf{Appendix~\ref{appendix:rlrr}} reports additional RL training diagnostics, checkpoint-selection details, and a cross-judge robustness probe.
\end{itemize}

\FloatBarrier
\section{Design space taxonomy}
\label{appendix:design-space}

\begin{figure*}[t]
    \centering
    \includegraphics[width=0.8\linewidth]{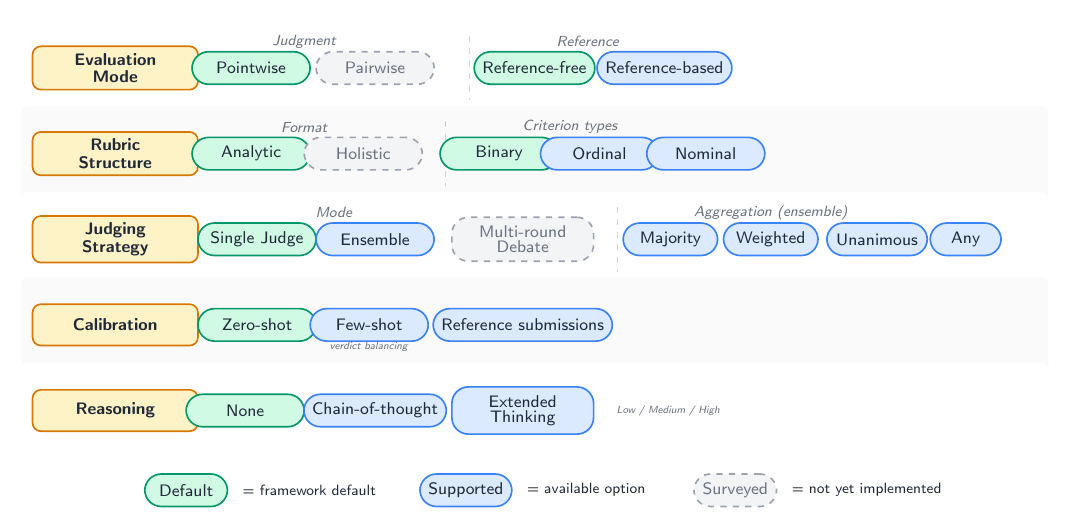}
    \caption{Design space for rubric-based LLM evaluation across five dimensions. Green pills indicate \texttt{Autorubric} defaults, blue pills indicate supported options, and gray dashed pills indicate paradigms covered in this paper but not yet implemented in the framework.}
    \label{fig:design-space}
\end{figure*}

\FloatBarrier
\section{Framework details}
\label{appendix:framework-details}

This appendix provides the code examples, educational foundations, and production infrastructure details summarized in the main text.

\paragraph{Version provenance.} Listings and framework defaults in this appendix target the public v1.5.3 release. The reported experiments are stored historical runs rather than v1.5.3 reruns: the primary ResearcherBench and CHARM artifacts were introduced with \texttt{Autorubric} 0.3.2 (commit \texttt{d81bea9}), while the separate CHARM sensitivity artifacts used their contemporaneous March 2026 code. Prompt and default changes after those runs are therefore documented as current API behavior, not retroactively attributed to the reported predictions.

\subsection{Code examples}
\label{appendix:code-examples}

The following listings illustrate \texttt{Autorubric}'s API for each major capability described in Sections~\ref{sec:background}--\ref{sec:failure-modes}. All listings target \texttt{Autorubric} v1.5.3; readers using later releases should consult the current public API documentation at \url{https://autorubric.org/docs/api/}.

\begin{code}{Defining an analytic rubric with weighted criteria.}
from autorubric import Rubric, Criterion

rubric = Rubric([
    Criterion(weight=10.0, requirement="States the correct answer"),
    Criterion(weight=8.0, requirement="Provides supporting evidence"),
    Criterion(weight=5.0, requirement="Uses clear, concise language"),
])
\end{code}

\begin{code}{Binary, ordinal, and nominal criterion types.}
from autorubric import Criterion

# Binary criterion (default)
Criterion(weight=10.0, requirement="The answer is factually correct")

# Ordinal criterion with explicit values
Criterion(
    weight=10.0,
    requirement="Rate the clarity of explanation",
    scale_type="ordinal",
    options=[
        {"label": "Unclear", "value": 0.0},
        {"label": "Somewhat clear", "value": 0.5},
        {"label": "Very clear", "value": 1.0},
    ]
)

# Nominal criterion (unordered categories)
Criterion(
    weight=5.0,
    requirement="Primary error type (if any)",
    scale_type="nominal",
    options=[
        {"label": "Factual error", "value": 0.0},
        {"label": "Logical error", "value": 0.0},
        {"label": "No error", "value": 1.0},
    ]
)
\end{code}

\begin{code}{Positive and negative (penalty) criteria with differential weights.}
from autorubric import Rubric

rubric = Rubric.from_dict([
    # Positive criteria (reward when met)
    {"weight": 10.0, "requirement": "Correctly identifies the main claim"},
    {"weight": 8.0, "requirement": "Cites at least two sources"},

    # Negative criteria (penalty when met)
    {"weight": -15.0, "requirement": "Contains hallucinated citations"},
    {"weight": -10.0, "requirement": "Contradicts established facts"},
])
\end{code}

\begin{code}{Single-judge and multi-judge ensemble grading.}
from autorubric import LLMConfig
from autorubric.graders import CriterionGrader, JudgeSpec

# Single judge
grader = CriterionGrader(
    llm_config=LLMConfig(model="openai/gpt-4.1-mini")
)

# Ensemble with three judges from different model families
grader = CriterionGrader(
    judges=[
        JudgeSpec(
            llm_config=LLMConfig(model="openai/gpt-4.1"),
            judge_id="gpt-4",
            weight=1.0,
        ),
        JudgeSpec(
            llm_config=LLMConfig(model="anthropic/claude-sonnet-4-5-20250929"),
            judge_id="claude-sonnet",
            weight=1.2,
        ),
        JudgeSpec(
            llm_config=LLMConfig(model="gemini/gemini-2.5-flash"),
            judge_id="gemini-flash",
            weight=1.0,
        ),
    ],
    aggregation="weighted",  # or "majority", "unanimous", "any"
)

# Grade a submission
result = await rubric.grade(to_grade=response, grader=grader, query=prompt)
score = f"{result.score:.2f}" if result.score is not None else "n/a"
agreement = (
    f"{result.mean_agreement:.1%}"
    if result.mean_agreement is not None else "n/a"
)
print(f"Score: {score}, Agreement: {agreement}")
\end{code}

\begin{code}{Few-shot calibration with balanced verdict sampling.}
from autorubric import FewShotConfig, LLMConfig, RubricDataset
from autorubric.graders import CriterionGrader

dataset = RubricDataset.from_file("labeled_data.json")

# Split dataset into training (for examples) and test
train_data, test_data = dataset.split_train_test(n_train=100, stratify=True, seed=42)

# Configure few-shot calibration
grader = CriterionGrader(
    llm_config=LLMConfig(model="openai/gpt-4.1-mini"),
    training_data=train_data,
    few_shot_config=FewShotConfig(
        n_examples=3,           # Include 3 examples per criterion
        balance_verdicts=True,  # Balance available label classes
        include_reason=True,    # Include labeled-example explanations
        seed=42,
    ),
)
\end{code}

\begin{code}{Ensemble with mixed reasoning levels per judge.}
from autorubric import LLMConfig, ThinkingConfig
from autorubric.graders import CriterionGrader, JudgeSpec

grader = CriterionGrader(
    judges=[
        JudgeSpec(
            llm_config=LLMConfig(
                model="anthropic/claude-sonnet-4-20250514",
                thinking=ThinkingConfig(level="high"),
            ),
            judge_id="claude-high",
            weight=2.0,
        ),
        JudgeSpec(
            llm_config=LLMConfig(model="openai/gpt-4.1"),
            judge_id="gpt-no-thinking",
            weight=1.0,  # no thinking
        ),
    ],
    aggregation="weighted",
)
\end{code}

\begin{code}{Option shuffling for position bias mitigation.}
from autorubric import LLMConfig
from autorubric.graders import CriterionGrader

# Position bias mitigation is enabled by default
grader = CriterionGrader(
    llm_config=LLMConfig(model="openai/gpt-4.1-mini"),
    shuffle_options=True,  # Default: randomize option order
    seed=42,               # Reproducible non-LLM randomness
)

# Disable shuffling (e.g., for manual prompt inspection)
grader = CriterionGrader(
    llm_config=LLMConfig(model="openai/gpt-4.1-mini"),
    shuffle_options=False,
)
\end{code}

\begin{code}{Strategies for handling unassessable criteria.}
from autorubric import CannotAssessConfig, CannotAssessStrategy, LLMConfig
from autorubric.graders import CriterionGrader

# Default: skip unassessable criteria (adjust denominator)
grader = CriterionGrader(llm_config=LLMConfig(model="openai/gpt-4.1-mini"))

# Conservative: treat cannot-assess as failure
grader = CriterionGrader(
    llm_config=LLMConfig(model="openai/gpt-4.1-mini"),
    cannot_assess_config=CannotAssessConfig(
        strategy=CannotAssessStrategy.FAIL
    ),
)

# Give partial credit (50%)
grader = CriterionGrader(
    llm_config=LLMConfig(model="openai/gpt-4.1-mini"),
    cannot_assess_config=CannotAssessConfig(
        strategy=CannotAssessStrategy.PARTIAL,
        partial_credit=0.5,
    ),
)
\end{code}

\begin{code}{Accessing per-criterion explanations for feedback.}
from autorubric import CriterionVerdict

report = await rubric.grade(
    to_grade=submission,
    grader=grader,
    query=prompt,
)

# Extract explanations for unmet criteria
for criterion_report in report.report or []:
    if criterion_report.final_verdict == CriterionVerdict.UNMET:
        print(f"Criterion: {criterion_report.criterion.requirement}")
        print(f"Reason: {criterion_report.final_reason}")
\end{code}

\subsection{Educational measurement and psychometrics}
\label{appendix:edu-measurement}

Educational measurement provides frameworks for designing valid and reliable assessments~\citep{mckeown2018heqco}. \texttt{Autorubric}'s design is informed by three principles from this literature. \emph{Unidimensionality}: each criterion should measure a single construct. A criterion that conflates multiple constructs (e.g., ``The response is accurate and well-written'') produces scores that are difficult to interpret. \texttt{Autorubric} supports criterion-level decomposition by evaluating each \texttt{Criterion.requirement} separately, but v1.5.3 does not verify that the requirement is semantically unidimensional. \emph{Behavioral anchors}: score levels should be defined by observable behaviors rather than evaluative adjectives. ``The response cites at least three peer-reviewed sources'' is behavioral; ``The response demonstrates excellent research'' is evaluative and subject to interpretation. For multi-choice criteria, option labels serve as behavioral anchors. \emph{Construct alignment}: criteria should align with the construct being measured. If the goal is to measure ``helpfulness,'' criteria should operationalize helpfulness (task completion, user satisfaction proxies) rather than proxies that may not correlate (length, formality). Construct alignment is a rubric design responsibility; \texttt{Autorubric} cannot enforce it but provides the structure for explicit criteria that can be audited for alignment.

From psychometrics~\citep{comer2009rubrics}, the framework provides quantitative methods for assessing measurement quality. \emph{Reliability metrics}: for categorical judgments, Cohen's $\kappa$ adjusts for chance agreement; for ordinal scales, quadratic weighted $\kappa$ accounts for the magnitude of disagreement. Psychometrics also uses intraclass correlation for continuous scores and multiple raters, but \texttt{Autorubric} v1.5.3 does not implement that statistic. \emph{Validity}: reliability is necessary but not sufficient---a measure can be reliable but not valid. Content validity asks whether criteria cover the intended domain; criterion validity asks whether scores correlate with external measures; construct validity asks whether the measure behaves as theory predicts. Distribution analysis via Earth Mover's Distance, following \citet{he2025llmjudge}, can reveal systematic biases (e.g., central tendency) that point estimates miss.

Some LLM-specific concerns---token-level position bias, prompt sensitivity, self-preference bias~\citep{llm-judge-self-bias-panickssery, llm-judge-self-bias-watoka}---have no educational precedent and require the mitigations in Section~\ref{sec:failure-modes}.

\begin{code}{Computing reliability metrics against human judgments.}
from autorubric import evaluate

# Run evaluation on dataset with ground truth
result = await evaluate(dataset, grader, show_progress=True)

# Compute metrics against human judgments
metrics = result.compute_metrics(
    dataset,
    bootstrap=True,        # Confidence intervals
    n_bootstrap=1000,
    per_judge=True,        # Per-judge breakdown for ensembles
)

# Access results
accuracy = metrics.criterion_accuracy
kappa = metrics.mean_kappa
rho = metrics.score_spearman.coefficient
print(f"Accuracy: {accuracy:.1%}" if accuracy is not None else "Accuracy: n/a")
print(f"Cohen's kappa: {kappa:.3f}" if kappa is not None else "Cohen's kappa: n/a")
print(f"Score correlation: {rho:.3f}" if rho is not None else "Score correlation: n/a")
print(metrics.summary())  # Formatted report
\end{code}

\subsection{Production infrastructure}
\label{appendix:production}

For production deployment, \texttt{Autorubric} provides response caching for reproducibility and cost control (keyed on the model, prompts, response schema, and standard sampling and thinking parameters), checkpoint-based resumable evaluation that stores intermediate results and selected grader settings, including the master seed---allowing continuation after interruptions and exact replication of the recorded non-LLM randomness, and per-provider rate limiting via semaphores to prevent throttling while maximizing throughput. The manifest is not a complete serialization of every prompt, model, or grader option, so exact reconstruction additionally requires retaining the evaluation code and inputs. Each LLM call records a transient cost via LiteLLM's \texttt{completion\_cost()}, and aggregation accumulates per-item and run-level totals. Per-item wall-clock \texttt{duration\_seconds} uses \texttt{time.perf\_counter()}; the returned \texttt{EvalResult} exposes total cost and \texttt{EvalTimingStats} (mean, min, max, p50, p95, and items-per-second). By default, \texttt{items.jsonl} stores per-item cost and duration, while \texttt{manifest.json} stores selected configuration fields, seed, status, and total wall time.

\begin{code}{Production infrastructure: caching, rate limiting, and checkpointing.}
from autorubric import EvalConfig, EvalRunner, LLMConfig
from autorubric.graders import CriterionGrader

# Caching: avoid redundant LLM calls
grader = CriterionGrader(
    llm_config=LLMConfig(
        model="openai/gpt-4.1-mini",
        cache_enabled=True,
        cache_dir=".autorubric_cache",
        cache_ttl=3600,  # 1 hour TTL
    )
)

# Rate limiting: respect provider quotas
grader = CriterionGrader(
    llm_config=LLMConfig(
        model="openai/gpt-4.1-mini",
        max_parallel_requests=10,  # Per-provider semaphore
    )
)

# Checkpointing: resumable batch evaluation
config = EvalConfig(
    experiment_name="my-eval",
    experiments_dir="./experiments",
    resume=True,  # Continue from checkpoint if interrupted
)
runner = EvalRunner(dataset=dataset, grader=grader, config=config)
result = await runner.run()

# Cost tracking
tokens = result.total_token_usage.total_tokens if result.total_token_usage else 0
cost = result.total_completion_cost or 0.0
print(f"Tokens: {tokens}")
print(f"Cost: {cost:.4f} USD")
\end{code}

\subsection{Default evaluation prompts}
\label{appendix:prompts}

Listings~\ref{listing:binary-prompt} and~\ref{listing:multi-choice-prompt} show abridged adaptations of the v1.5.3 base system prompts for binary and multi-choice criterion evaluation, respectively. They summarize the principal verdict and output rules rather than reproducing the constants verbatim. The complete, exact prompts are available in the tagged source at \url{https://github.com/delip/autorubric/blob/v1.5.3/src/autorubric/prompts.py}. When a default prompt is used, supplying labeled training data appends the corresponding few-shot addition. A custom prompt replaces the base and must include any desired few-shot instructions itself.

\begin{codeoutput}{Abridged adaptation of the v1.5.3 base prompt for binary criterion evaluation.\label{listing:binary-prompt}}
You are an expert evaluation judge. Your task is to determine whether
a single criterion is satisfied by a given submission. Be precise,
evidence-based, and consistent.

You will receive a <criterion_type> (positive or negative),
a <criterion>, and a <submission> to evaluate. Your verdict must
be one of:
- "MET": The thing described in the criterion IS present
- "UNMET": The thing described in the criterion IS NOT present
- "CANNOT_ASSESS": Insufficient evidence to determine (use rarely)

Evaluate this criterion independently. Do not let overall submission
quality influence your judgment.

CRITERION TYPES:
<criterion_type> indicates whether the criterion describes something
desirable (positive) or undesirable (negative). Your job is THE SAME
for both: determine if the thing described is present.

POSITIVE CRITERIA: desired traits that should be present.
NEGATIVE CRITERIA: active errors or mistakes. MET means the
submission advocates or states the problematic thing; UNMET means
it does NOT make this error or mentions it only to warn against it.

EVALUATION RULES:
- For numerical values: check specified ranges or exact matches.
- For factual claims: verify presence and accuracy.
- For required elements: confirm presence, count precisely.
- For exclusion requirements: confirm restricted content is absent.
- Be strict about factual accuracy but flexible about wording.
- Accept semantically equivalent statements or logical implications.

IMPLICIT SATISFACTION:
A criterion can be satisfied implicitly through context, tone, or
logical implication.

CANNOT_ASSESS VERDICT:
Use only when you genuinely cannot determine if the criterion is met
(e.g., missing attachments, garbled text). Do NOT use when you can
make a reasonable inference or when the criterion is simply not met.

RESPONSE FORMAT:
Respond with valid JSON:
{"criterion_status": "MET"|"UNMET"|"CANNOT_ASSESS",
 "explanation": "..."}

Provide a 1-2 sentence explanation. Cite specific text from the
submission as evidence for your verdict.
\end{codeoutput}

\begin{codeoutput}{Abridged adaptation of the v1.5.3 base prompt for multi-choice criterion evaluation.\label{listing:multi-choice-prompt}}
You are an expert evaluation judge. Your task is to select the best
matching option for a given submission from a set of predefined
choices. Be precise, evidence-based, and consistent.

You will receive a <question>, numbered <options>, and a <submission>.

EVALUATION RULES:
- Review ALL options before selecting.
- Base judgment on submission content, not assumptions about intent.
- Be strict about factual accuracy but flexible about wording.
- For ordinal scales, treat options as points on a continuum.
- Do not default to middle options out of uncertainty.
- When borderline, select the option whose description more precisely
  matches the specific evidence in the submission.

NA / NOT APPLICABLE OPTIONS:
Select NA only when the question genuinely cannot be answered
(missing attachments, garbled text). Do NOT select NA when you can
make a reasonable inference or when the submission simply does not
match well (select the closest match instead).

RESPONSE FORMAT:
Respond with valid JSON:
{"selected_option": <number>, "explanation": "..."}

Provide a 1-2 sentence explanation. Cite specific text from the
submission as evidence for your selection.
\end{codeoutput}

\subsection{Default hyperparameters}
\label{sec:default-hyperparams}

Table~\ref{tab:default-hyperparams} lists the principal v1.5.3 evaluation defaults. Few-shot calibration is opt-in: its defaults apply only when training data are supplied, and the settings used in each experiment are stated in the corresponding methods.

\begin{table}[h]
\centering
\small
\begin{tabularx}{\linewidth}{@{}llX@{}}
\toprule
\textbf{Component} & \textbf{Parameter} & \textbf{Default} \\
\midrule
Few-shot & examples per criterion & 3 \\
         & label-class balancing & on \\
         & include labeled-example explanation & off \\
Single-LLM mode & effective judges & 1 \\
Aggregation & binary & majority \\
            & ordinal & mean \\
            & nominal & mode \\
Multi-choice & option shuffling & on \\
             & automatic N/A option & on \\
Cannot-assess & strategy & SKIP \\
Scoring & normalized score & on \\
\bottomrule
\end{tabularx}
\caption{Principal framework evaluation defaults in v1.5.3. \texttt{CriterionGrader} requires either \texttt{llm\_config} or an explicit \texttt{judges} list; single-LLM mode is normalized internally to one effective judge. Few-shot defaults apply only when labeled training data are supplied. Reported experiments use the settings stated in their methods.}
\label{tab:default-hyperparams}
\end{table}

\subsection{Correctness verification}
\label{appendix:correctness}

Score calculation must be correct by construction. We verify \texttt{Autorubric}'s score aggregation against hand-computed examples covering standard positive criteria with varying weights, all four \texttt{CANNOT\_ASSESS} handling strategies, the penalty-only normalization branch in Equation~\ref{eq:score-aggregation}, mixed positive and negative weights, and multi-choice criteria with explicit option values. All tested cases agree with the hand calculations within floating-point tolerance. The test suite of more than 400 tests, covering edge cases in score computation, metric calculation, and ensemble aggregation, is included in the repository.

\FloatBarrier
\section{Rubric evaluation report example}
\label{appendix:rubric-evaluation-report-example}

\begin{codeoutput}{Sample evaluation report produced by Autorubric.}
====================================================
AutoRubric Demo: Criterion-Level Accuracy Evaluation
Model: gemini/gemini-2.5-flash
Thinking: medium
Max Parallel Requests: 10
=====================================================
Prompt: Explain the causes and effects of the Industrial Revolution.
Rubric Criteria (total positive weight: 80.0):
  1. Causes     [ +30.0]
  2. Effects    [ +30.0]
  3. Structure  [ +12.0]
  4. Britain    [  +8.0]
  5. Errors     [ -15.0]
----------------------------------------------------
Grading 11 items with EvalRunner...
(Checkpoints saved to experiments/ directory for resumption)
  Evaluating -------------- 11/11 (0.43/s) 0:00:25 0:00:00
Experiment saved to: experiments/slim-deer
====================
METRICS SUMMARY
====================
Items: 11, Criteria: 5
Criterion-Level Metrics:
  Accuracy:   92.2%
  Precision:  0.95
  Recall:     0.87
  F1:         0.91
  Mean Kappa: 0.846
Score-Level Metrics:
  RMSE:     0.3246
  MAE:      0.1489
  Spearman: 0.6413 (moderate positive)
  Kendall:  0.6527 (moderate positive)
  Pearson:  0.6699 (moderate positive)
Bias Analysis:
  Mean Bias:   -0.0875 (negative)
  Significant: No
Bootstrap CIs (95%):
  Accuracy: [84.3%, 98.0%]
  Kappa:    [0.667, 0.961]
  RMSE:     [0.0452, 0.5417]
Per-Criterion Breakdown:
Criterion                 Acc     Prec      Rec       F1    Kappa
-----------------------------------------------------------------
Causes                 100.0%     1.00     1.00     1.00    1.000
Effects                 90.0%     0.88     1.00     0.93    0.737
Structure               72.7%     1.00     0.57     0.73    0.492
Britain                100.0%     1.00     1.00     1.00    1.000
Errors                 100.0%     1.00     1.00     1.00    1.000

Total Cost: $0.090 (11 items, 124K tokens)
\end{codeoutput}

\FloatBarrier
\section{RiceChem additional results}
\label{appendix:ricechem}

\subsection{Dataset and conversion details}

The RiceChem dataset~\citep{sonkar2024automatedlonganswergrading} was introduced for automated long answer grading (ALAG), framed as a rubric entailment problem: given a student response and a rubric criterion, determine whether the response satisfies (entails) the criterion. The raw data comprises eight CSV files (four student answer files and four graded rubric files), one pair per question. Each graded rubric file contains per-student binary criterion annotations, numerical scores, and optional TA adjustment columns.

\paragraph{Criterion selection.} The original rubric files contain both positive criteria (knowledge demonstrated) and negative/flag columns (error indicators and blanks). Following \citet{sonkar2024automatedlonganswergrading}, who report 27 rubric items, we retain only positive criteria: 8 of 11 columns for Q1 (excluding \texttt{incorrect}, \texttt{Blank}, \texttt{Core charge calculation error}), 6 of 9 for Q2 (excluding \texttt{Incorrect statement included}, \texttt{Incorrect}, \texttt{Blank}), 7 of 9 for Q3 (excluding \texttt{Correct response}, \texttt{Incorrect/Blank response}), and 6 of 8 for Q4 (excluding \texttt{incorrect/misleading statement}, \texttt{incorrect/missing answer}).

\paragraph{Blank filtering.} Empty or blank-flagged submissions are excluded: 1 from Q1, 11 from Q2, 10 from Q3, and 2 from Q4.

\paragraph{Ground truth encoding.} Each binary annotation is mapped to a \texttt{CriterionVerdict}: \texttt{TRUE} $\rightarrow$ \texttt{MET}, \texttt{FALSE} $\rightarrow$ \texttt{UNMET}. The resulting verdict vectors enable direct computation of criterion-level agreement metrics.

\subsection{Per-question criteria and weights}

Table~\ref{tab:ricechem-questions} summarizes the rubric structure for each question.

\begin{table}[ht]
\centering
\small
\begin{tabular}{lcccc}
\toprule
\textbf{Question} & \textbf{Topic} & \textbf{Criteria} & \textbf{Students} & \textbf{Max Score} \\
\midrule
Q1 & Ionization energies (Coulomb's Law) & 8 & 327 & $\sim$8 \\
Q2 & Quantized absorption vs.\ photoejection & 6 & 317 & $\sim$8 \\
Q3 & Hybrid orbitals in methanimine & 7 & 298 & 9 \\
Q4 & Law of Multiple Proportions & 6 & 298 & $\sim$8 \\
\midrule
\textbf{Total} & & \textbf{27} & \textbf{1,240} & \\
\bottomrule
\end{tabular}
\caption{RiceChem question structure. All criteria are binary (MET/UNMET). Questions 1--3 expect responses of approximately 150 words; Question 4 approximately 75 words.}
\label{tab:ricechem-questions}
\end{table}

Tables~\ref{tab:ricechem-q1}--\ref{tab:ricechem-q4} list each criterion with its inferred weight and the proportion of students who satisfied it (MET rate).

\begin{table}[ht]
\centering
\small
\begin{tabular}{lcc}
\toprule
\textbf{Q1 Criterion} & \textbf{Weight} & \textbf{MET Rate} \\
\midrule
Correctly cites decreased electron-electron repulsion & 1.01 & 83.8\% \\
Relates decreased repulsion to decreased potential energy & 1.03 & 63.3\% \\
3rd and 4th electrons feel same core charge & 0.96 & 58.7\% \\
3rd and 4th electrons ionized from $n{=}3$ shell, same radius & 1.00 & 55.0\% \\
5th electron from $n{=}2$ shell feels higher core charge & 0.95 & 71.3\% \\
5th electron from $n{=}2$ shell has smaller radius & 0.98 & 83.2\% \\
Correctly explains PE--IE relationship (full) & 1.97 & 41.9\% \\
Partially explains PE--IE relationship & 0.99 & 17.1\% \\
\bottomrule
\end{tabular}
\caption{Q1 criteria: Silicon ionization energies. Students explain why successive ionization energies increase and why the 4th-to-5th jump is disproportionately large, using core charge and Coulomb's Law.}
\label{tab:ricechem-q1}
\end{table}

\begin{table}[ht]
\centering
\small
\begin{tabular}{lcc}
\toprule
\textbf{Q2 Criterion} & \textbf{Weight} & \textbf{MET Rate} \\
\midrule
Frequency proportional to energy of light & 1.93 & 57.4\% \\
Energy levels of an electron are quantized & 0.96 & 49.8\% \\
Fully explains energy/frequency condition & 2.04 & 27.1\% \\
Partially explains energy/frequency condition & 1.02 & 24.9\% \\
Minimum energy needed to eject electron & 0.96 & 72.6\% \\
Additional energy becomes kinetic energy & 1.98 & 42.0\% \\
\bottomrule
\end{tabular}
\caption{Q2 criteria: Light absorption vs.\ photoejection. Students reconcile quantized absorption (only specific frequencies excite electrons) with continuous photoejection (any frequency above threshold ejects electrons).}
\label{tab:ricechem-q2}
\end{table}

\begin{table}[ht]
\centering
\small
\begin{tabular}{lcc}
\toprule
\textbf{Q3 Criterion} & \textbf{Weight} & \textbf{MET Rate} \\
\midrule
Sentence 1 correct: VBT half-filled orbitals & 2.00 & 51.7\% \\
Sentence 2: correct number of hybrid orbitals & 2.00 & 40.9\% \\
Sentence 2: correct type (sp2) & 1.00 & 58.7\% \\
Sentence 3: nitrogen is hybridized & 1.00 & 69.1\% \\
Sentence 3: correct hybridization type (sp2) & 1.00 & 69.1\% \\
Sentence 3: hybrid orbital bonds described & 1.00 & 15.1\% \\
Sentence 3: unhybridized orbital bonds described & 1.00 & 24.5\% \\
\bottomrule
\end{tabular}
\caption{Q3 criteria: Hybrid orbitals in methanimine (CH$_2$NH). Students assess a deliberately flawed peer response, identifying errors in sp3 hybridization of carbon and the claim that nitrogen does not hybridize.}
\label{tab:ricechem-q3}
\end{table}

\begin{table}[ht]
\centering
\small
\begin{tabular}{lcc}
\toprule
\textbf{Q4 Criterion} & \textbf{Weight} & \textbf{MET Rate} \\
\midrule
Fixed mass of one element & 0.98 & 89.3\% \\
Mass data in Law of Multiple Proportions & 0.98 & 88.6\% \\
Combine to form compounds & 0.98 & 88.3\% \\
Integer/whole number ratio & 1.01 & 93.0\% \\
Whole numbers mean indivisible/discrete & 1.98 & 80.5\% \\
Indivisible unit of mass = atom & 2.00 & 67.8\% \\
\bottomrule
\end{tabular}
\caption{Q4 criteria: Law of Multiple Proportions. Students explain how this law provides evidence that matter is composed of atoms. Most criteria have MET rates above 80\%, making Q4 the easiest question.}
\label{tab:ricechem-q4}
\end{table}

\FloatBarrier
\subsection{Per-question and per-criterion reliability}
\label{appendix:ricechem-per-criterion-kappa}

Table~\ref{tab:ricechem-per-question-kappa} reports per-question aggregate Cohen's $\kappa$ for the paired 5-shot rerun used in the main text (661/819 correct). Q3 achieves the highest agreement ($\kappa = 0.831$), consistent with its error-identification task producing unambiguous MET/UNMET distinctions. Q1 is lowest ($\kappa = 0.435$), dragged down by the partial-credit criterion discussed below. Q2 and Q4 show moderate agreement ($\kappa \approx 0.49$), though Q4's high accuracy (86.2\%) masks low $\kappa$ due to high base rates on several criteria.

\begin{table}[ht]
\centering
\small
\begin{tabular}{lcccc}
\toprule
\textbf{Question} & \textbf{N} & \textbf{Acc.} & \textbf{$\kappa$} & \textbf{95\% CI} \\
\midrule
Q1 (ionization energies) & 256 & 73.4\% & 0.435 & [0.318, 0.544] \\
Q2 (absorption vs.\ photoejection) & 186 & 73.7\% & 0.492 & [0.380, 0.604] \\
Q3 (hybrid orbitals) & 203 & 91.6\% & 0.831 & [0.750, 0.902] \\
Q4 (Law of Multiple Proportions) & 174 & 86.2\% & 0.489 & [0.302, 0.656] \\
\midrule
\textbf{Aggregate} & 819 & 80.7\% & 0.587 & [0.531, 0.643] \\
\bottomrule
\end{tabular}
\caption{Per-question aggregate Cohen's $\kappa$ for the paired RiceChem rerun (5-shot, Gemini-3-Flash; 661/819 correct). Reported 95\% bootstrap CIs use 10{,}000 decision-level resamples and are descriptive because they do not cluster by student response. Each row pools all criteria within that question.}
\label{tab:ricechem-per-question-kappa}
\end{table}

Separately, Table~\ref{tab:ricechem-per-criterion-kappa} disaggregates an independent 5-shot diagnostic replicate (663/819 correct, 81.0\%). Its stochastic predictions are not the paired rerun used for the 80.7\% descriptive result. Three patterns emerge. First, Q3 criteria achieve the highest per-criterion $\kappa$ (six of seven above 0.76), likely because the task (identifying errors in a flawed peer response) produces clear-cut MET/UNMET distinctions. Second, criteria with very high MET rates ($>$89\%, e.g., Q4's \texttt{combine\_compounds} at 93.1\%) produce $\kappa \approx 0$ despite high accuracy, because chance agreement is also high---the judge can achieve 93\% accuracy by predicting MET uniformly. Third, \texttt{partial\_pe\_ie\_explanation} (Q1) shows negative $\kappa$ ($-0.385$), indicating systematic disagreement: the judge tends to award credit for partial explanations that the human grader did not, a known difficulty with partial-credit criteria in automated grading.

\begin{table}[ht]
\centering
\small
\begin{tabular}{llcccc}
\toprule
\textbf{Q} & \textbf{Criterion} & \textbf{N} & \textbf{Acc.} & \textbf{$\kappa$} & \textbf{95\% CI} \\
\midrule
Q1 & higher\_core\_charge & 32 & 96.9\% & 0.932 & [0.753, 1.000] \\
Q1 & same\_core\_charge & 32 & 93.8\% & 0.871 & [0.676, 1.000] \\
Q1 & decreased\_repulsion & 32 & 93.8\% & 0.833 & [0.524, 1.000] \\
Q1 & same\_shell\_radius & 32 & 81.2\% & 0.602 & [0.242, 0.867] \\
Q1 & smaller\_radius & 32 & 84.4\% & 0.452 & [$-$0.053, 0.818] \\
Q1 & repulsion\_potential\_energy & 32 & 65.6\% & 0.362 & [0.098, 0.632] \\
Q1 & full\_pe\_ie\_explanation & 32 & 65.6\% & 0.323 & [0.000, 0.629] \\
Q1 & partial\_pe\_ie\_explanation & 32 & 15.6\% & $-$0.385 & [$-$0.708, $-$0.113] \\
\midrule
Q2 & additional\_kinetic & 31 & 93.5\% & 0.865 & [0.650, 1.000] \\
Q2 & energy\_levels\_quantized & 31 & 80.6\% & 0.594 & [0.305, 0.860] \\
Q2 & full\_energy\_freq & 31 & 77.4\% & 0.553 & [0.268, 0.807] \\
Q2 & min\_energy\_eject & 31 & 83.9\% & 0.471 & [0.000, 0.839] \\
Q2 & freq\_proportional\_energy & 31 & 67.7\% & 0.365 & [0.097, 0.624] \\
Q2 & partial\_energy\_freq & 31 & 45.2\% & 0.108 & [$-$0.095, 0.326] \\
\midrule
Q3 & sentence3\_n\_hybridized & 29 & 96.6\% & 0.910 & [0.701, 1.000] \\
Q3 & sentence2\_correct\_type\_sp2 & 29 & 96.6\% & 0.901 & [0.633, 1.000] \\
Q3 & sentence1\_vbt\_half\_filled & 29 & 93.1\% & 0.848 & [0.613, 1.000] \\
Q3 & sentence3\_correct\_type\_sp2 & 29 & 93.1\% & 0.847 & [0.589, 1.000] \\
Q3 & sentence2\_correct\_number & 29 & 93.1\% & 0.828 & [0.525, 1.000] \\
Q3 & sentence3\_unhybridized\_bonds & 29 & 89.7\% & 0.765 & [0.482, 1.000] \\
Q3 & sentence3\_hybrid\_bonds & 29 & 79.3\% & 0.494 & [0.110, 0.812] \\
\midrule
Q4 & indivisible\_atom & 29 & 79.3\% & 0.586 & [0.289, 0.861] \\
Q4 & fixed\_mass & 29 & 93.1\% & 0.473 & [0.000, 1.000] \\
Q4 & mass\_data\_lomp & 29 & 93.1\% & 0.473 & [0.000, 1.000] \\
Q4 & whole\_numbers\_indivisible & 29 & 58.6\% & 0.028 & [$-$0.169, 0.261] \\
Q4 & combine\_compounds & 29 & 93.1\% & 0.000 & [0.000, 0.000] \\
Q4 & integer\_ratio & 29 & 89.7\% & 0.000 & [0.000, 0.000] \\
\midrule
& \textbf{Aggregate} & 819 & 81.0\% & 0.593 & [0.536, 0.651] \\
\bottomrule
\end{tabular}
\caption{Per-criterion Cohen's $\kappa$ in an independent RiceChem diagnostic replicate (5-shot, Gemini-3-Flash; 663/819 correct). This run is separate from the paired rerun. Reported 95\% bootstrap confidence intervals use 1{,}000 decision-level resamples; pooled summaries are descriptive because resampling is not clustered by student response. All criteria are binary (MET/UNMET).}
\label{tab:ricechem-per-criterion-kappa}
\end{table}

\FloatBarrier
\subsection{Weight inference}

The raw data does not encode per-criterion point values. We infer weights by solving a least-squares regression per question:
\begin{equation}
\text{Score}_i - \text{Adjustment}_i = \sum_{j=1}^{m} w_j \cdot \mathbf{1}[\text{criterion}_j = \text{TRUE}]
\end{equation}
where the \texttt{Adjustment} column captures manual TA score modifications. The inferred weights cluster around integer values (1 or 2 points per criterion), consistent with the rubric design. Table~\ref{tab:ricechem-r2} reports goodness-of-fit.

\begin{table}[ht]
\centering
\small
\begin{tabular}{lcc}
\toprule
\textbf{Question} & $R^2$ & \textbf{Inferred Weight Range} \\
\midrule
Q1 & 0.994 & 0.95 -- 1.97 \\
Q2 & 0.986 & 0.96 -- 2.04 \\
Q3 & 0.542 & 1.00 -- 2.00 \\
Q4 & 0.985 & 0.98 -- 2.00 \\
\bottomrule
\end{tabular}
\caption{Weight inference quality per question. Q3's lower $R^2$ is attributable to annotation artifacts (see text).}
\label{tab:ricechem-r2}
\end{table}

Q3's lower $R^2$ (0.542) has two identified causes. First, 13 of 15 students flagged with \texttt{Correct response = TRUE} received full marks (9 points) but have all individual criteria marked \texttt{FALSE}; these were graded holistically rather than criterion-by-criterion. Because \texttt{Correct response} is a summary flag (not a rubric criterion), it is excluded from the converted rubric, leaving these 13 responses with ground truth labels that undercount their actual performance. Second, 51 of the 298 released converted Q3 responses (17.1\%) carry manual \texttt{Adjustment} values of $\pm 1$ or $\pm 2$ points, introducing variance not captured by the binary criteria alone. For Q1, Q2, and Q4, the near-perfect $R^2$ values ($\geq 0.985$) confirm that the inferred weights faithfully recover the original scoring rubric.

\FloatBarrier
\subsection{Score distributions}

\begin{table}[ht]
\centering
\small
\begin{tabular}{lcccc}
\toprule
\textbf{Question} & \textbf{Mean} & \textbf{Std Dev} & \textbf{Min} & \textbf{Max} \\
\midrule
Q1 & 5.09 & 2.10 & 0.0 & 7.9 \\
Q2 & 3.92 & 2.20 & 0.0 & 7.9 \\
Q3 & 4.18 & 2.60 & 0.0 & 9.0 \\
Q4 & 6.65 & 1.75 & 0.0 & 7.9 \\
\bottomrule
\end{tabular}
\caption{RiceChem score distributions per question. Scores are $\sum_j w_j \cdot \mathbf{1}[\text{criterion}_j = \text{MET}]$ using inferred weights. Q4 has the highest average; Q2 is the most challenging, with several criteria below 30\% MET rate.}
\label{tab:ricechem-scores}
\end{table}

\begin{figure}[ht]
    \centering
    \includegraphics[width=0.9\linewidth]{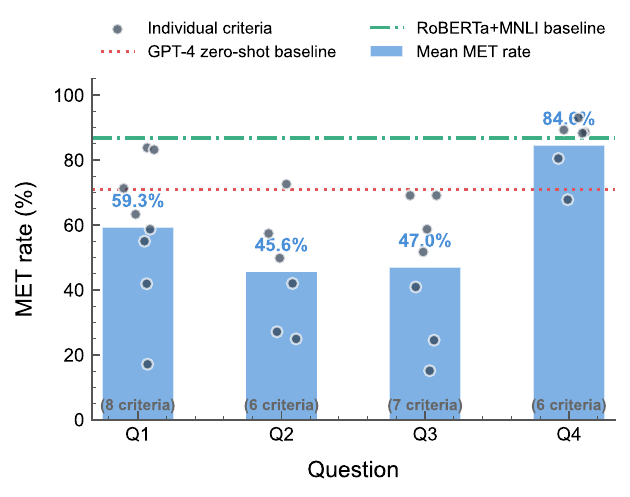}
    \caption{Mean MET rate by question on RiceChem with individual criterion rates overlaid.}
    \label{fig:ricechem-per-question}
\end{figure}

\begin{figure}[ht]
    \centering
    \includegraphics[width=0.9\linewidth]{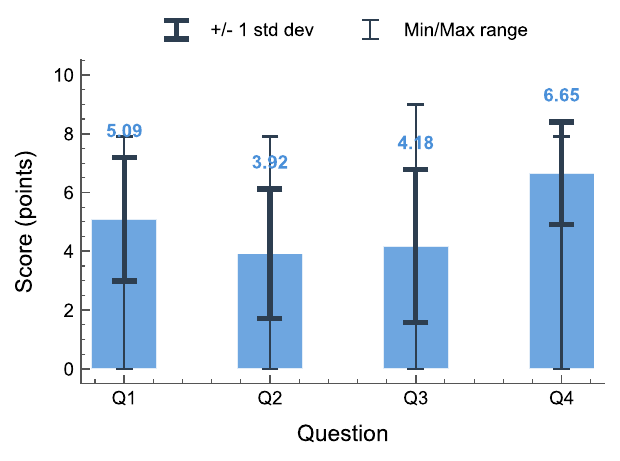}
    \caption{Student score distributions on RiceChem.}
    \label{fig:ricechem-score-distributions}
\end{figure}

\begin{figure}[ht]
    \centering
    \includegraphics[width=0.9\linewidth]{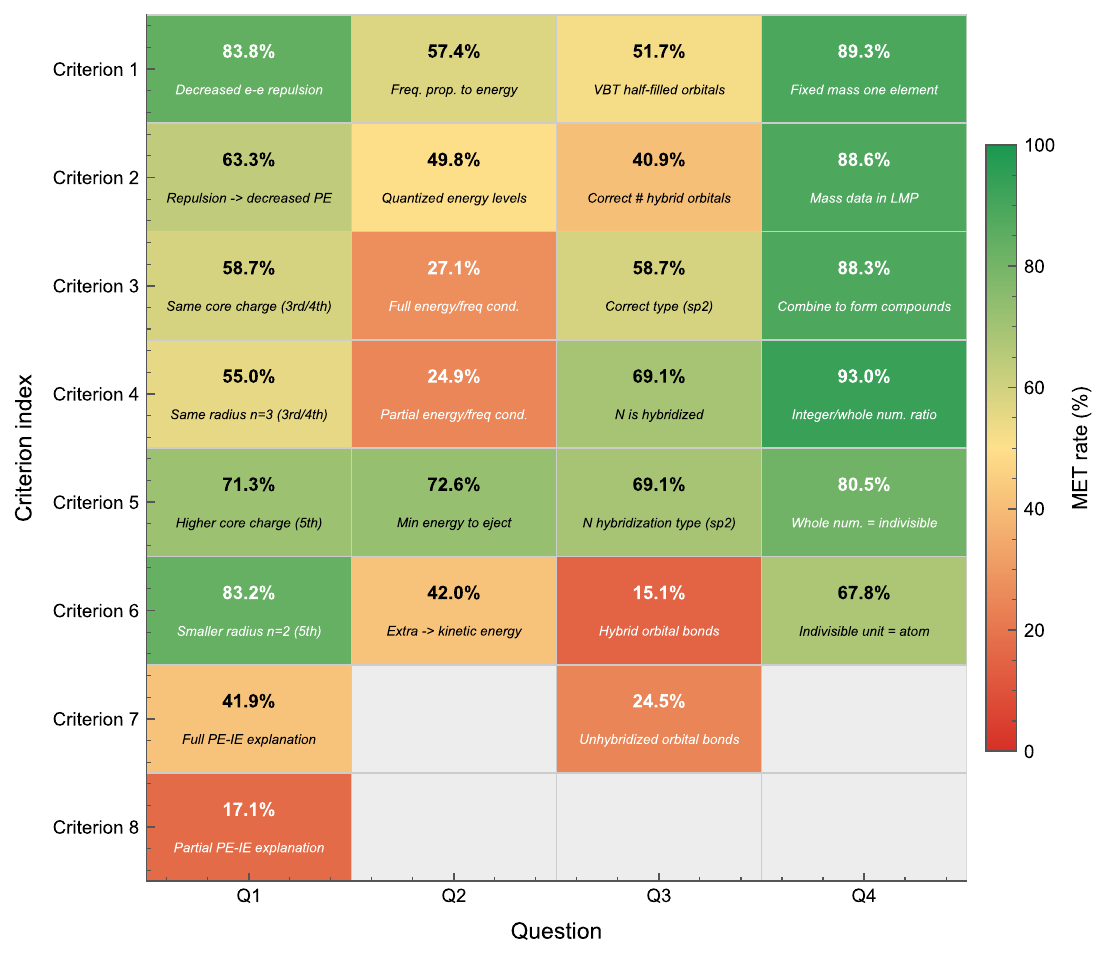}
    \caption{MET rates across all 27 RiceChem criteria organized by question.}
    \label{fig:ricechem-met-heatmap}
\end{figure}

\FloatBarrier
\subsection{Paired comparison and exploratory few-shot sweep}

The primary paired rerun compares 0-shot and 5-shot predictions on the same 819 held-out response--criterion decisions: accuracy increases from 78.0\% (639/819) to 80.7\% (661/819), with 54 decisions improving and 32 degrading. Because each response contributes multiple correlated criterion decisions, these counts and per-question differences are descriptive; we do not claim significance from the unclustered decision-level comparison.

Table~\ref{tab:ricechem-fewshot} instead reports the original exploratory sweep across five shot counts. It is a separate stochastic run, so its 0-shot and 5-shot accuracies (77.2\% and 80.0\%) should not be combined with the paired test above. In both runs, few-shot exemplars come exclusively from the 80\% training split with verdict-balanced sampling; no exemplar appears in the 10\% test split, and the fixed train--test partition is shared across shot counts.

\begin{table}[ht]
\centering
\small
\begin{tabular}{ccc}
\toprule
\textbf{Shots} & \textbf{Accuracy} & \textbf{Cost (USD)} \\
\midrule
0  & 77.2\% & \$0.51 \\
3  & 79.0\% & \$0.77 \\
5  & 80.0\% & \$0.92 \\
10 & 79.7\% & \$0.84 \\
20 & 80.8\% & \$1.07 \\
\bottomrule
\end{tabular}
\caption{Original exploratory RiceChem few-shot sweep. These values come from a separate stochastic run from the paired 0-shot/5-shot comparison. The non-monotonic observed costs reflect run-to-run cache-hit variation.}
\label{tab:ricechem-fewshot}
\end{table}

\begin{figure}[ht]
    \centering
    \includegraphics[width=0.7\linewidth]{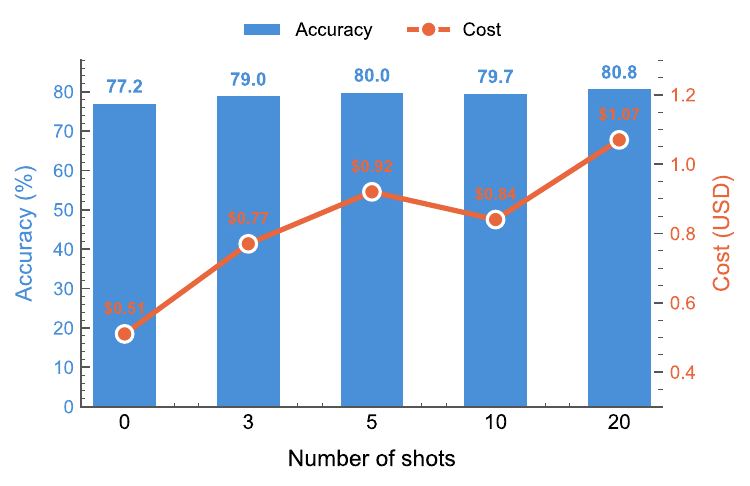}
    \caption{Accuracy versus cost in the original exploratory RiceChem few-shot sweep, a separate run from the paired comparison.}
    \label{fig:fewshot-ricechem}
\end{figure}

\FloatBarrier
\subsection{Cold-start results}

Following \citet{sonkar2024automatedlonganswergrading}, Table~\ref{tab:ricechem-coldstart} reports cold-start performance where models are trained on three questions and tested on the held-out fourth. This simulates rubric application to a new question type.

\begin{table}[ht]
\centering
\small
\begin{tabular}{lcc}
\toprule
\textbf{Held-out Question} & \textbf{Accuracy} & \textbf{F1} \\
\midrule
Q1 & 65.9\% & 0.705 \\
Q2 & 68.7\% & 0.629 \\
Q3 & 66.7\% & 0.633 \\
Q4 & 60.6\% & 0.717 \\
\bottomrule
\end{tabular}
\caption{Cold-start results from \citet{sonkar2024automatedlonganswergrading} (RoBERTa+MNLI, train on 3 questions, test on held-out 4th). These represent supervised transfer learning performance for comparison with zero/few-shot approaches.}
\label{tab:ricechem-coldstart}
\end{table}

\FloatBarrier
\subsection{Prior results from Sonkar et al.}

For reference, Table~\ref{tab:ricechem-prior} reproduces the in-distribution results from \citet{sonkar2024automatedlonganswergrading}, evaluated on the rubric entailment task with an 80-10-10 split. Fine-tuned model results are averaged across 5 random seeds; metrics are micro-averaged across all criterion--response pairs in the test set. The NLI-based approach (RoBERTa-large + MNLI) substantially outperforms zero-shot GPT-4; both approaches treat grading as pairwise entailment between student response and rubric criterion.

\begin{table}[ht]
\centering
\small
\begin{tabular}{lcc}
\toprule
\textbf{Model} & \textbf{Accuracy} & \textbf{F1} \\
\midrule
RoBERTa-large + MNLI & 86.8\% & 0.888 \\
BART-large + MNLI & 85.4\% & 0.876 \\
RoBERTa-large & 84.1\% & 0.864 \\
GPT-4 (zero-shot) & 70.9\% & 0.689 \\
\bottomrule
\end{tabular}
\caption{In-distribution results from \citet{sonkar2024automatedlonganswergrading} on RiceChem rubric entailment. NLI pretraining provides a substantial boost over standard fine-tuning, and both supervised approaches outperform zero-shot GPT-4.}
\label{tab:ricechem-prior}
\end{table}

\FloatBarrier
\section{ResearcherBench additional results}
\label{appendix:researcherbench}

\subsection{Dataset description}

ResearcherBench~\citep{xu2025researcherbenchevaluatingdeepai} is a benchmark for evaluating deep AI research systems on genuinely open-ended, frontier scientific questions that require synthesis across papers, expert judgment, and nuanced reasoning. The benchmark contains 65 expert-curated questions across 34 AI research subjects, with responses from 11 AI systems (7 evaluated in the original paper, 4 additional from the public repository).

Table~\ref{tab:rb-question-categories} summarizes the three question categories. Open Consulting questions, which comprise roughly half the benchmark, require subjective interpretation and strategic insight, qualities that make automated evaluation particularly challenging.

\begin{table}[ht]
\centering
\small
\begin{tabular}{lcp{8cm}}
\toprule
\textbf{Category} & \textbf{Count} & \textbf{Description} \\
\midrule
Open Consulting   & 33 & Emerging trends, strategic insights, subjective interpretation and expert judgment \\
Literature Review  & 20 & Synthesizing findings across multiple papers, comparing methodologies, identifying trends \\
Technical Details  & 12 & Precise explanations of methodologies, implementations, or theoretical concepts \\
\bottomrule
\end{tabular}
\caption{ResearcherBench question categories. Open Consulting questions, which require the most subjective judgment, form the majority of the benchmark.}
\label{tab:rb-question-categories}
\end{table}

\subsection{Rubric structure}

Each question carries its own per-item rubric of weighted binary criteria. There is no global rubric; all criteria are question-specific. Rubrics were designed by experienced AI researchers (masters and PhD students) through a three-step process: insight extraction from reference materials, human annotation of criteria, and quality control review.

Each criterion has an integer weight from 1 to 3 reflecting its importance: weight 1 for nice-to-have details (35\% of criteria), weight 2 for supporting points (51\%), and weight 3 for core findings (14\%).

Table~\ref{tab:rb-rubric-stats} reports summary statistics for criteria counts across the 65 questions.

\begin{table}[ht]
\centering
\small
\begin{tabular}{lc}
\toprule
\textbf{Statistic} & \textbf{Value} \\
\midrule
Total criteria & 931 \\
Mean per question & 14.3 \\
Median per question & 14 \\
Standard deviation & 3.1 \\
Minimum & 6 \\
Maximum & 21 \\
Interquartile range & 12--16.5 \\
\bottomrule
\end{tabular}
\caption{Rubric criteria statistics across the 65 ResearcherBench questions. All criteria are binary (MET/UNMET).}
\label{tab:rb-rubric-stats}
\end{table}

\subsection{Coverage score}

The ResearcherBench coverage score is defined as:
\begin{equation}
\text{Coverage} = \frac{\sum_{i} w_i \cdot c_i}{\sum_{i} w_i}, \quad c_i \in \{0, 1\}
\end{equation}
where $w_i$ is the criterion weight and $c_i$ indicates whether criterion $i$ is met. This is identical to \texttt{Autorubric}'s normalized \texttt{EvaluationReport.score} calculation for binary criteria with positive weights (Equation~\ref{eq:score-aggregation}), so no score transformation is needed when using \texttt{Autorubric} to evaluate ResearcherBench. The \texttt{raw\_score} field instead stores the unnormalized weighted sum.

\subsection{Systems evaluated}

We evaluate three Deep AI Research Systems from the benchmark:

\begin{itemize}
    \item \textbf{OpenAI DeepResearch}: OpenAI's deep research agent
    \item \textbf{Gemini DeepResearch}: Google's deep research system (Gemini-2.5-Pro)
    \item \textbf{Grok3 DeepSearch}: xAI's deep search system
\end{itemize}

The original benchmark includes additional systems (Grok3 DeeperSearch, Perplexity Deep Research, GPT-4o Search Preview, Sonar Reasoning Pro) and four systems from the public repository (Claude, Doubao, Mita, Perplexity Sonar). We selected the three popular systems above to keep the evaluation costs reasonable.

\subsection{Cross-judge agreement analysis}

With only three systems, aggregate rank correlations are degenerate (Spearman $\rho = 1.0$, Kendall $\tau = 1.0$ between our two judges). Per-question analysis provides a more informative picture.

\paragraph{Per-question rank concordance.} For each of the 65 questions, we rank the three systems by coverage score under each judge and measure agreement. The two judges agree on the full three-system ranking for only 15 of 65 questions (23.1\%). They agree on the top-ranked system 57\% of the time and the bottom-ranked system 52\% of the time---better than the 33\% expected by chance, but far from unanimous.

\paragraph{Per-question Spearman.} Computing Spearman $\rho$ between judges across the three systems on each question yields a mean $\rho = 0.433$ [95\% bootstrap CI: 0.266, 0.584] over 61 questions with non-degenerate rankings. Exact reverse rankings ($\rho = -1$) occur on 3 of these 61 questions (4.9\%).

\paragraph{Cross-judge score correlations.} A complementary view asks whether judges agree on \emph{which questions are hard} for a given system, correlating the 65 per-question scores across judges. These Spearman correlations are moderate to strong: $\rho = 0.71$ for OpenAI, $\rho = 0.54$ for Gemini, and $\rho = 0.82$ for Grok3 (all $p < 0.001$). Judges thus agree more on question difficulty than on system ordering.

\paragraph{Cross-study comparison.} Alignment with \citet{xu2025researcherbenchevaluatingdeepai}'s Sonnet-3.5 assessment yields Spearman $\rho = 0.5$ and Kendall $\tau = 0.33$, with the disagreement arising from a reversal of the top two systems---consistent with the overlapping bootstrap CIs reported in Table~\ref{tab:researcherbench-results}.

\subsection{Paired permutation test for system ranking}

Because bootstrap CIs overlap for the top two systems (Table~\ref{tab:researcherbench-results}), we apply a two-sided paired permutation test to the 65 per-question score differences (Gemini DeepResearch minus OpenAI DeepResearch). Table~\ref{tab:rb-paired-permutation} reports results for each judge.

\begin{table}[ht]
\centering
\small
\begin{tabular}{lcccc}
\toprule
\textbf{Judge} & \textbf{Mean $\Delta$} & \textbf{95\% CI} & \textbf{Cohen's $d$} & \textbf{$p$ (perm.)} \\
\midrule
Sonnet-4.5 & +0.073 & [0.027, 0.117] & 0.39 & 0.003* \\
Gemini-3-Flash & +0.039 & [-0.025, 0.097] & 0.16 & 0.219 \\
\bottomrule
\end{tabular}
\caption{Paired permutation test for Gemini DeepResearch vs.\ OpenAI DeepResearch coverage scores (65 questions, 9{,}999 permutations). $\Delta$ = Gemini DR $-$ OpenAI DR. $^*p < 0.05$.}
\label{tab:rb-paired-permutation}
\end{table}

Under Sonnet-4.5, the difference is significant ($p = 0.003$, $d = 0.39$), with the bootstrap CI excluding zero. Under Gemini-3-Flash, the CI includes zero and the test is not significant ($p = 0.219$). The top-two ranking is therefore judge-dependent: Sonnet-4.5 separates the systems, while Gemini-3-Flash does not distinguish them at the 5\% level.

\subsection{Per-criterion agreement by weight band}

To assess whether judges agree more on high-importance criteria, we compare criterion-level verdicts across judges for all 2{,}793 paired judgments (931 criteria $\times$ 3 systems). Table~\ref{tab:rb-weight-agreement} reports raw agreement and Cohen's $\kappa$ by criterion weight band.

\begin{table}[ht]
\centering
\small
\begin{tabular}{l cc cc cc cc}
\toprule
& \multicolumn{2}{c}{\textbf{W=1}} & \multicolumn{2}{c}{\textbf{W=2}} & \multicolumn{2}{c}{\textbf{W=3}} & \multicolumn{2}{c}{\textbf{Overall}} \\
\cmidrule(lr){2-3} \cmidrule(lr){4-5} \cmidrule(lr){6-7} \cmidrule(lr){8-9}
\textbf{System} & \textbf{Agr.} & \textbf{$\kappa$} & \textbf{Agr.} & \textbf{$\kappa$} & \textbf{Agr.} & \textbf{$\kappa$} & \textbf{Agr.} & \textbf{$\kappa$} \\
\midrule
OpenAI DeepResearch & 71.2\% & 0.412 & 80.2\% & 0.542 & 81.9\% & 0.558 & 77.2\% & 0.497 \\
Gemini DeepResearch & 73.0\% & 0.417 & 76.6\% & 0.389 & 77.2\% & 0.314 & 75.4\% & 0.398 \\
Grok3 DeepSearch & 78.5\% & 0.569 & 85.9\% & 0.707 & 83.5\% & 0.644 & 82.9\% & 0.654 \\
\midrule
\textbf{Pooled} & 74.2\% & 0.477 & 80.9\% & 0.566 & 80.8\% & 0.532 & 78.5\% & 0.532 \\
\bottomrule
\end{tabular}
\caption{Per-weight-band inter-judge agreement (Sonnet-4.5 vs.\ Gemini-3-Flash) on ResearcherBench criterion verdicts. W=1: nice-to-have ($N=990$), W=2: supporting ($N=1{,}422$), W=3: core ($N=381$). Agreement is raw percentage; $\kappa$ is Cohen's kappa.}
\label{tab:rb-weight-agreement}
\end{table}

\begin{figure}[ht]
    \centering
    \includegraphics[width=0.9\linewidth]{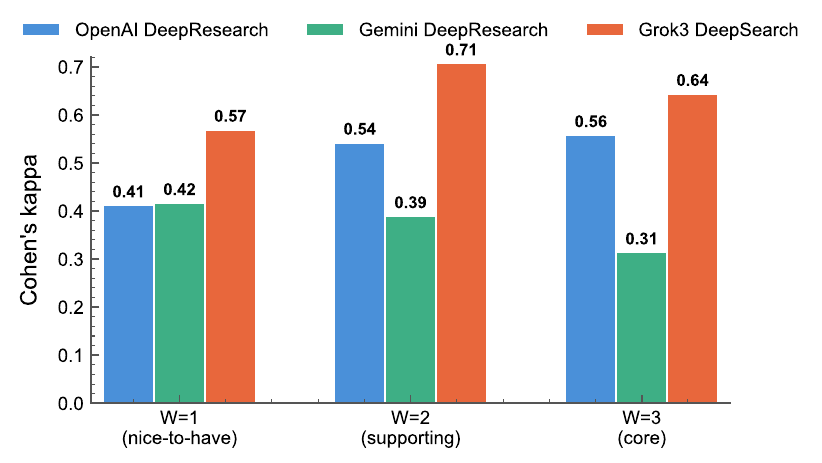}
    \caption{Cohen's $\kappa$ by criterion weight band and system on ResearcherBench. Grok3 (clearly weakest system) shows highest inter-judge agreement; Gemini DeepResearch (top-ranked system) shows lowest, especially on core criteria.}
    \label{fig:rb-weight-agreement}
\end{figure}

Weight-1 (nice-to-have) criteria show the lowest agreement ($\kappa = 0.477$), while weight-2 and weight-3 criteria are comparable ($\kappa = 0.566$ and $0.532$). Across systems, Grok3 shows the highest agreement ($\kappa = 0.654$), consistent with being the clearly weakest system where both judges identify the same unmet criteria. Gemini DeepResearch shows the lowest agreement ($\kappa = 0.398$), particularly on core criteria ($\kappa = 0.314$), suggesting the top system is hardest to evaluate consistently. The disagreement direction is asymmetric: of 600 disagreements, 452 (75\%) are cases where Gemini-3-Flash marks MET while Sonnet-4.5 marks UNMET, consistent with Gemini-3-Flash's higher-scoring tendency reported in Section~\ref{appendix:rb-calibration}. Figure~\ref{fig:rb-weight-agreement} visualizes the per-system kappa by weight band.

\FloatBarrier
\subsection{Disagreement taxonomy}

To examine whether disagreements concentrate in particular rubric constructs, we classify each criterion's requirement text into non-exclusive categories using keyword patterns (Table~\ref{tab:rb-disagreement-taxonomy}).

\begin{table}[ht]
\centering
\small
\begin{tabular}{lcccc}
\toprule
\textbf{Rubric construct} & \textbf{$N$} & \textbf{Disagree} & \textbf{Rate} & \textbf{$\Delta$} \\
\midrule
Quantitative & 30 & 11 & 36.7\% & +15.2\% \\
Critical analysis & 513 & 143 & 27.9\% & +6.4\% \\
Other & 183 & 42 & 23.0\% & +1.5\% \\
Mention & 1{,}107 & 238 & 21.5\% & +0.0\% \\
Explanation & 2{,}061 & 435 & 21.1\% & $-$0.4\% \\
Comparison & 597 & 118 & 19.8\% & $-$1.7\% \\
Enumeration & 27 & 4 & 14.8\% & $-$6.7\% \\
Depth & 84 & 12 & 14.3\% & $-$7.2\% \\
\midrule
\textbf{Overall} & 2{,}793 & 600 & 21.5\% & --- \\
\bottomrule
\end{tabular}
\caption{Disagreement rates by rubric construct type (pooled across 3 systems, 931 criteria $\times$ 3 = 2{,}793 judgments). Categories are non-exclusive. $\Delta$ is the difference from the overall disagreement rate. Quantitative criteria ($N=30$) and critical-analysis criteria show elevated disagreement; enumeration and depth criteria show reduced disagreement.}
\label{tab:rb-disagreement-taxonomy}
\end{table}

Critical-analysis criteria---those requiring evaluation of strengths, limitations, or assessment quality---show the highest disagreement rate among well-represented categories (27.9\% vs.\ 21.5\% overall). Quantitative-threshold criteria show even higher disagreement (36.7\%) but on a small sample ($N=30$). In contrast, criteria matching enumeration and depth keywords show the lowest disagreement rates (14--15\%). Because the categories are non-exclusive and keyword-derived, this analysis is descriptive and does not establish that rewriting criteria would improve reliability.

\subsection{Cross-judge score differences}
\label{appendix:rb-calibration}

Gemini-3-Flash assigns higher mean coverage scores than Sonnet-4.5 for all three systems. The differences (Gemini-3-Flash minus Sonnet-4.5) are $+0.151$ for OpenAI DeepResearch, $+0.118$ for Gemini DeepResearch, and $+0.039$ for Grok3 DeepSearch, with an equal-weight mean difference of $+0.103$. The consistently positive direction indicates a higher-scoring tendency for Gemini-3-Flash in this evaluation. However, the varying magnitudes leave open judge-by-system interactions, and without independent criterion adjudication these differences do not establish which judge is better calibrated.

\subsection{Cost analysis}

Evaluation cost varies substantially across both judge models and evaluated systems. Table~\ref{tab:rb-cost-analysis} summarizes the cost comparison.

\begin{table}[ht]
\centering
\small
\begin{tabular}{lcc}
\toprule
\textbf{System} & \textbf{Sonnet-4.5} & \textbf{Gemini-3-Flash} \\
\midrule
OpenAI DeepResearch & \$37.98 & \$6.72 \\
Gemini DeepResearch & \$61.23 & \$9.77 \\
Grok3 DeepSearch    & \$12.37 & \$1.99 \\
\bottomrule
\end{tabular}
\caption{Evaluation cost (USD) per system and judge. Gemini-3-Flash is 5.7--6.3$\times$ cheaper than Sonnet-4.5. Cost differences across systems reflect differences in response length.}
\label{tab:rb-cost-analysis}
\end{table}

The total cost for evaluating all three systems across all 65 questions with both judges is \$130.06 (5,586 criterion-level judgments). Gemini DeepResearch is the most expensive to evaluate because its responses are the longest, requiring more tokens per criterion evaluation. Grok3 DeepSearch is the cheapest for the same reason.

\begin{figure}[ht]
    \centering
    \includegraphics[width=0.9\linewidth]{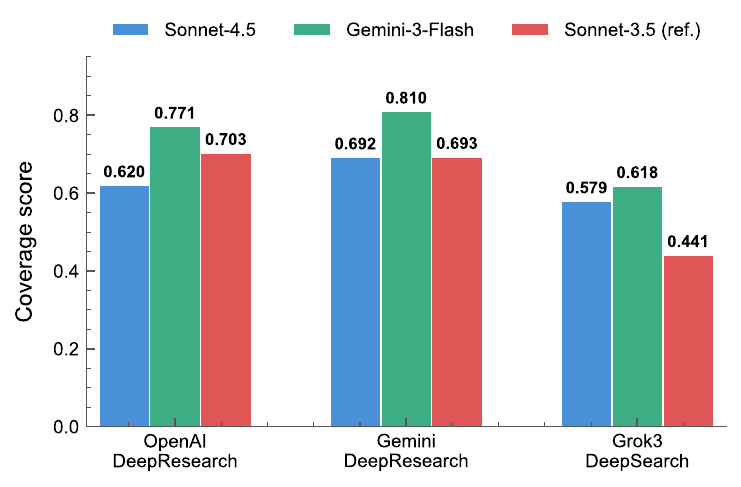}
    \caption{Cross-judge coverage scores on ResearcherBench. Both \texttt{Autorubric} judges produce the same aggregate ranking despite positive cross-judge score differences, but bootstrap CIs overlap for the top two systems (Table~\ref{tab:researcherbench-results}).}
    \label{fig:rb-cross-judge}
\end{figure}

\begin{figure}[ht]
    \centering
    \includegraphics[width=0.9\linewidth]{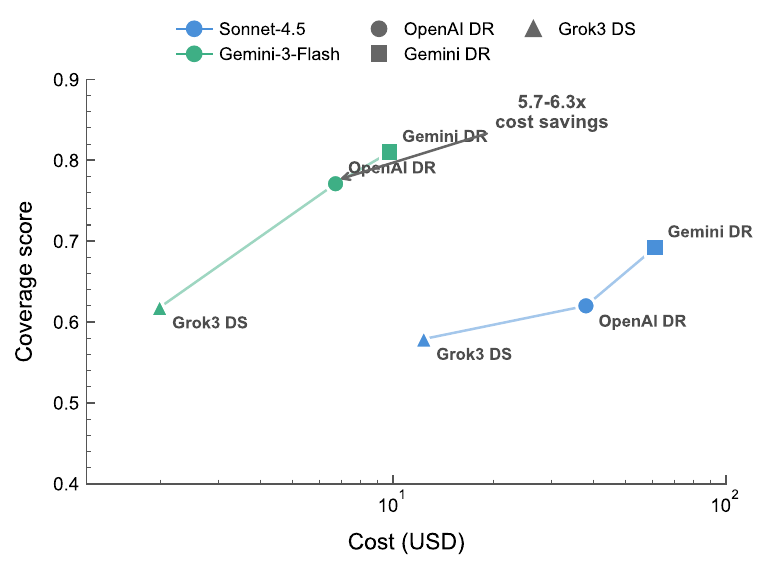}
    \caption{Cost vs.\ coverage on ResearcherBench. At 5.7--6.3$\times$ lower cost, Gemini-3-Flash assigns higher coverage scores while preserving the aggregate system ranking; no independent adjudication establishes which judge is better calibrated.}
    \label{fig:rb-cost-performance}
\end{figure}

\begin{figure}[ht]
    \centering
    \includegraphics[width=0.9\linewidth]{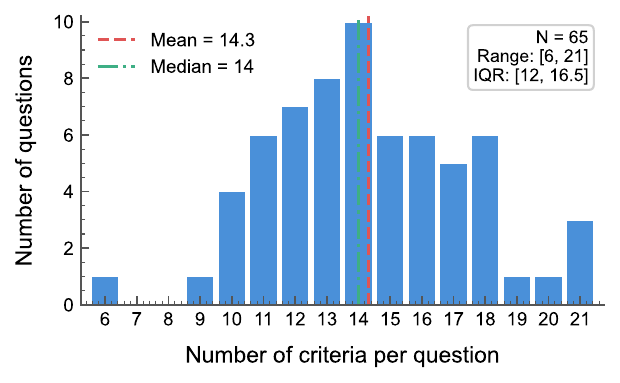}
    \caption{Distribution of criteria counts per question in ResearcherBench (65 questions).}
    \label{fig:rb-criteria-distribution}
\end{figure}

\FloatBarrier
\subsection{Rubric statistics}

Across the 65 questions, criteria counts range from 6 to 21 (mean 14.3, median 14, SD 3.1; IQR 12--16.5). Weight distribution: 35\% weight-1 (nice-to-have), 51\% weight-2 (supporting), 14\% weight-3 (core). All criteria are binary.

\FloatBarrier
\section{CHARM-100 dataset}
\label{appendix:charm100-dataset}

This appendix provides the full specification of the CHARM-100 (\textbf{Ch}atbot \textbf{A}ssessment with Mixed \textbf{R}ubric \textbf{M}etrics) dataset introduced in Section~\ref{sec:charm100}. The benchmark evaluates LLM-as-a-judge systems on their ability to apply heterogeneous rubric criteria simultaneously.

\subsection{Motivation}

Several chatbot evaluation benchmarks exist, but they uniformly adopt a single scale type across all criteria. MT-Bench~\citep{zheng2023judging} and Vicuna Bench use 1--5 Likert scales; Arena-Hard-Auto~\citep{li2025arenahard} uses pairwise ordinal judgments; WildBench~\citep{lin2024wildbench} uses binary checklists aggregated into a composite score; and HelpSteer2~\citep{wang2024helpsteer2} and LLM-Rubric~\citep{hashemi2024llmrubric} use multi-dimensional Likert ratings. Real-world evaluation rubrics rarely consist of a single criterion type. A benchmark restricted to one measurement scale gives an incomplete picture of judge capabilities: it cannot reveal whether a judge handles binary verification differently from ordinal placement, or whether nominal classification produces different error patterns than either.

The CHARM-100 dataset was created to fill this gap and to exercise the capabilities of rubric evaluation frameworks that support heterogeneous criterion types. Because the rubric contains binary, ordinal, and nominal criteria in a single evaluation pass, it enables criterion-type-specific error analysis and type-appropriate agreement metrics (weighted kappa for ordinal, unweighted kappa for nominal and binary).

\subsection{Dataset overview}

\begin{table}[ht]
\centering
\small
\begin{tabular}{@{}ll@{}}
\toprule
\textbf{Property} & \textbf{Value} \\
\midrule
Name & \texttt{charm-100} \\
Size & 100 annotated samples \\
Format & JSON (\texttt{RubricDataset} schema) \\
Language & English \\
Source & Synthetically authored chatbot conversations \\
Criteria per sample & 6 \\
Criterion types & 1 binary, 4 ordinal, 1 nominal \\
Unique system prompts & 100 \\
Unique topic domains & 35+ \\
\bottomrule
\end{tabular}
\caption{CHARM-100 dataset overview.}
\label{tab:charm100-overview}
\end{table}

Each sample is a single-turn chatbot conversation (one system message, one user message, one assistant response) with synthetically assigned reference labels for all six criteria. In the tables and code, ``ground truth'' denotes this dataset field; it should be read as the operational reference annotation rather than independently adjudicated truth.

\subsection{Annotation schema}
\label{appendix:charm100-schema}

The rubric combines ordinal, nominal, and binary criteria. Each sample receives exactly six ground truth labels, one per criterion. Table~\ref{tab:charm100-full-criteria} provides the full schema.

\begin{table}[ht]
\centering
\small
\begin{tabularx}{\linewidth}{@{}llccX@{}}
\toprule
\textbf{Criterion} & \textbf{Type} & \textbf{Weight} & \textbf{Scale} & \textbf{Options (value)} \\
\midrule
satisfaction & Ordinal & 10.0 & 4-pt & V.\ dissatisfied (0.0), S.\ dissatisfied (0.33), \\
 & & & & S.\ satisfied (0.67), V.\ satisfied (1.0) \\
\addlinespace
helpfulness & Ordinal & 8.0 & 4-pt & Not helpful (0.0), Slightly (0.33), \\
 & & & & Moderately (0.67), Very (1.0) \\
\addlinespace
naturalness & Ordinal & 5.0 & 4-pt & Robotic (0.0), Mechanical (0.33), \\
 & & & & Mostly natural (0.67), Very natural (1.0) \\
\addlinespace
response\_length & Nominal & 4.0 & 3-class & Too brief (0.0), Too verbose (0.0), \\
 & & & & Just right (1.0) \\
\addlinespace
factual\_accuracy & Binary & 10.0 & MET/UNMET & MET (1.0), UNMET (0.0) \\
\addlinespace
specificity & Ordinal & 6.0 & 5-pt & V.\ vague (0.0), S.\ vague (0.33), \\
 & & & & Mod.\ specific (0.67), V.\ specific (1.0), N/A \\
\bottomrule
\end{tabularx}
\caption{Full CHARM-100 annotation schema. The nominal criterion response\_length assigns value 0.0 to both failure modes (too brief, too verbose). The specificity criterion includes an N/A option excluded from scoring.}
\label{tab:charm100-full-criteria}
\end{table}

\textbf{Scoring semantics.} The four ordinal criteria are scored using the \texttt{value} field (0.0--1.0) associated with each option. Because these criteria have a natural ordering, inter-rater agreement is measured with weighted (quadratic) kappa. The nominal criterion \texttt{response\_length} has two failure modes that both receive value 0.0 and one success mode at 1.0; because the failure modes have no ordering relative to each other, agreement is measured with unweighted Cohen's kappa. The binary criterion uses MET/UNMET verdicts scored as 1.0 and 0.0. The \texttt{specificity} criterion includes an N/A option (flagged with \texttt{na: true}) for questions that do not call for concrete recommendations (e.g., definitional or philosophical questions). Samples labeled N/A are excluded from the specificity score denominator under the configured \texttt{SKIP} strategy. The final score is a weighted sum of per-criterion scores, normalized by total positive weight (all weights in this rubric are positive). The reported March 2026 runs predate v1.5.3's automatic N/A-option injection; to reproduce their original label space with v1.5.3, set \texttt{auto\_na\_option=False}, leaving only specificity's author-defined N/A option.

\subsection{Design principles}
\label{appendix:charm100-design}

The dataset was constructed according to four principles.

\textbf{Description-derived quality strata.} The released JSON does not contain an explicit quality-tier field. For the second-annotation audit, the recorded generator deterministically maps description prefixes and failure keywords into five quality strata plus an Edge Case stratum (Table~\ref{tab:charm100-tiers}). These post hoc tags support stratified sampling; they are not independently authored annotations.

\begin{table}[ht]
\centering
\small
\begin{tabular}{@{}lcp{7.5cm}@{}}
\toprule
\textbf{Stratum} & \textbf{Count} & \textbf{Description} \\
\midrule
Excellent & 15 & All or nearly all top marks. Specific, accurate, well-structured, natural. \\
Good & 12 & Mostly high ratings with one or two minor shortcomings (e.g., slightly verbose or mildly mechanical). \\
Mediocre/Mixed & 27 & Split verdicts across criteria with deliberate conflicts. \\
Below Average & 15 & Generic, mechanical, missing key information, or deflective. Partially addresses the question but falls short on several dimensions. \\
Poor & 11 & Useless, evasive, broken, or containing dangerous misinformation. \\
Edge Cases & 20 & Borderline or debatable labels designed to challenge automated judges. \\
\bottomrule
\end{tabular}
\caption{Description-derived CHARM-100 quality strata produced by the fixed audit classifier ($N=100$). The tags are heuristic metadata used for the second-annotation subset, not an explicit field in the released JSON.}
\label{tab:charm100-tiers}
\end{table}

\textbf{Criteria independence.} Cross-criteria conflicts were deliberately introduced. A response might be factually wrong but naturally written, or technically correct but robotic. These conflicts are designed to expose reliance on a single ``overall quality'' heuristic projected across all criteria.

\textbf{Non-trivial responses.} Even descriptions classified into the lowest-quality strata correspond to substantive text with identifiable flaws, not empty outputs or obviously broken formatting. The benchmark tests identification of subtle problems in realistic-looking text.

\textbf{Diversity.} The 100 samples use 100 unique system prompts across more than 35 topic domains and more than 25 distinct assistant personas. This reduces the risk that a judge's accuracy is inflated by topic-specific heuristics or memorized patterns.

\subsection{Topic and domain coverage}
\label{appendix:charm100-topics}

For descriptive coverage, we assigned each sample post hoc to one of nine
mutually exclusive primary-topic clusters, using the main information need in
the user message rather than the system persona or incidental context. This
annotation is not part of the CHARM-100 grading ground truth and is not used in
any evaluation. The counts below cover all 100 items. Coverage is broad
(6--17 items per cluster; normalized entropy $H/\log 9=0.98$), with no single
topic accounting for more than 17\% of the dataset; it is not intended to
estimate deployment traffic.

% Primary-topic audit of charm100.json in 1-based JSON order.
% Source SHA-256: b0ac1a0b9d5eca2c2d4a9c5d56f11759e8c29cf88d75b524f5f2ba300b08db3c
% Computing & Technology (9): 1,11,12,23,44,50,56,73,86
% Science & Mathematics (14): 3,6,21,35,43,46,53,60,64,69,76,80,82,92
% Health & Wellness (12): 4,13,16,18,24,38,49,57,71,81,89,100
% Home & Practical Skills (17): 2,5,8,10,14,29,41,48,52,59,66,75,78,84,88,93,96
% Finance, Law & Consumer Affairs (9): 7,9,37,61,62,72,74,85,94
% Work & Business (6): 19,22,30,47,51,65
% Humanities, Social Sciences & Education (12): 17,26,27,28,36,58,68,70,79,83,87,97
% Arts, Hobbies & Animal Care (10): 15,20,25,33,34,42,54,55,63,90
% Language, Culture & Social Life (11): 31,32,39,40,45,67,77,91,95,98,99

\begin{table}[ht]
\centering
\small
\begin{tabularx}{\linewidth}{@{}p{3.8cm}Xc@{}}
\toprule
\textbf{Cluster} & \textbf{Domains} & \textbf{Count} \\
\midrule
Computing \& Technology & Programming, ML/AI, cybersecurity, digital privacy, software operations, computer support & 9 \\
Science \& Mathematics & Mathematics, statistics, physics, chemistry, biology, astronomy, environmental science & 14 \\
Health \& Wellness & Medicine, nutrition, mental health, sleep, exercise, rehabilitation, first aid & 12 \\
Home \& Practical Skills & Cooking, baking, home repair, electrical safety, automotive maintenance, gardening & 17 \\
Finance, Law \& Consumer Affairs & Budgeting, saving and investing, taxation, legal rights, major purchases & 9 \\
Work \& Business & Project management, presentations, interviewing, career development, negotiation, entrepreneurship & 6 \\
Humanities, Social Sciences \& Education & History, philosophy, literature, economics, geography, art history, psychology, study and research methods & 12 \\
Arts, Hobbies \& Animal Care & Photography, music and audio production, gaming, crafts, creative writing, pets, aquariums & 10 \\
Language, Culture \& Social Life & Travel, etiquette, language learning and translation, media literacy, parenting, relationships & 11 \\
\bottomrule
\end{tabularx}
\caption{Post hoc primary-topic coverage of CHARM-100 ($N=100$). Each item receives one cluster; counts sum to 100. The largest cluster contains 17\% of items ($H/\log 9=0.98$).}
\label{tab:charm100-topics}
\end{table}

\subsection{Ground truth label distributions}
\label{appendix:charm100-distributions}

Reference labels are broadly distributed for the four ordinal criteria, whose majority-class baselines are 27--36\% and normalized entropies are 0.93--0.98. Response length and factual accuracy are intentionally more skewed, with majority baselines of 66\% and 72\%, respectively. CHARM-100 therefore spans relatively balanced and imbalanced criteria rather than being uniformly balanced across all six.

\begin{table}[ht]
\centering
\small
\begin{tabularx}{\linewidth}{@{}lXcc@{}}
\toprule
\textbf{Criterion} & \textbf{Labels (count)} & \textbf{Majority \%} & \textbf{Norm.\ Entropy} \\
\midrule
satisfaction & V.\ dissat.\ (20), S.\ dissat.\ (33), S.\ sat.\ (28), V.\ sat.\ (19) & 33\% & 0.98 \\
helpfulness & Not (19), Slightly (27), Moderately (34), Very (20) & 34\% & 0.98 \\
naturalness & Robotic (9), Mechanical (25), Mostly (30), Very (36) & 36\% & 0.93 \\
response\_length & Brief (20), Verbose (14), Just right (66) & 66\% & 0.79 \\
factual\_accuracy & MET (72), UNMET (28) & 72\% & 0.86 \\
specificity & V.\ vague (13), S.\ vague (27), Mod.\ (26), Very (25), N/A (9) & 27\% & 0.95 \\
\bottomrule
\end{tabularx}
\caption{Ground truth distributions per criterion. Mean normalized entropy across criteria is 0.92. Naturalness is the most skewed ordinal criterion, leaning toward the upper end; response\_length has the lowest entropy because ``Just right'' accounts for 66\% of this synthetically constructed dataset.}
\label{tab:charm100-gt-distributions}
\end{table}

The satisfaction and helpfulness criteria have the most uniform distributions (normalized entropy 0.98). Naturalness is the most skewed ordinal criterion (0.93): even synthetically authored responses tend to sound at least somewhat natural, making ``Robotic/unnatural'' the hardest label to construct convincingly. Response length has the lowest entropy (0.79), driven by the intentional 66\% prevalence of ``Just right.'' The factual-accuracy split is 72/28; these designed prevalences characterize CHARM-100 and are not estimates of deployment base rates.

\begin{figure}[ht]
    \centering
    \includegraphics[width=0.9\linewidth]{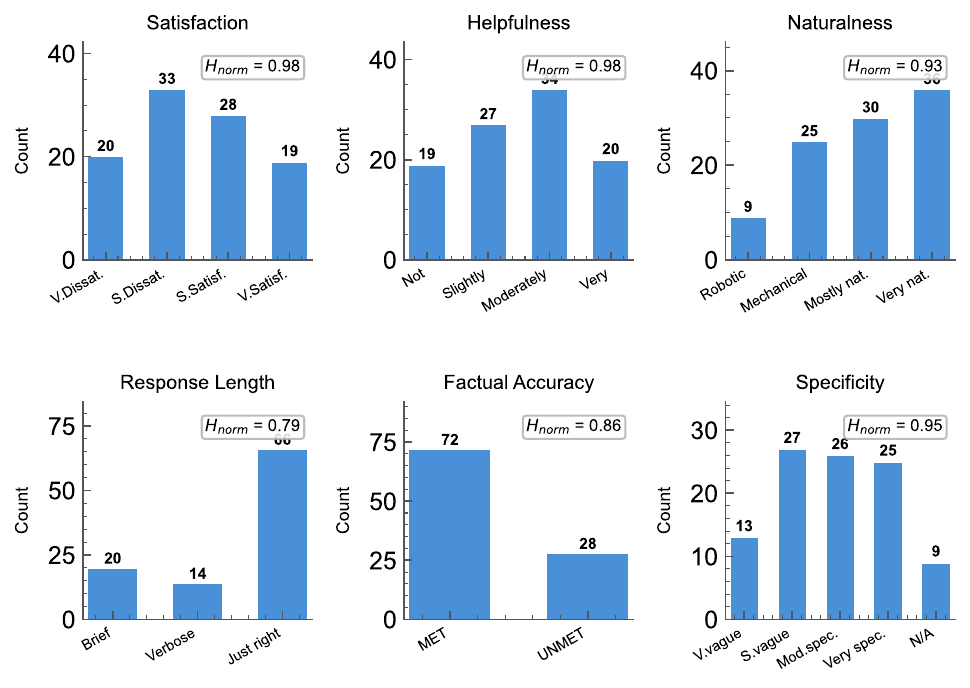}
    \caption{Ground truth label distributions for all six CHARM-100 criteria, with normalized entropy annotated.}
    \label{fig:charm100-gt-distributions}
\end{figure}

\subsection{Response anti-pattern taxonomy}
\label{appendix:charm100-antipatterns}

The dataset includes a structured taxonomy of response failure modes distributed across the description-derived strata. These anti-patterns provide coverage of the failures automated judges must detect.

\textbf{Factual failures.} The 28 UNMET samples for factual accuracy exhibit six patterns: (1) outright errors (confidently stating incorrect facts), (2) subtle hallucinations (mostly correct with one non-obvious error requiring domain knowledge), (3) outdated information, (4) confident confabulation (inventing plausible but fictitious details such as nonexistent laws or fabricated studies), (5) partially correct with critical error (sound reasoning depending on one wrong premise), and (6) misleading framing (individually true statements arranged to support a false conclusion).

\textbf{Helpfulness failures.} Five patterns test whether a judge can distinguish looking informative from being useful: evasive/deflective responses that ask for already-provided information, tangential responses that answer a related but different question, over-hedged responses where qualifiers eliminate all actionable advice, circular responses that restate the question as the answer, and information dumps that provide encyclopedic text without addressing the specific question.

\textbf{Naturalness failures.} Patterns include overly formal/stilted prose inappropriate for conversational settings, excessive disclaimers front-loading paragraphs of caveats, bullet-only walls without connective prose, patronizing tone, and robotic enumeration without context or transitions.

\textbf{Length failures.} Three subcategories: too-brief responses (one-liners for questions requiring detail), too-verbose responses (bloated repetition, unnecessary preamble, kitchen-sink coverage), and preamble-heavy responses (useful content buried under unnecessary context-setting). The third category tests whether a judge evaluates length based on total word count or on the ratio of useful content to filler.

\textbf{Compound anti-patterns.} The most diagnostically valuable samples combine failures across criteria: factually wrong but naturally written, correct but robotic, helpful but excessively verbose, natural but vague, condescending but technically specific, empathetic but without actionable advice. These force the judge to evaluate each criterion independently rather than relying on a single quality signal.

\subsection{Edge cases}
\label{appendix:charm100-edge}

Ten to twenty samples are deliberately ambiguous or borderline, serving as stress tests for automated judges. These fall into several categories:

\begin{itemize}
    \item \textbf{Criteria conflicts}: Factually wrong but well-written, or correct but robotic. Tests whether a judge maintains independent assessments or collapses into a single ``overall quality'' judgment.
    \item \textbf{Factual borderlines}: Mostly correct with one subtle error, reasonable rounding versus wrong numbers, previously-correct-but-now-outdated information.
    \item \textbf{Length borderlines}: Three sentences may be appropriate for a simple question but insufficient for a complex one. Tests whether the judge adjusts length expectations to question complexity.
    \item \textbf{N/A ambiguity}: Questions ranging from clearly definitional (N/A appropriate for specificity) to clearly requiring actionable advice, with a subset at the boundary.
    \item \textbf{Context mismatches}: Correct advice for the wrong context (wrong jurisdiction, wrong climate, wrong cultural setting).
    \item \textbf{Polite refusals}: Appropriate safety redirects that nonetheless fail to answer the question.
    \item \textbf{Satisfaction-helpfulness divergence}: Correct but unsatisfying responses, or entertaining but unhelpful ones.
    \item \textbf{Oversimplification vs.\ accessibility}: Tests the boundary at which simplification becomes misleading.
\end{itemize}

\subsection{Sample format}

Each item in the dataset has three fields: \texttt{submission} (a JSON-encoded string of the messages array), \texttt{description} (a human-readable summary of quality characteristics, not used in scoring), and \texttt{ground\_truth} (an array of six labels in fixed criterion order). All conversations are single-turn: one system message, one user message, one assistant response. User messages have a median length of 19 words. Assistant responses range from 9 to 497 words (mean 167, standard deviation 98).

\subsection{Summary statistics}

\begin{table}[ht]
\centering
\small
\begin{tabularx}{\linewidth}{@{}lX@{}}
\toprule
\textbf{Statistic} & \textbf{Value} \\
\midrule
Total samples & 100 \\
Unique topic domains & 35+ \\
Unique system prompts & 100 \\
User message length (median) & 19 words \\
Assistant response length (min / median / max) & 9 / 180 / 497 words \\
Assistant response length (mean / stdev) & 167 / 98 words \\
Criteria per sample & 6 \\
Ordinal criteria & 4 (satisfaction, helpfulness, naturalness, specificity) \\
Nominal criteria & 1 (response\_length) \\
Binary criteria & 1 (factual\_accuracy) \\
Samples with N/A labels & 9 (specificity only) \\
Mean normalized entropy (across criteria) & 0.92 \\
\bottomrule
\end{tabularx}
\caption{CHARM-100 summary statistics.}
\label{tab:charm100-summary-stats}
\end{table}

\subsection{Second-annotation agreement}
\label{appendix:charm100-iaa}

To audit the stability of the CHARM-100 reference labels, a second annotation pass independently judged a fixed, seed-42, description-stratified subset of 50 items (50\% of the dataset): 8 Excellent, 6 Good, 14 Mediocre/Mixed, 8 Below Average, 6 Poor, and 8 Edge Case. The second pass used the annotation guidelines (criteria definitions and scale options) without access to the reference labels. Table~\ref{tab:charm100-iaa} reports per-criterion Cohen's $\kappa$ between the reference labels and the second annotations.

\begin{table}[ht]
\centering
\small
\begin{tabular}{llccc}
\toprule
\textbf{Criterion} & \textbf{Type} & \textbf{Weighting} & \textbf{$N$} & \textbf{$\kappa$} \\
\midrule
Satisfaction    & Ordinal (4) & Quadratic & 50 & 0.781 \\
Helpfulness     & Ordinal (4) & Quadratic & 50 & 0.777 \\
Naturalness     & Ordinal (4) & Quadratic & 50 & 0.870 \\
Response length & Nominal (3) & Unweighted & 50 & 0.506 \\
Factual accuracy & Binary (2) & Unweighted & 50 & 0.622 \\
Specificity     & Ordinal (4) & Quadratic & 47 & 0.565 \\
\midrule
\textbf{Macro-average} &      &           & --- & \textbf{0.687} \\
\bottomrule
\end{tabular}
\caption{Agreement between the reference labels and a second annotation pass on a fixed description-stratified 50-item subset of CHARM-100. Ordinal criteria use quadratic weighting; binary and nominal criteria use unweighted Cohen's $\kappa$. For specificity, three N/A--N/A pairs are excluded, leaving 47 complete cases and four effective ordered levels. The displayed mean (0.687) is a descriptive macro-average across these differently weighted coefficients.}
\label{tab:charm100-iaa}
\end{table}

Naturalness has the highest agreement ($\kappa = 0.870$), followed by satisfaction ($0.781$) and helpfulness ($0.777$). Response length ($0.506$) and specificity ($0.565$) have the lowest agreement, consistent with the greater subjectivity of these criteria---particularly the boundary between ``just right'' and ``too brief.'' Factual accuracy ($0.622$) falls between these groups; disagreements typically involve borderline cases where one annotation treats an omission as a factual gap and the other does not.

The descriptive macro-average is $\kappa=0.687$. Because it averages quadratic-weighted and unweighted coefficients across different criterion types, it is not a single common-scale reliability statistic and we do not interpret it against a qualitative threshold. The original labels were assigned during synthetic authoring and only half the dataset received a second annotation, so this is an audit of reference-label stability rather than a full-dataset human--human reliability estimate. The recorded subset, label key, guidelines, and summary statistics are included with the paper artifacts in \path{data/charm100_iaa_*}.

\subsection{Limitations}
\label{appendix:charm100-limitations}

\textbf{Synthetic construction.} All conversations were authored by a language model rather than collected from a production chatbot system. The distribution of response patterns, error types, and conversational styles may not fully represent real-world deployments. Synthetic authoring provides precise control over the joint distribution of quality labels but reduces ecological validity.

\textbf{Partial second-annotation coverage.} Agreement was measured on a fixed stratum-sampled 50-item subset (50\% of the dataset), with the subset counts reported above; it is not proportional to the six description-derived strata. Criterion-level $\kappa$ estimates use $n=50$, except specificity ($n=47$ after excluding N/A--N/A pairs), and have substantial sampling uncertainty. The remaining 50 items have reference labels only.

\textbf{English only.} All prompts, queries, and responses are in English. The evaluation criteria assume English-language conventions for naturalness, formality, and specificity. Results may not generalize to multilingual settings.

\textbf{Single-turn only.} All conversations are single-turn. Multi-turn dialogue quality dimensions (coherence across turns, context tracking, topic management, conversational repair) are not captured.

\textbf{Fixed rubric.} The rubric is fixed at six criteria with predetermined scales and weights. Alternative rubric designs (different granularity, different criteria such as creativity or safety, domain-specific criteria) are not represented. The benchmark tests a judge's ability to apply this particular rubric, not to adapt to arbitrary rubrics.

\textbf{Static snapshot.} The dataset does not capture how chatbot quality evolves over time, across model versions, or in response to changing user expectations.

\FloatBarrier
\section{CHARM-100 evaluation results}
\label{appendix:charm100-results}

This section reports the full evaluation results for CHARM-100 using Gemini-3-Flash as the judge. For the dataset description, see Appendix~\ref{appendix:charm100-dataset}.

\begin{table}[ht]
\centering
\small
\begin{tabular}{llccccc}
\toprule
\textbf{Criterion} & \textbf{Type} & \textbf{Exact Acc.} & \textbf{Adj. Acc.} & \textbf{$\kappa$} & \textbf{Spearman} & \textbf{EMD} \\
\midrule
factual\_accuracy & Binary  & 87.0\% & ---   & 0.642\textsuperscript{u} & --- & --- \\
naturalness       & Ordinal & 58.0\% & 93.0\% & 0.719\textsuperscript{q} & 0.743 & 0.370 \\
satisfaction      & Ordinal & 42.0\% & 85.0\% & 0.648\textsuperscript{q} & 0.786 & 0.650 \\
specificity       & Ordinal & 39.5\% & 86.4\% & 0.549\textsuperscript{q} & 0.698 & 0.716 \\
helpfulness       & Ordinal & 38.0\% & 85.0\% & 0.625\textsuperscript{q} & 0.747 & 0.650 \\
response\_length  & Nominal & 81.0\% & ---   & 0.552\textsuperscript{u} & --- & --- \\
\bottomrule
\end{tabular}
\caption{Per-criterion results on CHARM-100 (100 samples, 6 criteria) with Gemini-3-Flash as judge. Specificity metrics use 81 complete cases after excluding the 19 pairs with an N/A label on either side; its exact and adjacent accuracies are 32/81 and 70/81, respectively. $\kappa$\textsuperscript{u} = unweighted Cohen's $\kappa$; $\kappa$\textsuperscript{q} = quadratic-weighted $\kappa$ (appropriate for ordinal scales, credits near-misses). EMD = Earth Mover's Distance between ground truth and predicted label distributions (ordinal steps; lower is better). Adjacent accuracy is within one ordinal step.}
\label{tab:charm100-per-criterion}
\end{table}

\begin{figure*}[ht]
    \centering
    \includegraphics[width=\linewidth]{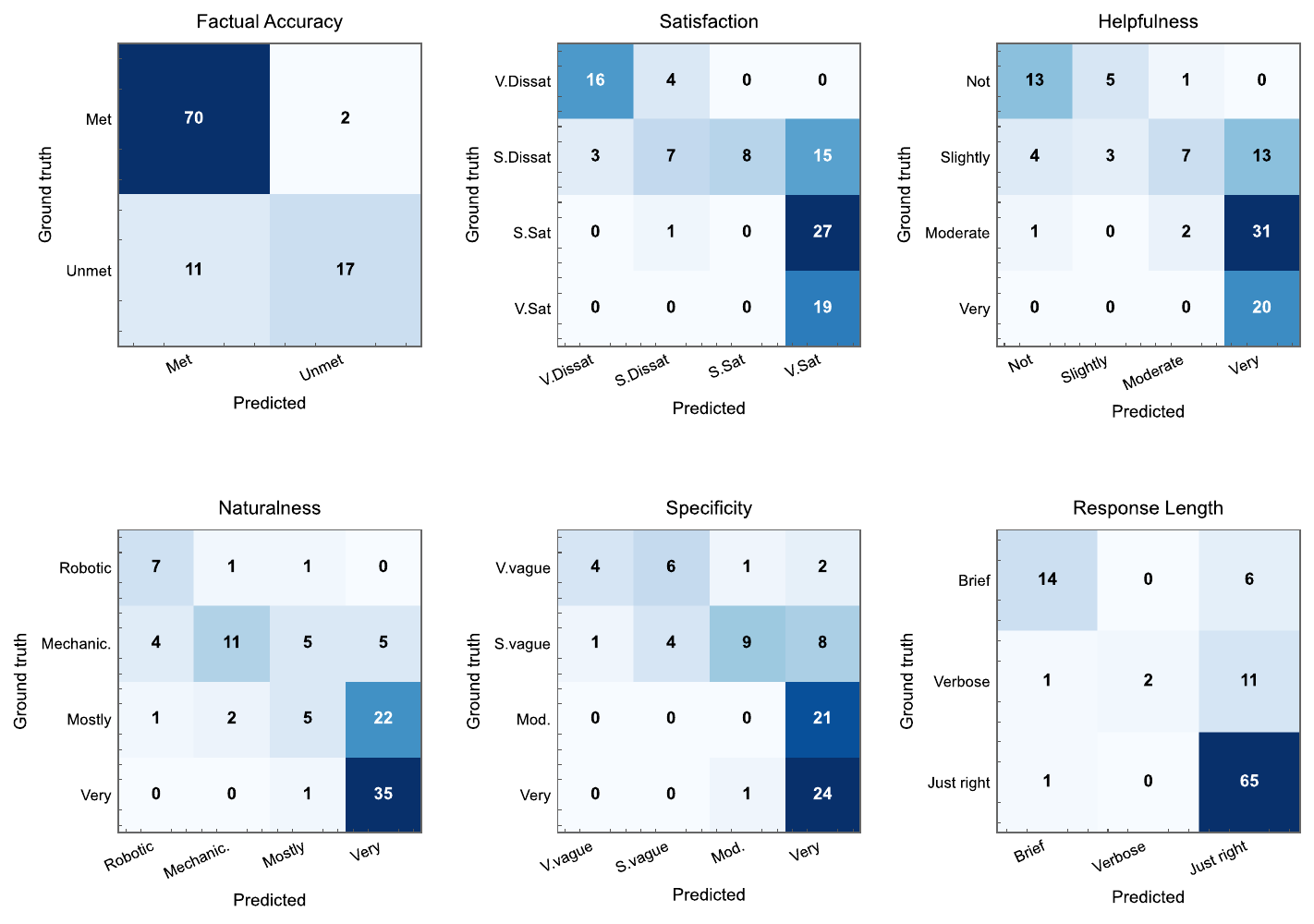}
    \caption{Confusion matrix heatmaps for all six CHARM-100 criteria. Ordinal criteria show a clear positive bias: predictions cluster toward the highest category. The specificity panel contains the 81 complete cases with neither label N/A.}
    \label{fig:charm100-confusion-heatmaps}
\end{figure*}

\begin{figure*}[ht]
    \centering
    \includegraphics[width=\linewidth]{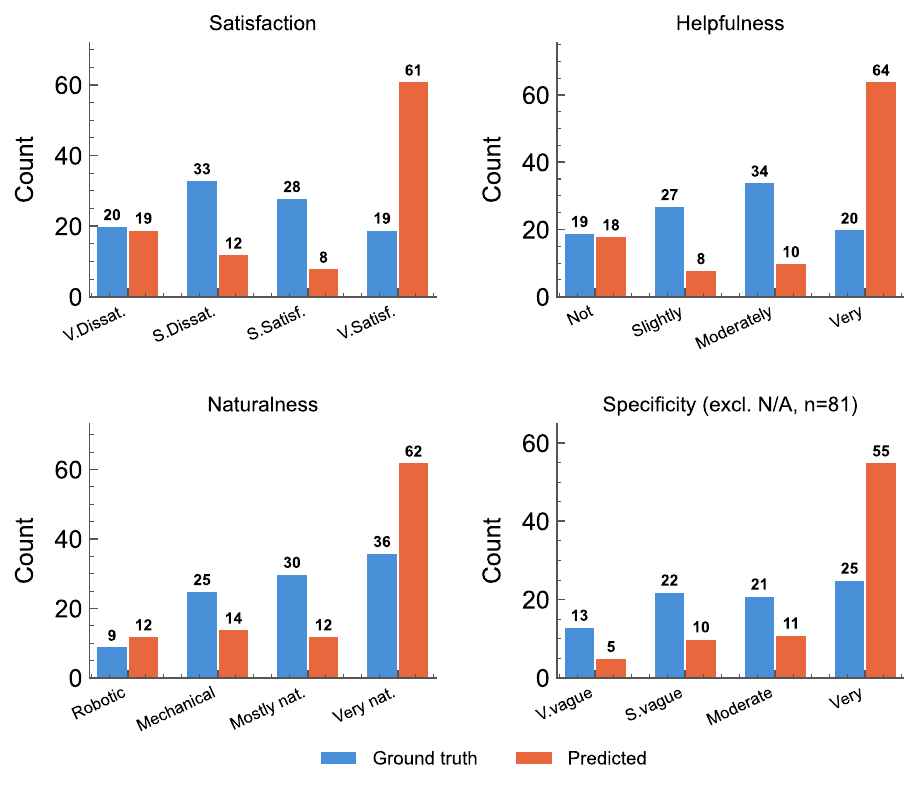}
    \caption{Ground truth vs.\ predicted label distributions for the four ordinal CHARM-100 criteria. The judge systematically over-predicts the highest category and under-predicts intermediate categories.}
    \label{fig:charm100-prediction-shift}
\end{figure*}

\subsection{Confusion matrices}

\textbf{Factual accuracy} (binary, rows = ground truth, columns = predicted):

\begin{table}[ht]
\centering
\small
\begin{tabular}{lrr}
\toprule
 & \textbf{MET} & \textbf{UNMET} \\
\midrule
\textbf{MET}   & 70 & 2 \\
\textbf{UNMET} & 11 & 17 \\
\bottomrule
\end{tabular}
\caption{Factual accuracy confusion matrix. High recall (0.97) but moderate precision (0.86) indicates the model occasionally marks incorrect responses as correct.}
\label{tab:charm100-factual-cm}
\end{table}

\textbf{Satisfaction} (ordinal, 4-point):

\begin{table}[ht]
\centering
\small
\begin{tabular}{lrrrr}
\toprule
 & \textbf{V.\ Dissat.} & \textbf{S.\ Dissat.} & \textbf{S.\ Sat.} & \textbf{V.\ Sat.} \\
\midrule
\textbf{V.\ Dissatisfied} & 16 & 4 & 0 & 0 \\
\textbf{S.\ Dissatisfied} & 3 & 7 & 8 & 15 \\
\textbf{S.\ Satisfied}    & 0 & 1 & 0 & 27 \\
\textbf{V.\ Satisfied}    & 0 & 0 & 0 & 19 \\
\bottomrule
\end{tabular}
\caption{Satisfaction confusion matrix. The model predicts ``Somewhat satisfied'' for only 8 samples (vs.\ 28 in ground truth) and assigns ``Very satisfied'' to 61 samples vs.\ 19 in ground truth.}
\label{tab:charm100-satisfaction-cm}
\end{table}

\textbf{Helpfulness} (ordinal, 4-point):

\begin{table}[ht]
\centering
\small
\begin{tabular}{lrrrr}
\toprule
 & \textbf{Not helpful} & \textbf{Slightly} & \textbf{Moderately} & \textbf{Very} \\
\midrule
\textbf{Not helpful at all} & 13 & 5 & 1 & 0 \\
\textbf{Slightly helpful}   & 4 & 3 & 7 & 13 \\
\textbf{Moderately helpful}  & 1 & 0 & 2 & 31 \\
\textbf{Very helpful}        & 0 & 0 & 0 & 20 \\
\bottomrule
\end{tabular}
\caption{Helpfulness confusion matrix. ``Moderately helpful'' receives 10 predictions vs.\ 34 in ground truth; ``Very helpful'' receives 64 vs.\ 20.}
\label{tab:charm100-helpfulness-cm}
\end{table}

\textbf{Naturalness} (ordinal, 4-point):

\begin{table}[ht]
\centering
\small
\begin{tabular}{lrrrr}
\toprule
 & \textbf{Robotic} & \textbf{Mechanical} & \textbf{Mostly nat.} & \textbf{Very nat.} \\
\midrule
\textbf{Robotic/unnatural}       & 7 & 1 & 1 & 0 \\
\textbf{Somewhat mechanical}     & 4 & 11 & 5 & 5 \\
\textbf{Mostly natural}          & 1 & 2 & 5 & 22 \\
\textbf{Very natural/human-like} & 0 & 0 & 1 & 35 \\
\bottomrule
\end{tabular}
\caption{Naturalness confusion matrix. The easiest ordinal criterion (weighted $\kappa = 0.719$), with most errors pulling toward ``Very natural.''}
\label{tab:charm100-naturalness-cm}
\end{table}

\textbf{Specificity} (ordinal, 4-point; $N=81$ complete cases, with pairs containing N/A on either side excluded):

\begin{table}[ht]
\centering
\small
\begin{tabular}{lrrrr}
\toprule
 & \textbf{Very vague} & \textbf{S.\ vague} & \textbf{Mod.\ spec.} & \textbf{Very spec.} \\
\midrule
\textbf{Very vague}          & 4 & 6 & 1 & 2 \\
\textbf{Somewhat vague}      & 1 & 4 & 9 & 8 \\
\textbf{Moderately specific} & 0 & 0 & 0 & 21 \\
\textbf{Very specific}       & 0 & 0 & 1 & 24 \\
\bottomrule
\end{tabular}
\caption{Specificity confusion matrix ($N=81$ complete cases; N/A on either side excluded). Exact accuracy is 32/81 (39.5\%), and adjacent accuracy is 70/81 (86.4\%). This is the hardest ordinal criterion ($\kappa = 0.549$); all ``Moderately specific'' samples are misclassified as ``Very specific.''}
\label{tab:charm100-specificity-cm}
\end{table}

\textbf{Response length} (nominal, 3-class):

\begin{table}[ht]
\centering
\small
\begin{tabular}{lrrr}
\toprule
 & \textbf{Too brief} & \textbf{Too verbose} & \textbf{Just right} \\
\midrule
\textbf{Too brief}   & 14 & 0 & 6 \\
\textbf{Too verbose}  & 1 & 2 & 11 \\
\textbf{Just right}   & 1 & 0 & 65 \\
\bottomrule
\end{tabular}
\caption{Response length confusion matrix. The model detects brevity (0.70 recall) but misses verbosity (0.14 recall). Of 14 verbose responses, 11 are classified as ``Just right.''}
\label{tab:charm100-length-cm}
\end{table}

\subsection{Aggregate metrics}

\begin{table}[ht]
\centering
\small
\begin{tabular}{lcc}
\toprule
\textbf{Metric} & \textbf{Value} & \textbf{95\% Bootstrap CI} \\
\midrule
Accuracy (binary criterion) & 87.0\% & [80.0\%, 93.0\%] \\
Mean $\kappa$ (all criteria; weighted for ordinal) & 0.623 & [0.451, 0.794] \\
Mean EMD (ordinal criteria) & 0.597 & --- \\
Spearman correlation (scores) & 0.810 & --- \\
Kendall correlation (scores) & 0.663 & --- \\
RMSE (scores) & 0.246 & [0.212, 0.274] \\
Mean bias & +0.170 & --- \\
\bottomrule
\end{tabular}
\caption{CHARM-100 aggregate metrics. The positive mean bias indicates that the model rates responses higher than the reference labels on average; no inferential test is reported for this statistic.}
\label{tab:charm100-aggregate}
\end{table}

\subsection{N/A handling analysis}

Across all 100 items, the judge predicts N/A 16 times versus 9 reference labels. The N/A-status confusion counts are TP${}=6$, FP${}=10$, FN${}=3$, and TN${}=81$ (87.0\% raw agreement; Cohen's $\kappa=0.412$). Treating N/A as positive, precision is 37.5\%, recall is 66.7\%, and Jaccard similarity is 31.6\%: six of nine reference N/A cases are recovered, but 10 of 16 predictions are false positives. Specificity label metrics exclude any pair with N/A on either side ($N=81$). By contrast, score aggregation under \texttt{SKIP} omits only predicted N/A verdicts from each item's weighted-score denominator.

\subsection{Configuration-sensitivity details}
\label{appendix:ablation}

\paragraph{Per-criterion-type breakdown.}
Table~\ref{tab:ablation-per-type} disaggregates the single-run configuration comparisons by criterion type. Removing few-shot examples changes binary exact accuracy by $-1.2$, $-3.7$, and $-15.0$pp for Gemini, GPT, and LLaMA, respectively; ordinal adjacent accuracy changes by $-2.5$, $-2.7$, and $+20.5$pp; and nominal exact accuracy changes by $+2.5$, $-2.5$, and $+21.2$pp. The sign reversals across criterion and model types reinforce that these runs measure configuration sensitivity plus output variability, not an isolated causal effect of few-shot calibration.

\begin{table*}[ht]
\centering
\small
\begin{tabular}{llcccccc}
\toprule
& & \multicolumn{2}{c}{\textbf{Binary}} & \multicolumn{2}{c}{\textbf{Ordinal}} & \multicolumn{2}{c}{\textbf{Nominal}} \\
\textbf{Configuration} & \textbf{Model} & Acc & $\kappa$ & Adj.Acc & $\kappa_q$ & Acc & $\kappa$ \\
\midrule
Default & Gemini-3-Flash & 90.0\% & 0.689 & 91.4\% & 0.690 & 81.2\% & 0.626 \\
 & GPT-5.4-nano & 72.5\% & 0.362 & 83.2\% & 0.502 & 72.5\% & 0.404 \\
 & LLaMA-3.1-8B & 41.2\% & 0.026 & 31.9\% & -0.008 & 18.8\% & 0.000 \\
\addlinespace
$-$Shuffle & Gemini-3-Flash & 90.0\% & 0.689 & 90.3\% & 0.679 & 77.5\% & 0.556 \\
 & GPT-5.4-nano & 72.5\% & 0.362 & 85.0\% & 0.462 & 70.0\% & 0.341 \\
 & LLaMA-3.1-8B & 68.8\% & 0.264 & 31.9\% & -0.016 & 18.8\% & 0.000 \\
\addlinespace
$-$Few-shot & Gemini-3-Flash & 88.8\% & 0.657 & 88.9\% & 0.661 & 83.8\% & 0.638 \\
 & GPT-5.4-nano & 68.8\% & 0.329 & 80.5\% & 0.488 & 70.0\% & 0.326 \\
 & LLaMA-3.1-8B & 26.2\% & 0.022 & 52.4\% & 0.120 & 40.0\% & 0.018 \\
\addlinespace
+Ens($k{=}3$, maj.) & Gemini-3-Flash & 90.0\% & 0.689 & 90.4\% & 0.677 & 82.5\% & 0.643 \\
 & GPT-5.4-nano & 72.5\% & 0.362 & 82.9\% & 0.485 & 70.0\% & 0.323 \\
 & LLaMA-3.1-8B & 51.2\% & 0.111 & 32.2\% & -0.002 & 20.0\% & 0.008 \\
\addlinespace
+Ens($k{=}5$, maj.) & Gemini-3-Flash & 90.0\% & 0.689 & 91.1\% & 0.690 & 81.2\% & 0.622 \\
 & GPT-5.4-nano & 72.5\% & 0.362 & 83.5\% & 0.509 & 70.0\% & 0.341 \\
 & LLaMA-3.1-8B & 67.5\% & 0.270 & 32.2\% & -0.002 & 18.8\% & 0.000 \\
\addlinespace
+Ens($k{=}3$, mean) & Gemini-3-Flash & 90.0\% & 0.689 & 90.4\% & 0.678 & 81.2\% & 0.622 \\
 & GPT-5.4-nano & 72.5\% & 0.362 & 84.1\% & 0.496 & 70.0\% & 0.357 \\
 & LLaMA-3.1-8B & 70.0\% & 0.281 & 32.8\% & -0.007 & 20.0\% & 0.008 \\
\addlinespace
Bare (no mitigations) & Gemini-3-Flash & 88.8\% & 0.657 & 88.5\% & 0.630 & 81.2\% & 0.593 \\
 & GPT-5.4-nano & 68.8\% & 0.329 & 79.7\% & 0.486 & 68.8\% & 0.297 \\
 & LLaMA-3.1-8B & 26.2\% & 0.004 & 51.6\% & -0.030 & 31.2\% & -0.114 \\
\bottomrule
\end{tabular}
\caption{Per-criterion-type breakdown of the single-run CHARM-100 configuration-sensitivity analysis. Binary = factual\_accuracy (1 criterion); Ordinal = satisfaction, helpfulness, naturalness, specificity (4 criteria, mean reported); Nominal = response\_length (1 criterion). $\kappa_q$ = quadratic-weighted $\kappa$.}
\label{tab:ablation-per-type}
\end{table*}

\paragraph{Cross-family ensemble.}
The cross-family ensemble reconstructs a 3-judge panel from the individual Default runs (one per model family). For each item and criterion, verdicts from the three models are combined via majority vote. It achieves 57.1\% pooled exact accuracy, $\kappa = 0.626$, $\rho = 0.769$, and RMSE $= 0.200$. Relative to Gemini alone, exact accuracy and $\kappa$ are lower (60.4\% and 0.679 for Gemini), while score RMSE is slightly lower (0.213 for Gemini). Inter-family agreement (61.9\%) is substantially lower than same-model ensembles (86--98\%), and the cross-family panel costs the sum of all three judges' inference. The stored runs therefore show no general accuracy or reliability advantage over the strongest single judge.

\FloatBarrier
\section{Agent skill improvement details}
\label{appendix:skill-improvement}

This appendix provides additional details on the agent skill improvement application (Section~\ref{sec:skill-improvement}).

\textbf{Tradeoff analysis.} Two tradeoffs emerge from per-criterion analysis. First, thoroughness competes with conciseness: the revised skill's detailed structural requirements cause \texttt{concise\_review} to drop from 100\% to 50\%, while the curated skill maintains 100\%. The revised skill is tightly coupled to the rubric and model; a different rubric might reward different tradeoffs. Second, \texttt{factual\_misrepresentation} remains a recurring error: even with explicit source-checking constraints, Llama 3.1 8B fabricates details in 40--60\% of reviews in this 10-paper setup. The longer, more structured output format creates more opportunities for hallucination, and this revision does not resolve the problem.

\textbf{Implications.} Per-criterion pass rates provide a more actionable optimization signal than aggregate scores: the revision LLM can target specific failing criteria rather than optimizing a scalar. In this setup, a small model improves from 0.47 to 0.85 after rubric-guided skill revision. Under the primary judge, the revised point estimate slightly exceeds the expert-curated point estimate of 0.82, but the overlapping CIs and absence of a revised-versus-curated superiority test do not establish outperformance. Factual-grounding errors remain frequent and are not resolved by this revision.

\begin{figure}[ht]
    \centering
    \includegraphics[width=0.9\linewidth]{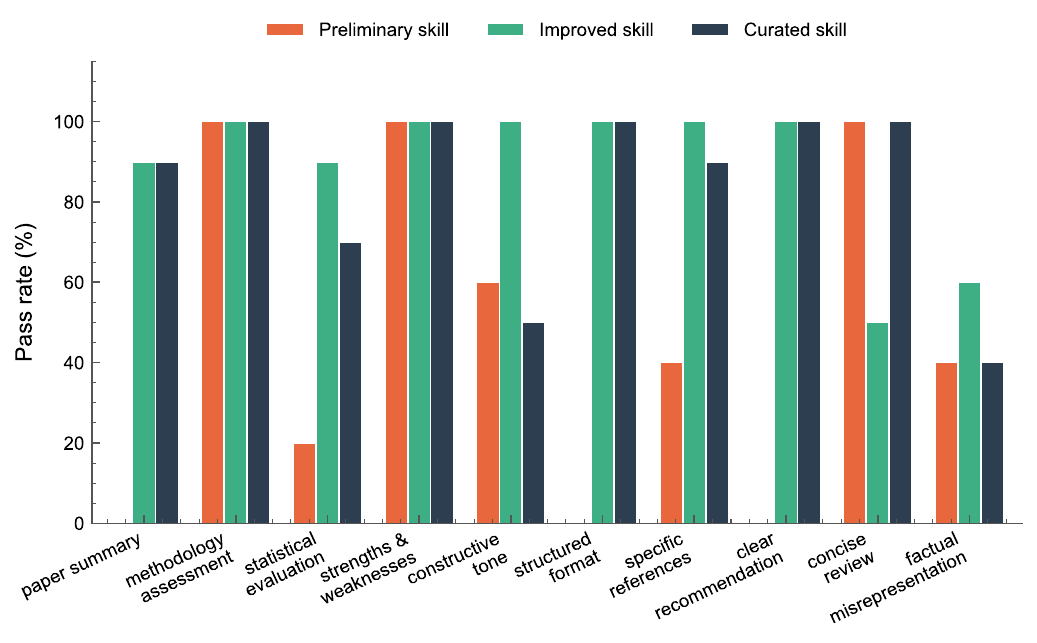}
    \caption{Per-criterion pass rates across three skill conditions (10 papers, 10 criteria). The improved skill matches or exceeds the curated skill on 9 of 10 criteria in this run.}
    \label{fig:skill-threeway}
\end{figure}

\subsection{Cross-model robustness probes}

The original experiment uses Gemini-3-Flash as both the rubric grading judge and the revision model, raising two concerns: (1) the revision model may produce skill instructions that align especially well with the same model's grading preferences (self-preference), and (2) the grading judge may have systematic biases that inflate post-revision scores regardless of which model revised the skill.

\textbf{Cross-revision probe.} To examine concern (1), we rerun the full pipeline with GPT-5.4 as the revision model while keeping Gemini-3-Flash as the rubric grading judge. Each revision arm comprises 3 independent executions on the same 10-paper set (30 scores per revision row). The displayed vague and expert-curated comparators pool their separately generated controls across both arms (6 executions, 60 scores per comparator row). We compute descriptive 95\% IID per-score bootstrap intervals over 10{,}000 resamples; these intervals do not account for repeated paper identity across executions. The conditions independently generate and grade their vague-skill reviews, so the revisers receive separately sampled pass rates and failure explanations; they also independently generate and grade their post-revision reviews. The comparison therefore changes the reviser together with upstream feedback and downstream generation and grading realizations. The resulting point estimates are similar (0.85 vs.\ 0.86) with overlapping CIs, but this pattern neither isolates revision-model self-preference nor establishes equivalence. Both point estimates slightly exceed the pooled expert-curated point estimate of 0.82, but the revised-versus-curated CIs overlap and no superiority test was conducted.

\textbf{Cross-grading probe.} To examine concern (2), we take the Gemini-revised skill from each run, generate fresh reviews with Llama 3.1 8B, and grade them with GPT-5.4-mini as a second rubric grading judge. Table~\ref{tab:skill-cross-judge} reports both sets of results. Under GPT-5.4-mini, the vague-to-revised point-estimate increase persists ($0.58 \to 0.74$), while the revised point estimate is below the expert-curated point estimate (0.78); their CIs overlap. Relative to the original Gemini-graded run, the GPT-5.4-mini run assigns a higher score to the vague condition and a lower score to the revised condition. Because the second run uses freshly generated reviews, these between-run differences reflect both review sampling and judge choice and cannot estimate relative judge calibration or grading-judge bias. The repeated vague-to-revised direction is therefore an end-to-end robustness observation under a joint change in generated reviews and grader. Testing transfer to a perturbed rubric remains a direction for future work; the current experiment does not test rubric-specific overfitting.

\begin{table}[ht]
\centering
\small
\begin{tabular}{lcc}
\toprule
\textbf{Condition} & \textbf{Gemini-3-Flash [95\% CI]} & \textbf{GPT-5.4-mini [95\% CI]} \\
\midrule
Vague skill & 0.47 [0.44, 0.51] & 0.58 [0.52, 0.63] \\
Expert curated & 0.82 [0.79, 0.85] & 0.78 [0.74, 0.83] \\
\midrule
Gemini-3-Flash revision & 0.85 [0.81, 0.89] & 0.74 [0.69, 0.78] \\
GPT-5.4 revision & 0.86 [0.82, 0.90] & --- \\
\bottomrule
\end{tabular}
\caption{Skill-revision robustness results under two graders. In the Gemini-3-Flash column, vague and expert-curated rows pool 6 runs $\times$ 10 papers ($N=60$), while each revision row uses 3 runs $\times$ 10 papers ($N=30$). The GPT-5.4-mini column uses 3 runs $\times$ 10 freshly generated reviews per reported row. Column differences therefore conflate review sampling and judge choice and are not cross-judge calibration estimates. Both end-to-end evaluations show a large vague-to-Gemini-revised point-estimate increase. Under Gemini grading, the two revision point estimates (0.85 and 0.86) slightly exceed the expert-curated point estimate (0.82); under GPT grading, the Gemini-revised point estimate (0.74) is below the expert-curated point estimate (0.78). Revised-versus-curated CIs overlap in every reported comparison, and no superiority or equivalence test was conducted. The 95\% CIs are descriptive IID per-score bootstrap intervals over 10{,}000 resamples and do not account for repeated paper identity across runs.}
\label{tab:skill-cross-judge}
\end{table}

\begin{figure*}[ht]
    \centering
    \includegraphics[width=\linewidth]{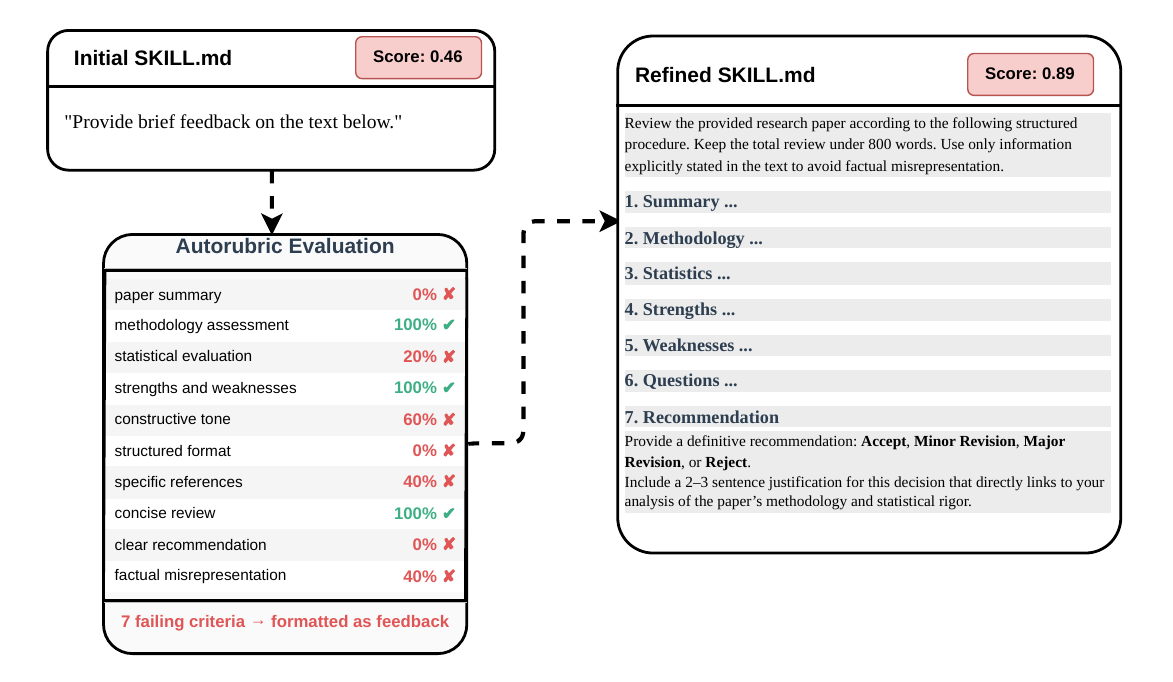}
    \caption{Concrete example from one rubric-guided revision run. The initial one-line skill (run score 0.46) is evaluated against 10 binary criteria; 7 failing criteria are formatted as structured feedback for the revision LLM, which produces a detailed skill with explicit sections addressing each deficiency (run score 0.89). These illustrative run scores differ from the pooled point estimates of 0.47 and 0.85 reported in Figure~\ref{fig:skill-improvement}.}
    \label{fig:skill-transformation}
\end{figure*}

\FloatBarrier
\section{Cross-benchmark reliability}
\label{appendix:cross-benchmark}

In the primary CHARM-100 configuration, Gemini-3-Flash shows scale-extreme clustering on the four ordinal criteria: adjacent accuracy is high (85--93\%) despite lower exact accuracy (38--58\%), with the judge avoiding intermediate categories. These observations motivate evaluating coarser scales or adjacent accuracy for similar rubrics, but CHARM-100 does not compare scale widths. Because criterion identity is confounded with type and $\kappa$ weighting differs by type, these data do not support binary criteria as a universal default.

\begin{figure}[ht]
    \centering
    \includegraphics[width=0.9\linewidth]{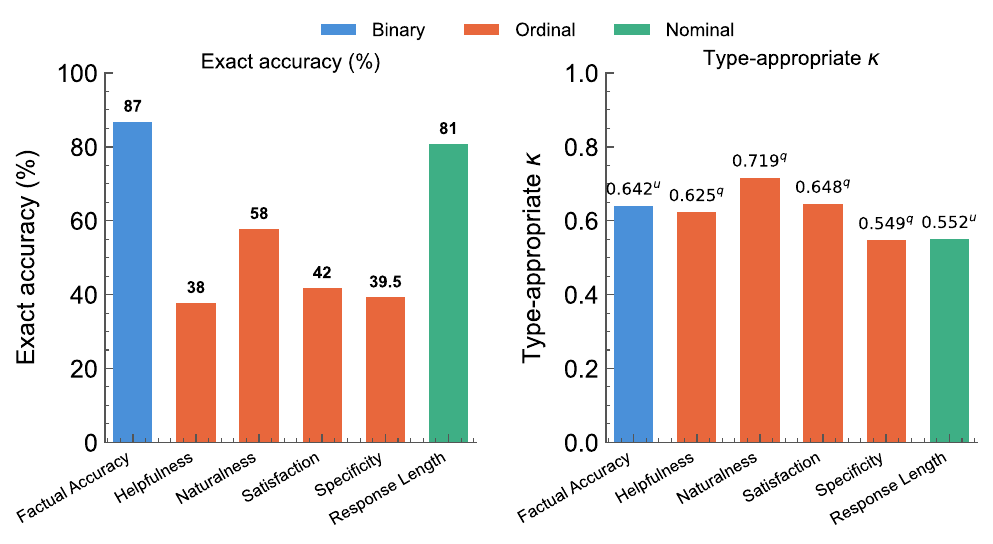}
    \caption{CHARM-100 criterion-level agreement under Gemini-3-Flash, grouped by type. Ordinal exact accuracy (44.4\%) is lower than factual accuracy (87\%) and response-length accuracy (81\%), while ordinal adjacent accuracy is 87.3\%. The displayed $\kappa$ is unweighted for binary and nominal criteria and quadratic-weighted for ordinal criteria; these values characterize criterion-specific agreement rather than a common-metric ranking of types.}
    \label{fig:cross-benchmark-reliability}
\end{figure}

\FloatBarrier
\section{RL training details}
\label{appendix:rlrr}

This appendix provides additional training diagnostics for the RL experiment in Section~\ref{sec:rlrr}.

\begin{figure}[ht]
    \centering
    \includegraphics[width=0.9\linewidth]{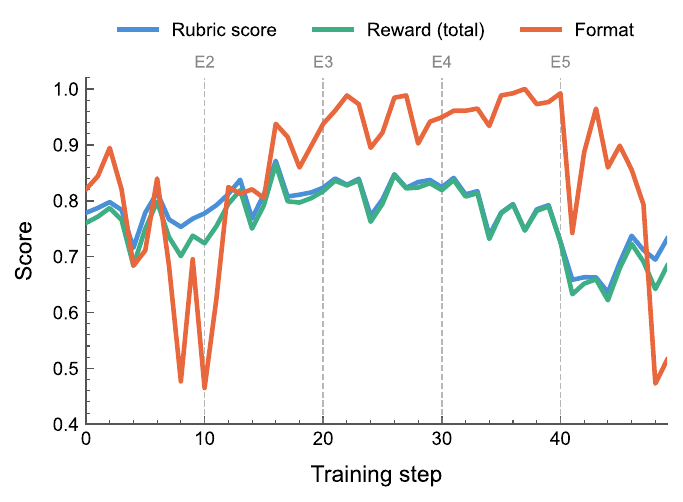}
    \caption{Training rubric score, reward, and format compliance across 50 steps (5 epochs, boundaries marked). Rubric scores improve through epoch~3, then show late-training degradation.}
    \label{fig:rlrr-training-curve}
\end{figure}

Figure~\ref{fig:rlrr-training-curve} shows per-step training metrics. Mean training rubric score improves from 0.774 (epoch 1) to 0.825 (epoch 3), then declines in epochs 4 and 5 (0.793 and 0.698), consistent with late-training degradation on the 321 training prompts.

\begin{figure}[ht]
    \centering
    \includegraphics[width=0.9\linewidth]{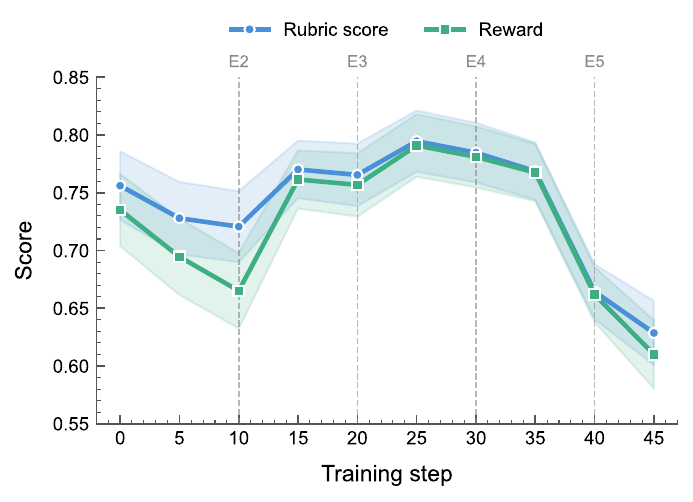}
    \caption{Rubric score on the 81-prompt validation split at ten monitored checkpoints, with standard error bands. Step~25, selected after observing this curve, has the highest mean score (0.795 versus 0.756 at step~0; Cohen's $d = 0.26$, 95\% CI $[0.04, 0.51]$). All checkpoints at steps 15--35 have positive mean differences, but no independent test split was evaluated.}
    \label{fig:rlrr-eval-progression}
\end{figure}

Figure~\ref{fig:rlrr-eval-progression} shows evaluation on the 81-prompt validation split every five steps. Step~25 is the observed maximum (0.795) among the ten monitored checkpoints and is selected for reporting; this was not a prespecified early-stopping rule, and training continued through step~49. The mean validation score falls to 0.629 by the step-45 evaluation. Checkpoints at steps 15--35 all have positive mean differences from the step-0 baseline, but these repeated measurements on the selection split are descriptive rather than independent confirmation.

\begin{figure*}[ht]
    \centering
    \includegraphics[width=\linewidth]{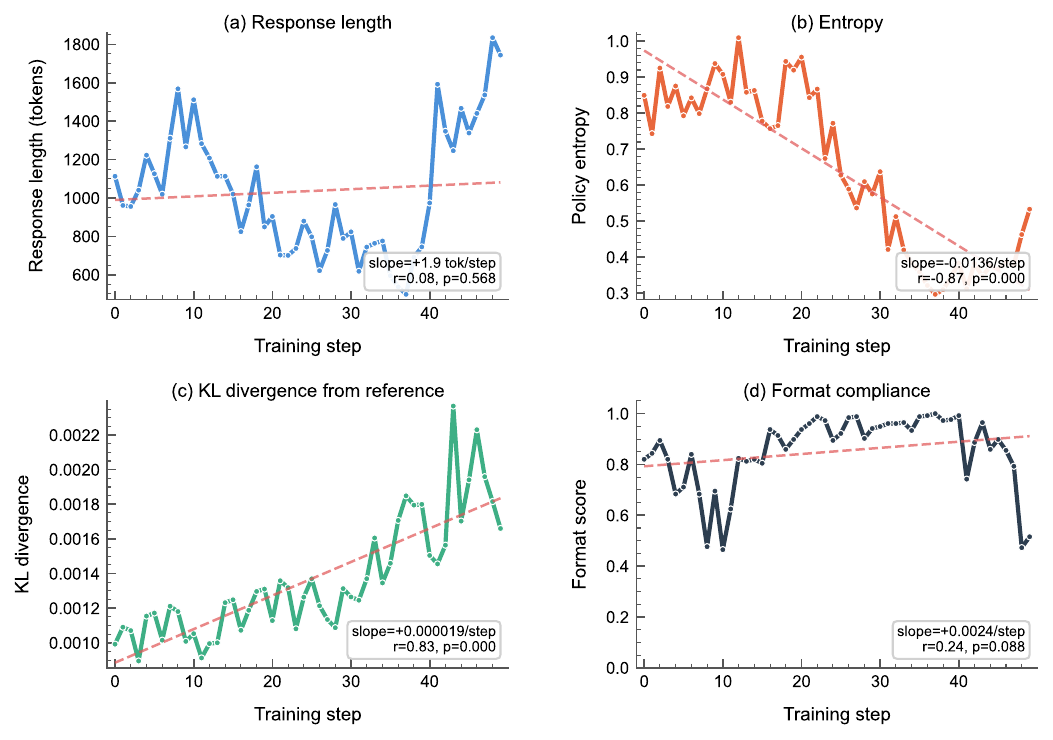}
    \caption{Training health indicators across 50 steps. (a) Response length is non-monotonic: the epoch mean falls from 1{,}158 tokens in epoch~1 to 782 in epoch~3, then rises to 1{,}452 in epoch~5; over all batches, the linear slope is +1.9 tokens/step ($r = 0.08$, $p = 0.568$). (b) Entropy declines from 1.0 to 0.3. (c) KL divergence from the reference policy stays below 0.003. (d) Format compliance fluctuates around 0.80 with no detectable trend.}
    \label{fig:rlrr-training-health}
\end{figure*}

Figure~\ref{fig:rlrr-training-health} tracks four health indicators. Response length is non-monotonic across training, while KL divergence remains bounded and format compliance is stable. These diagnostics do not show a simple monotonic verbosity-exploitation pattern, but they cannot exclude other forms of reward misspecification. The entropy decline in later epochs accompanies the late-training score decline.

\subsection{Cross-judge robustness probe}

The RL experiment uses \texttt{gemini-3-flash-preview} as both the training reward judge and the evaluation judge. To probe judge dependence, we regenerate fresh responses from the base model and the step-25 checkpoint on all 81 validation prompts (temperature 1.0, matching training conditions) and grade both sets with GPT-5.4-mini as an independent judge. Because these are not the same sampled responses used by the original judge, the comparison changes both response realization and judge and therefore cannot isolate judge coupling.

Table~\ref{tab:rlrr-cross-judge} compares the two evaluations. Under GPT-5.4-mini, the trained-model mean also rounds to 0.795, while the base-model mean is higher (0.772 vs.\ 0.756), reducing the observed difference from +0.039 to +0.023. The nominal paired two-sided Wilcoxon signed-rank result is inconclusive ($p = 0.243$; Cohen's $d = 0.13$, 95\% bootstrap CI $[-0.09, 0.35]$). Per prompt, 25 validation cases improve, 38 remain unchanged, and 18 regress.

The attenuated, directionally consistent difference and the IFEval transfer results (Section~\ref{sec:rlrr}) are useful robustness probes, but neither provides a confirmatory estimate of the selected checkpoint's in-domain effect.

\begin{table}[ht]
\centering
\begin{tabular}{lcc}
\toprule
 & Gemini-3-Flash & GPT-5.4-mini \\
 & (original) & (independent) \\
\midrule
Base mean score & 0.756 & 0.772 \\
Trained mean score & 0.795 & 0.795 \\
$\Delta$ & +0.039 & +0.023 \\
Paired two-sided Wilcoxon $p$ & 0.032 & 0.243 \\
Cohen's $d$ [95\% CI] & 0.26 [0.04, 0.51] & 0.13 [$-$0.09, 0.35] \\
Rubric-score-1.0 count & 21 $\to$ 30 & 25 $\to$ 26 \\
\bottomrule
\end{tabular}
\caption{Cross-judge robustness probe on 81 AdvancedIF validation prompts. The GPT-5.4-mini column uses freshly regenerated responses, so differences between columns reflect both judge and generation variability. Reported Wilcoxon $p$-values are nominal, paired, two-sided, and unadjusted for checkpoint selection.}
\label{tab:rlrr-cross-judge}
\end{table}

\end{document}